\documentclass[]{ijimai_article}

\usepackage{orcidlink}
\usepackage{float}
\usepackage{placeins}
\usepackage{multirow} 
\usepackage{pifont} 

\usepackage{zref-user} 
\usepackage{xstring}
\usepackage{mfirstuc}
\usepackage{etoolbox}
\usepackage{xparse}
\usepackage{xstring}    
\usepackage{ifthen}     

\newcommand{\Capitalize}[1]{%
  \StrLeft{#1}{1}[\firstletter]%
  \StrGobbleLeft{#1}{1}[\rest]%
  \MakeUppercase{\firstletter}\lowercase{\rest}%
}

\newcommand{\customrefaux}[3]{%
  \StrBefore{#1}{:}[\prefix]%
  \Capitalize{\prefix}%
  \ifnum#2=2 s\fi.~\ref{#1}%
}

\newcommand{\customref}[2][0]{%
  \ifnum#1=1
    \ref{#2}%
  \else
    \StrBefore{#2}{:}[\prefix]%
    \Capitalize{\prefix}.~\ref{#2}%
  \fi
}

\newcommand{\customrefs}[2][0]{%
  \ifnum#1=1
    \ref{#2}%
  \else
    \StrBefore{#2}{:}[\prefix]%
    \Capitalize{\prefix}s.~\ref{#2}%
  \fi
}

\newcounter{rq}
\renewcommand{\therq}{\arabic{rq}}
\newcommand{\rqref}[1]{RQ.~\ref{#1}}
    
\newcommand{\RQitem}[2]{%
  \refstepcounter{rq}%
  \item[\textbf{RQ\therq}] \label{#1}#2
}

\newcommand{\department}{Computer Systems Engineering Dpt.}
\newcommand{\affilUPM}{Universidad Politécnica de Madrid}

\newcommand{\departJames}{Department of Computer Science and Engineering}
\newcommand{\affilJames}{Seoul National University of Science and Technology}

\newcommand{\myAuthor}[4]{\textbf{#1~#2}\textsuperscript{#3}\orcidlink{#4}}

\title{Decoding Latent Spaces: Assessing the Interpretability of Time Series Foundation Models for Visual Analytics}

\author{
    \myAuthor{Inmaculada}{Santamaria-Valenzuela}{1}{0000-0002-7497-8795}
    \and \myAuthor{Victor}{Rodriguez-Fernandez}{1}{0000-0002-8589-6621}
    \and \myAuthor{Javier}{Huertas-Tato}{1}{0000-0003-4127-5505}
    \and \myAuthor{Jong}{Hyuk Park}{2}{0000-0003-1831-0309}
    \and \myAuthor{David}{Camacho}{1}{0000-0002-5051-3475}
}

\affiliation{
\textsuperscript{1} \department, \affilUPM, Madrid (Spain) \\ 
\textsuperscript{2} \departJames, \affilJames, Seoul (Korea)}

\mailinfo{E-mail address: mi.santamaria@upm.es, victor.rfernandez@upm.es, javier.huertas.tato@upm.es, jhpark1@seoultech.ac.kr, david.camacho@upm.es}

\abstract{The present study explores the interpretability of latent spaces produced by time series foundation models, focusing on their potential for visual analysis tasks. Specifically, we evaluate the MOMENT family of models, a set of transformer-based, pre-trained architectures for multivariate time series tasks such as: imputation, prediction, classification, and anomaly detection. We evaluate the capacity of these models on five datasets to capture the underlying structures in time series data within their latent space projection and validate whether fine tuning improves the clarity of the resulting embedding spaces. Notable performance improvements in terms of loss reduction were observed after fine tuning. Visual analysis shows limited improvement in the interpretability of the embeddings, requiring further work. Results suggest that, although Time Series Foundation Models such as MOMENT are robust, their latent spaces may require additional methodological refinements to be adequately interpreted, such as alternative projection techniques, loss functions, or data preprocessing strategies. Despite the limitations of MOMENT, foundation models supose a big reduction in execution time and so a great advance for interactive visual analytics.
}

\begin{document}

\maketitle

\section{Introduction}
\label{sec:introduction}
\fancyFirstWord{Visual} analytics are essential for interpreting large time series data on fields like finance, enabling human understanding and supporting algorithm assisted decision making~\cite{wu2025requirement,zhang2025visual}. In this field, understanding Deep Learning (DL) models' embedding space has been a major focus. Different interactive visualization tools support dataset analysis by exploring these embedding spaces~\cite{tensorflow_embedding_projector, umap_interactive_viz, rodriguezfernandez2023deepvats}.

In other domains like computer vision and natural language processing, Transfer Learning (TL) has proven effective at reusing the model knowledge accross datasets enabling faster analysis and reducing training time in new scenarios. Bringing this efficiency into the visual analytics of time series could significantly enhance the interactivity. 
However, unlike image or text data, TS does not always exhibit  shared semantic information between datasets and domains -the same change on the shape of a TS could be an anomaly in a machine analysis but not in stock analysis. This raises important questions about how well TL can be adapted to TS~\cite{woo2023pushing}. However, TL is used for different TS tasks, such as forecasting workload on cloud platforms \cite{zuo2025mixed} or forecasting influenza hospitalization with small datasets \cite{meyer2025prospective}. Also, Fawaz et al.~\cite{fawaz2018transfer} motivates practitioners to no longer train models from scratch for TS classification but instead to fine-tune pre-trained models, based on the idea that models can be adapted. But this still raises various questions. First, ``is the size of the dataset the only important thing or is context explicitly neccesary too?'' Second, ``even in the same domain, is it possible to swap semantics between datasets?'', ``is it enough with a fine-tune or does a TS model need extra semantic information for building the TS?''. 

\begin{figure}[!thb]
\centering
\setlength{\tabcolsep}{4pt}
\begin{tabular}{ccc}
& \textbf{Trend} & \textbf{Auto-correlation} \\
\raisebox{0.15\linewidth}{\textbf{PCA}}
 &
\includegraphics[height=0.25\linewidth, pagebox=artbox]{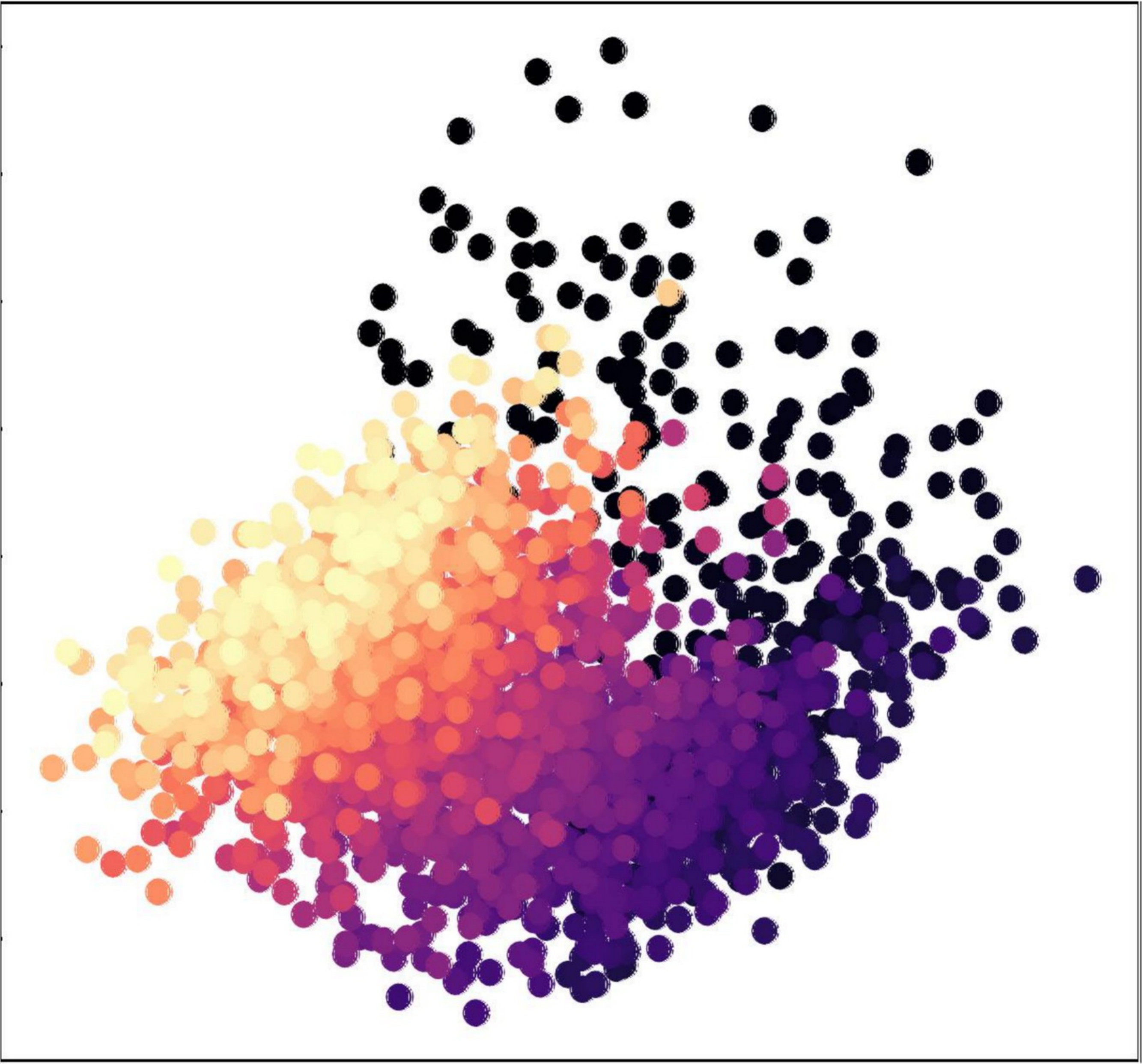} & 
\includegraphics[height=0.25\linewidth, pagebox=artbox]{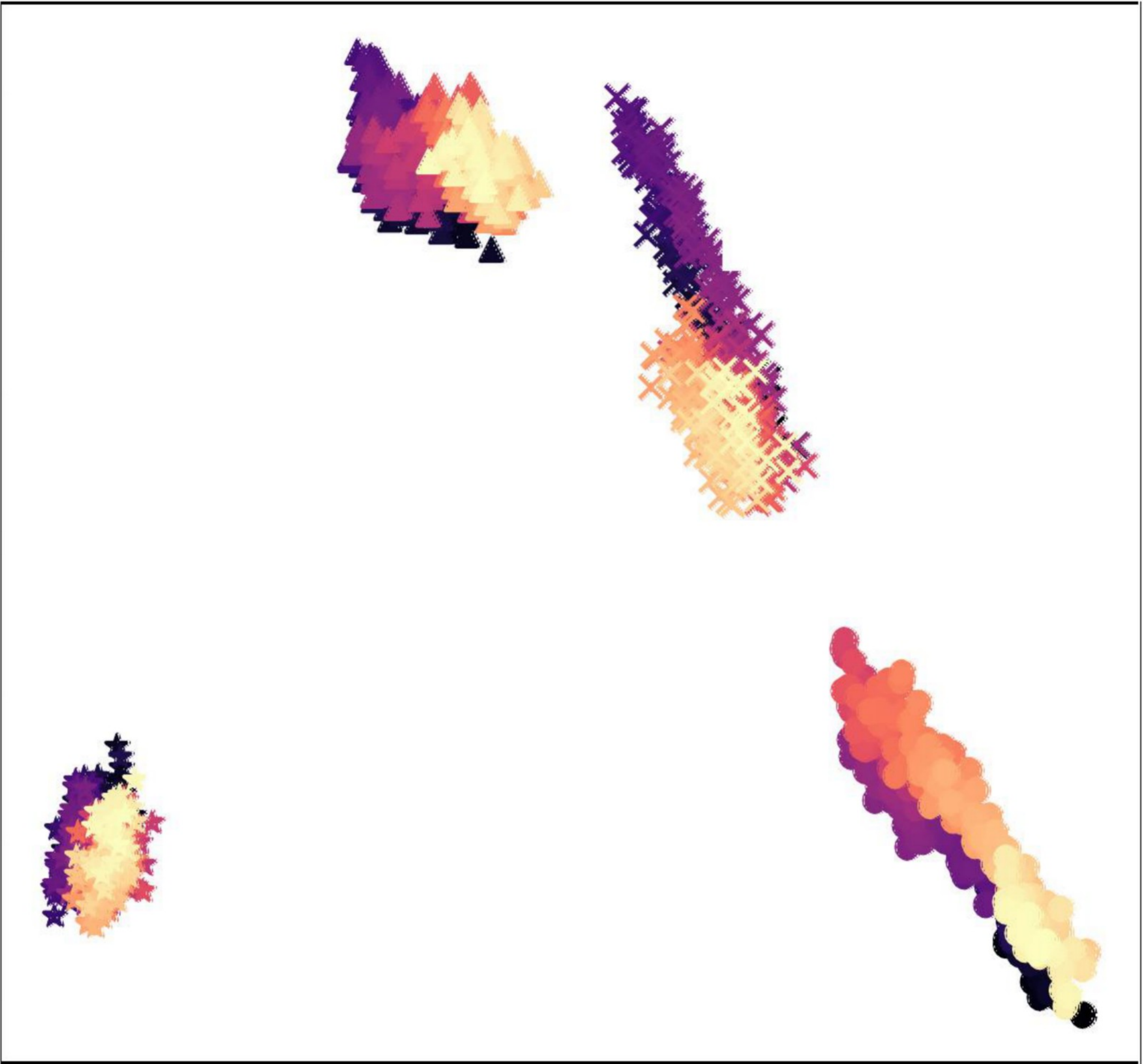} \\
\raisebox{0.17\linewidth}{\textbf{t-SNE}}
 &
 \hspace{-12pt}
\includegraphics[height=0.25\linewidth, pagebox=artbox]{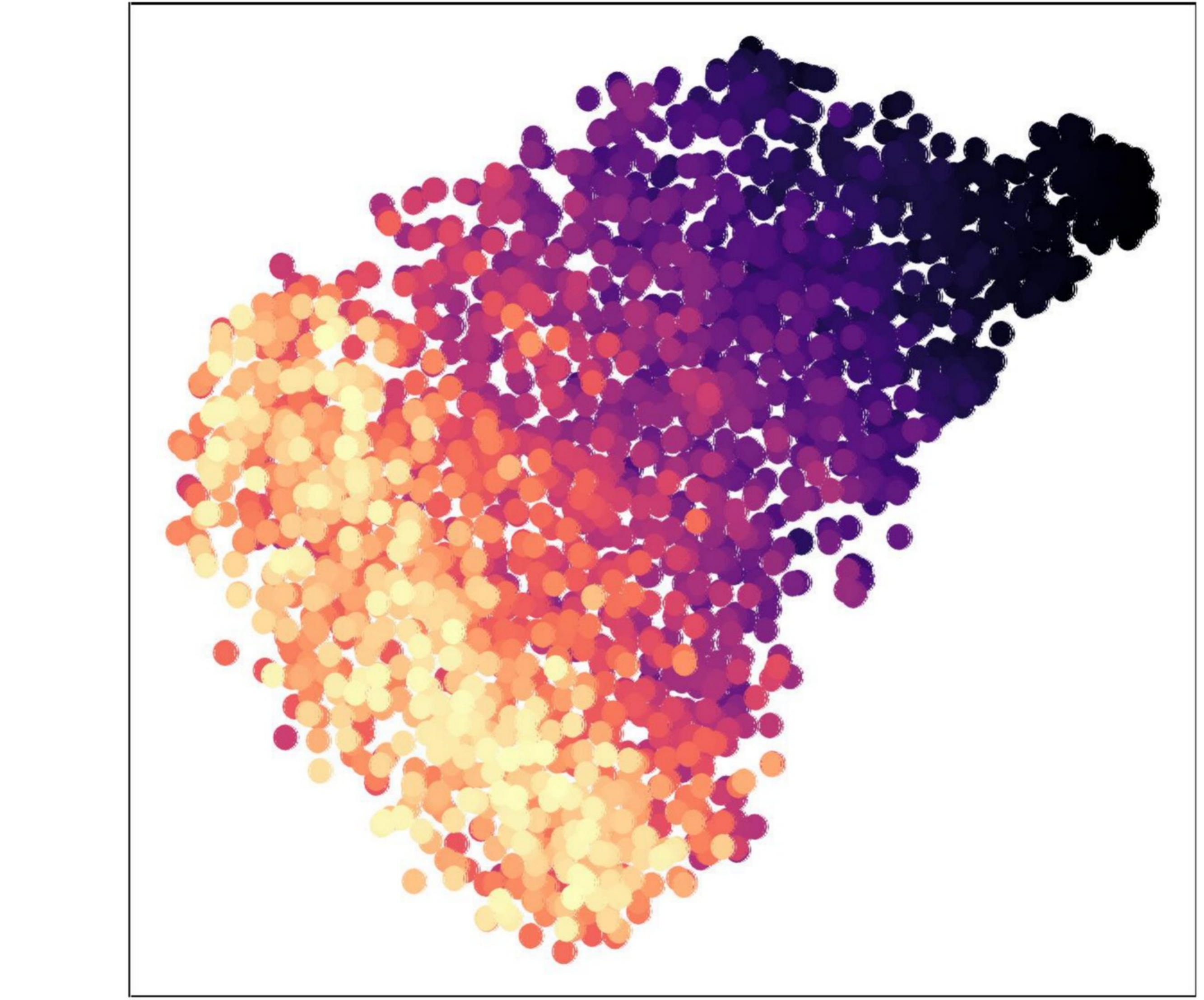} & 
\includegraphics[height=0.25\linewidth, pagebox=artbox]{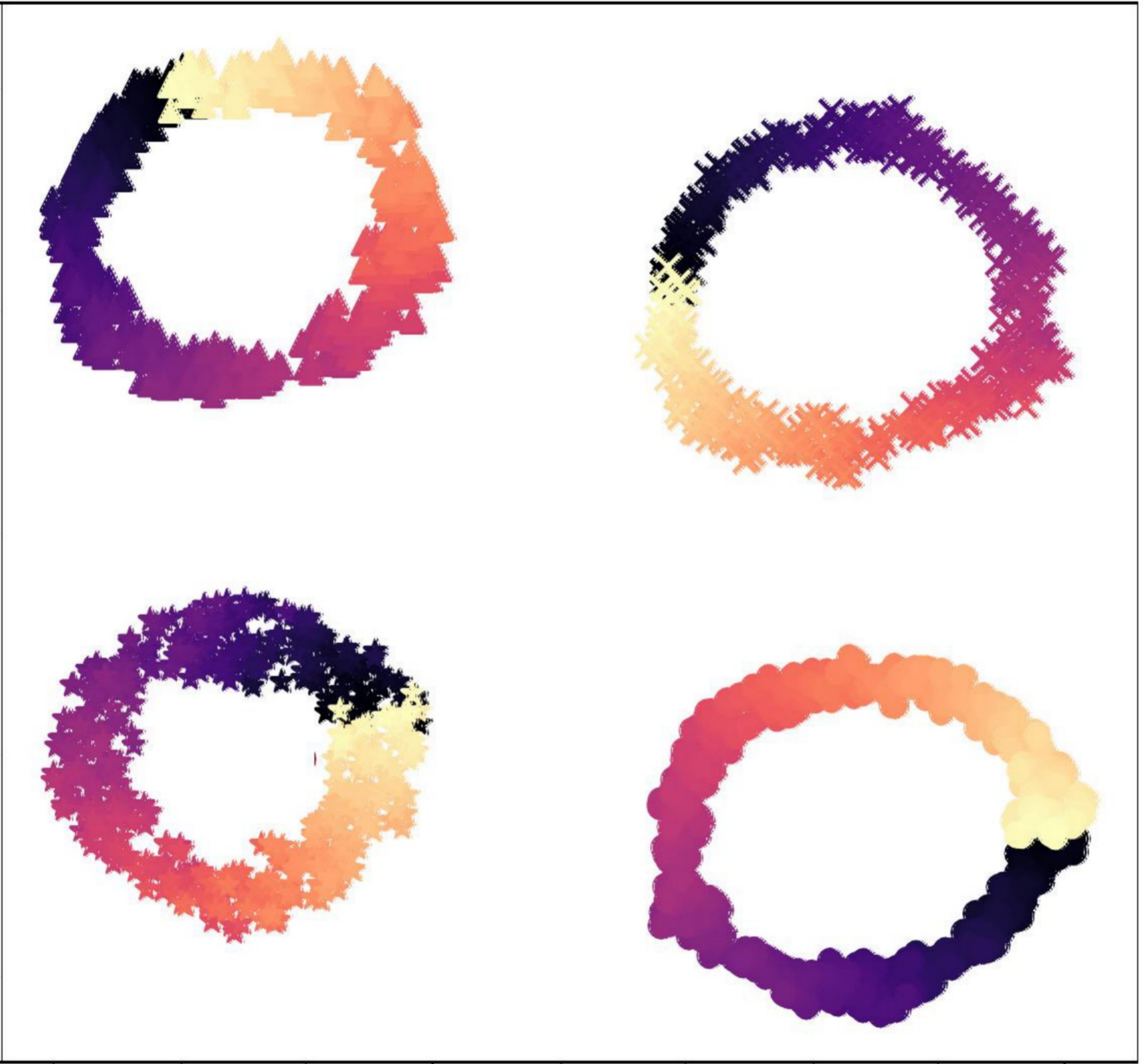} \\
\end{tabular}
\caption{MOMENT's embedding space for synthetically generated datasets projection using PCA (top) and t-SNE (bottom) projections.  Shows subtle trend and auto-correlation are captured. This suggests that TSFM embeddings may encode rich structural features that could be exploited for interactive exploration. Figure modified from~\cite{goswami2024moment}.}
\label{fig:momentfm:projections}
\end{figure}

Following this path, recent studies have shown great advances on time series foundation models (TSFM), with good performance in the different tasks (classification, anomaly detection, etc.), leading to their widespread adoption in the field. New TSFM~\cite{goswami2024moment, ansari2024chronos, woo2024unified} are appearing that could enhance visual analytics by giving more interpretable, fast and interactive analysis. Specifically, the MOMENT family~\cite{goswami2024moment} has been analyzed to show interpretable embedding projections for different synthetic datasets (see~\customref{fig:momentfm:projections}) and to have good performance against the state-of-the-art~\cite{goswami2024moment}.

In this context, two main questions arise for the field of visual analytics: 

\begin{description}[leftmargin=1.5em, labelindent=0em, align=left]
    \RQitem{rq:moment:1} Do quantitative improvements in the classical tasks result in embeddings space interpretability enhancement?
    
    \RQitem{rq:moment:2} How much tuning is required to obtain high quality descriptive clusters for a specific time series?
\end{description}

We explore these questions by conducting both statistical and visual analyses using DeepVATS~\cite{rodriguezfernandez2023deepvats} as the evaluation framework. Initially developed as a Deep Learning-based tool, DeepVATS is a Visual Analytics tool for time series that aimed to analyze the embedding space of long time series (TS) by leveraging DL models to extract vectorized representations, providing a more intuitive comprehension of their structure. 

However, the high computational cost of this approach limits interactivity as dataset size increases. To enhance the interactiveness - and with an eye in the Data Mining field - MPlot (similarity matrix plot) was integrated into DeepVATS~\cite{Santamaria-Valenzuela2024, santamariavalenzuela2024deepvats, shahcheraghi2024introducing}. Although computationally expensive, these methods have been successfully scaled for large-scale TS exploration. The integration of MPlot into DeepVATS has significantly enhanced trend detection capabilities for univariate time series while also providing a fast preview of the series’ properties~\cite{Santamaria-Valenzuela2024}. This combination facilitates an initial exploration of univariate TS while the Deep Learning model is still being trained, improving the overall workflow efficiency. 

While this integrations supports fast previews and initial exploration, the question remains wether taking a step back to the DL field, TSFM embeddings can offer deeper insights - not just a preview - of both univariate and multivariate time series properties. To analyze this apportation, we integrate MOMENT into DeepVATS and analyze its embedding space using the same datasets and projection techniques as in Rodriguez-Fernandez et al.~\cite{rodriguezfernandez2023deepvats} to ensure a fair comparison.

The findings suggest that quantitative improvements are not necessarily linked to the embeddings precision, and that even minimal fine-tuning can significantly change a model's internal knowledge. This indicates that extensive training may not be required for effective adaptation, being a great idea to integrate foundation models into visual analytics tools. In the next steps, we aim to refine this fine-tune aproach to produce clearer embeddings that better capture the diverse properties of time series data for specific tasks. 

Thus, the central contribution of this work is the introduction of the MOMENT mechanism into DeepVATS resulting - to the best of or knowledge - on the first integration of foundation models into visual analytics tools. This integration has been done enabling the adaptation and fine-tuning of any pretrained model in a zero-shot mode. This approach offers a flexible and efficient way for visually analyzing large time series data.

The next sections are structured as follows. Section~\ref{sec:foundations} introduces foundation models for time series and, more specifically, the MOMENT family. Section~\ref{sec:integration} shows the analysis of four synthetic datasets and one real-data dataset within deepVATS using MOMENT models and analyzing it's advantages and disadvantages. Section~\ref{sec:conclusions} summarizes the analysis and proposes future lines for improving visual interactive analytic tools for time series.

\section{Background \label{sec:foundations}}
\begin{figure}[!htb]
    \centering
    \includegraphics[width = 1\linewidth, pagebox=artbox]{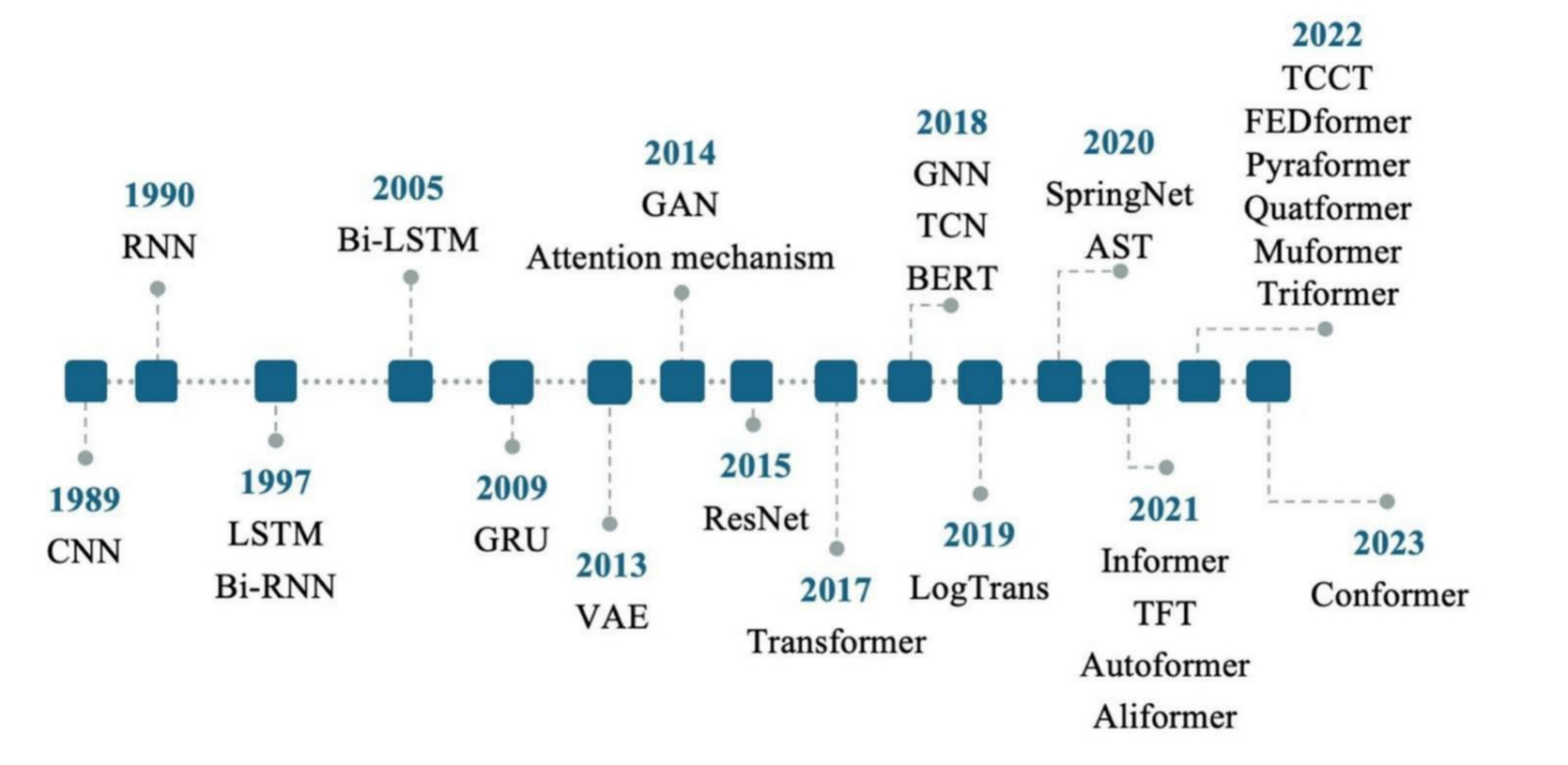}
    \caption{Historical development of time series forecasting deep learning. Figure obtained from~\cite{su2025systematic}.}
    \label{fig:transformers_history}
\end{figure}

The past two years have been especially important for the development of time series models. At the beginning, Transformers were thought to lose temporal correlation, making doubts about their use for time series analysis~\cite{zeng2023transformers}. Few time later Wu et al.~\cite{wu2021autoformer} added a new AutoCorrelation mechanism that fixed the temporal correlation problem, making transformers adaptable to time series. Figure~\ref{fig:transformers_history} shows how, only in 2023, different transformer models~\cite{vaswani2017attention} were adapted to time series and their history still continues in the present year (see timexer~\cite{wang2025timexer}, which adds exogeneous variates, or Medformer~\cite{wang2025medformer}, developed for specific medical classification tasks). Their flexibility made them easy to adapt for pre-trained models and foundation models, showing great results in other fields like image~\cite{dhekane2025transfer, awais2025foundation} or text~\cite{rupaadvancing,yang2025recent} leads to the option of using foundation models for time series. 

At first, transfer learning was thought not to be appropriate for TS as ~\cite{woo2023pushing} describe: ``Unlike image and text data which naturally share semantic information across datasets and domains, time series data may not enjoy such properties of transferability as the semantics of time series data may be unique to their dataset or domain. As such, it is still unclear how time series models can benefit from pretraining and transfer learning''. However, in the same year, transfer learning for specific tasks appeared, motivating practitioners to no longer train models from scratch when classifying time series but instead fine-tune pre-trained models~\cite{fawaz2018transfer,zuo2025mixed,meyer2025prospective}. Eventually, following the positive path,  the time series foundation models appeared, increasingly gaining importance in the field~\cite{dai2025data, garza2023timegpt, goswami2024moment,woo2024unified}. 

\subsection{Most influential time series foundation models}

To the best of our knowledge, the most influential TSFMs, based on the transformer architecture, are the following five: 
\begin{description}
    \item[Chronos] An univariate time series forecasting model based on language architectures~\cite{ansari2024chronos}.
    \item[MOIRAI] A Masked Encoder-based Universal time series Forecasting Transformer, Trained on a Large-scale Open time series Archive (LOTSA) featuring over 27B observations across nine domains (energy, climate, cloud operations, transport, web, sales, finance, nature, and healthcare)~\cite{woo2024unified}. 
    \item[Lag-LLaMa] A decoder-only transformer pre-trained with lag features, leveraging LLaMA to univariate probabilistic time series forecasting~\cite{rasul2023lag}.
    \item[TimeGPT-1] A proprietary time-series foundation model designed for generative forecasting across diverse domains~\cite{garza2023timegpt}.
    \item[MOMENT] A multivariate time series foundation model, designed for multitask forecasting across different datasets~\cite{goswami2024moment}.
\end{description}
\subsection{Visual Analytics for Time Series}

Embedding space visualization is an important tool used to interpret how DL models internally represent high-dimensional data. These projections - typically computed using techniques like t-SNE~\cite{van2008visualizing}, UMAP~\cite{mcinnes2018umap} or the more recent PacMAP~\cite{pacmap} - allow analysts to explore the structure of the data, detect outliers, and evaluate how well a model captures the intrinsecal characteristics. This is used across diferent domains such as genomics~\cite{bhatnagar2024combining} or healthcare-related time series~\cite{stym2025dafted}. These areas usually represent data in high-dimensional tabular matrices, and visualizing their embeddings enables researches to observe patterns, disease progressions, or treatment effects. 

Tools like TimeClustser~\cite{ali2019timecluster} project embeddings extracted from sliding-windowed time series into scatter plots to identify intrinsecal behaviours, while DeepVATS~\cite{rodriguezfernandez2023deepvats} combines masked autoencoders and data mining tools for better capturing patterns. In addition, it integrates an interactive interface that allows users to link points in embedding space regions with raw temporal behaviour, enhancing the interpretability within the embedding space analysis.

Focusing on temporal tabular data, in the healtcare domain, Stym-Popper et al.~\cite{stym2025dafted} analyze tabular information that could be summarized as time series, using a DL model embedding space projection to support hypertension diagnosis. Adding interactive to such visual analyses could provide additional insigths into patient evolution. In the sports and transport domain, Adrienko et al.~\cite{andrienko2024human} use visual embeddings projections to 
expose groups of time windows with similar feature values, enabling interactive selection of groups for inspection and eventual labeling. However, the tool does not show an interactive relation between the scatter plot and the raw data. Combining their labelling cappacity with the interactiveness of DeepVATS would greatly add interpretability to the results.

\subsection{DeepVATS application}
DeepVATS~\cite{rodriguezfernandez2023deepvats} is a Deep Visual Analytics tool for Time Series that focuses on the modelling of intrinsecal characteristics of the time series through Deep Learning models, showing the results in a visual interactive plot of their latent space and the original plot. The DL backbone model is a TimeInception based masked time series autoencoder (MTSAE) that uses a masked value prediction callback (MVP) from `\texttt{tsai}` package~\cite{tsai, ismail2020inceptiontime}. The application is thought for four main tasks: detection of anomalies, segmentation, pattern detection, and trend detection.  


All the foundation models have an embedding space that should be able to detect those characteristics. In particular, MOMENT has already been trained for multitasks including imputation, anomaly detection, and classification; therefore, it should achieve fair results within those tasks. In fact, Goswami et al.~\cite{goswami2024moment} defends that the embedding space using the PCA and TSNE dimensionality reduction techniques shows that the models fit adequately to the trend of the timeseries, showing interpretable embedding spaces (see~\customref{fig:momentfm:projections}). Thus, the integration of MOMENT into DeepVATS could aid in the achievement of an interactive application with few training times for large time series, the integration integrate both tools to test the interpretability of MOMENT within deepVATS benchmark. 

\subsection{The MOMENT family}

\begin{figure}[!ht]
    \centering
    \includegraphics[width=1\linewidth, pagebox=artbox]{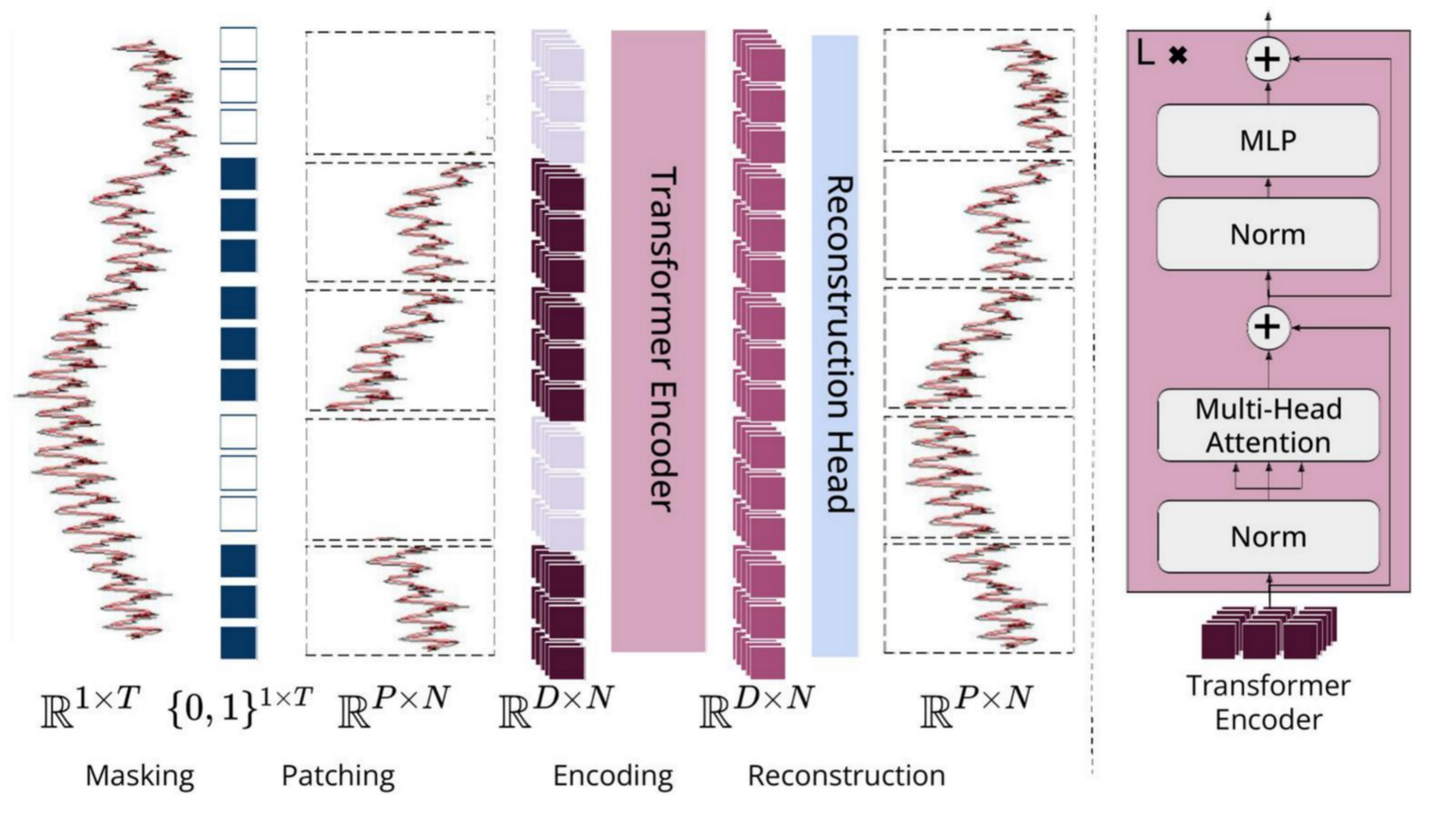}
    \caption{Overview of MOMENT. A time series is broken into $P$ disjoint fixed-length ($N$) patches. Each of them is mapped into a $D$-dimensional patch embedding. During pretraining, patches as  uniformly masked at random by replacing their embeddings. The goal of pre-training is to learn patch embeddings which can be used to reconstruct the input time series using a light-weight reconstruction head. It is important to advise that, eventhough the image shows the univariate process, the model is defined for multivariate time series. Figure obtained from Fig.~$3$ of ~\cite{goswami2024moment}.  }
    \label{fig:momentfm}
\end{figure}

\begin{table}[!htb]
    \caption{Model sizes and architectures of MOMENT pre-trained variants and the  model.}
    \label{tab:momentfm:models}
    \centering
    \setlength{\tabcolsep}{8pt}
    \resizebox{0.9\linewidth}{!}{

        \begin{tabular}{cccc}   
        \hline
            \textbf{Model} & 
            \textbf{Layers} & 
            \textbf{Hidden Dimension} & 
            \textbf{Parameters ($\approx)$}\\
            \arrayrulecolor{black}\hline
            \noalign{\vskip 2pt}
            MOMENT-1 Small & 8  & 512  &  38 M \\ 
            \arrayrulecolor{black!40}\hline
            MOMENT-1 Base   & 12 & 768  &  113 M\\ 
            \arrayrulecolor{black!40}\hline
            MOMENT-1 Large   & 24 & 1024 &  346 M\\
            \noalign{\vskip 1pt}
            \arrayrulecolor{black}\hline
            \noalign{\vskip 2pt}
            \textbf{DeepVATS-MTSAE}   & \textbf{6} &  \textbf{128} & \textbf{457K} \\ \hline
        \end{tabular}
    }

\end{table}

The MOMENT TSFM family relates to transformer models (see~\customref{fig:momentfm}) with three different size versions pre-trained: small, base, and large (see~\customref{tab:momentfm:models}). Also, each version can be loaded in four different modes (tasks): reconstruction, embedding, forecasting, and classification. Each task modifies the structure of the transformer (see~\customref{fig:momentfm}) to use a specific output head based on the task to be solved, without modifying the base model. 

Thus, MOMENT learns general representations of the time series and adapts to different tasks by adding a specific head for each task: 
\begin{itemize}
    \item \textbf{Reconstruction task.} Used for imputation and anomaly detection. This version is trained to reconstruct missing values within masked patches of the time series. 
    \item \textbf{Embedding task.} Used for representation learning, clustering, and classification. In this case, MOMENT generates a fixed embedding vector for each time series, capturing its essential features.
    \item \textbf{Forecasting task.} Used to predict future values in the time series. MOMENT learns representations and maps them to a forecast horizon. 
    \item \textbf{Classification task.} Used for labeling the time series based on learned patterns. MOMENT extracts representations and maps them to output classes.
\end{itemize}

Since the embedding task generates general latent representations of the time series, it is the most adequate for its integration into DeepVATS as it is not limited to a specific task and models the intrinsecal characteristics of the time series. From now on the notation ``MOMENT-small'', ``MOMENT-base'' and ``MOMENT-large'' is used to refer to the different sizes of the embedding task version of the model. 

\section{Integration of MOMENT into DeepVATS}
\label{sec:integration}

In order to analyze the integration of MOMENT into DeepVATS in an equitable manner, we analyze and visualize latent spaces for most of the data sets analyzed in Rodriguez-Fernandez et a.~\cite{rodriguezfernandez2023deepvats} and Santamaria-Valenzuela et al. ~\cite{santamariavalenzuela2024deepvats}: synthetic datasets generated by additive decomposition \texttt{S1}, \texttt{S2}, \texttt{S3}, the synthetic multivariate series \texttt{M-Toy}~\cite{law2019stumpy}, and the \texttt{Kohl's} dataset, based on real data~\cite{matsubara2015web}. Dataset \texttt{S1} is used to analyze segmentation, \texttt{S2} for anomaly detection, \texttt{S3} for trends, \texttt{M-toy} for anomaly detection, and \texttt{Kohl's} for both segmentation and trend detection\footnote{The code necessary for replication as well as the full-sized screenshots are available at \href{https://github.com/misantamaria/deepvats-foundation/tree/paper}{github:misantamaria/deepvats-foundation}.}.

The goal is to check that MOMENT achieves at least the capabilities of MTSAE. This architecture has proven to have interpretable latent space projections for anomalies, patterns, and segments, lacking that interpretability in detecting trends. The addition of MOMENT to DeepVATS would provide more interactive and interpretable latent space plots, solving the question ~\ref{rq:moment:1}.

\subsection{zero-shot analysis of S1 using MOMENT-small}
\begin{figure}[!ht]
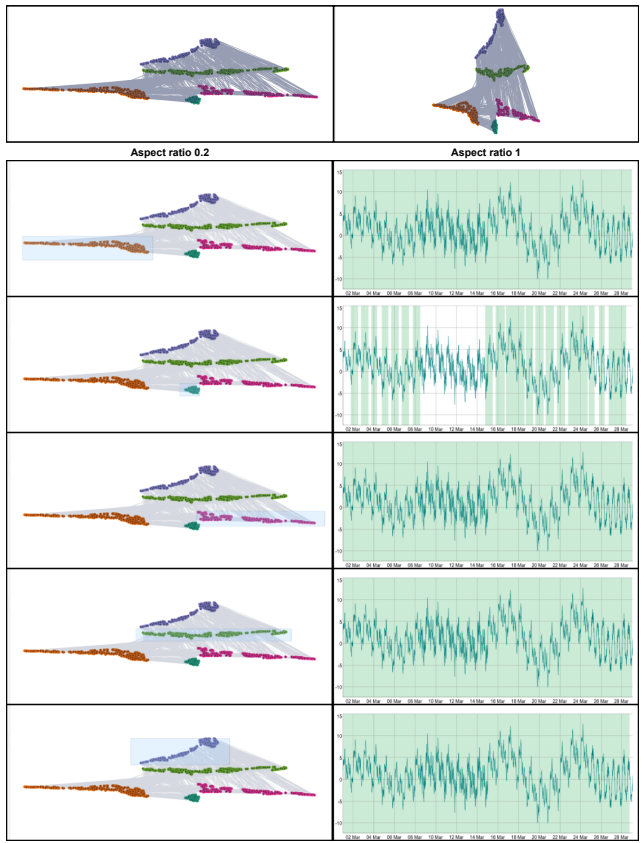

    \centering
    \includegraphics[width=1\linewidth, pagebox=artbox]{pic/moment/small/S1-zeroshot-global.pdf}
    \includegraphics[width=1\linewidth, pagebox=artbox]{pic/moment/small/S1-zeroshot-global-cluster.pdf}
    \caption{Execution of MOMENT-small for \texttt{S1} taking the mean in batches of $20$ minutes. The execution is done for window length $54$ and stride $2$. 
    } 
    \label{fig:s1:moment:zero-shot:global}
\end{figure}

The \texttt{S1} time series is designed to have four segments ($df_{\ast}$), giving a good example to test the segmentation task within the application. The execution of MOMENT for \texttt{S1} in the same configuration as in the original research paper results in five very interlaced clusters, like a tangle. Four of them cover the whole series and the other one seems to be getting the whole series but not the $df_2$ segment (see ~\customref{fig:s1:moment:zero-shot:global}).  Using a zoom in all the clusters  (see ~\customrefs{fig:s1:moment:zeroshot:cluster_1},~\ref{fig:s1:moment:zeroshot:cluster_2},~\ref{fig:s1:moment:zeroshot:cluster_3}, ~\ref{fig:s1:moment:zeroshot:cluster_4},~\ref{fig:s1:moment:zeroshot:cluster_5}) we can detect that MOMENT is trying to get patterns, but they do not seem to be very clear. Also, the only segment that MOMENT can detect correctly is $df_2$. 

This fact leads to two questions. The first one is: ``Is the size of the window hiding important information from MOMENT as it tries to learn more specific information?''. And second, ``Can I force MOMENT to learn specific information of the time series with few shots of the data?''. Based on the fact that foundation models work better when fine tuning is applied, we focus on the second one, analyzing the amount of dataset MOMENT needs to know to get better insights of it in the inference step. 

\subsection{Fine-tune in DeepVATS}

For fine tuning, a wrapper has been built so that any torch-based model can easily be included in deepVATS's fine-tuning process. The fine-tuning is done like an imputation task training, in a similar way to  training, thought for the imputation task: based on the percentage of the training dataset and masking part, use and validate the model for the reconstruction task using both a validation and train dataset. 

The goal is to check if this reinforcement makes MOMENT fit the time series better and to check if the embeddings space plot shows this enhancement. The data will be selected from parts of the time series to make sure we don't focus on a specific/special part of the time series.

The fine-tuning algorithm is based on four main parameters:
\begin{itemize}
    \item \texttt{window\_lengths}. Hosts the window sizes that will be used for the batches in the fine-tuning.
    \item \texttt{training\_percent}. This value is the percentage of the total percentage of the dataset that will be used to get the training batches. 
    \item \texttt{valid\_percent}. This value is the percentage of the total percentage of the dataset that will be used to obtain the validation batches.
    \item \texttt{mask\_percent}. This value is the percentage that will be masked in each batch. 
\end{itemize}

The selection of the window lengths and the position of the selected windows within the time series resulted in two different versions of the fine-tuning algorithms, modelled in the code based on a \texttt{mix\_windows} flag:

\begin{itemize}
    \item \textbf{\texttt{mix\_windows} False.} This version loops through \texttt{window\_lengths}, splitting the total dataset for the curreng \texttt{wlen} size and randomly selecting (within the \texttt{training\_percent} and \texttt{valid\_percent} percentages) windows across the time series. 
    \item \textbf{\texttt{mix\_windows} True.}  This version splits the time series using the validation and training percentages. The validation and training datasets are built from the beginning of the time series. The first part will be the training dataset and the second part, the validation dataset. The batches are dinamically built getting a random window length from \texttt{window\_length} in each batch. To ensure the integrity of the training and validation batches, the values of the lengths are cached so each validation is done within the same batches. 
\end{itemize}

The first version did not show much improvement. It seemed to be skewed, as each window size was retraining the full model just after the previous one, maybe leading to catastrophic forgetting of the previous model at each iteration. Thus, we tested mixing the window sizes to get a more stochastic process, one that ensured a global learning of the time series for the model. 

This rationale led to the second version, where we got improvements of more than $20\%$ loss decrease. However, this second version has a difficulty in selecting random parts of the time series to make sure the model learns the global shape of it. The problem is that we cannot directly divide the time series correctly in different window sizes. Maybe we could fix it by taking the largest window for the division and later reducing the batch window size by getting the first or last values so we can fuse both fine-tune modes.

\subsection{Statistical Loss improvement analysis} 

To select the values for the main parameters of the fine-tune of the MOMENT models, a previous analysis using the Kohl's time series is done. For each version of the MOMENT embedding model (small, base, and large), the validation percentage is fixed to $0.3$ and four parameters are analyzed. 

The first parameter is the number of epochs for the training. The model is trained until $20$ epochs, saving model with smaller loss in the training step as as the final model. Once the training is complete, we fix the best epoch as the one where the optimal model was saved. This approach reduces the execution time in the visualization by selecting the most suitable number of epochs.  

The second parameter is the \texttt{dataset\ percentage}, which is tested for $\{0.15,\,0.2,\,0.25,\,0.3\}$. This percentage is the part of the dataset used as shot - training dataset -  for the fine-tune of the different model. 

The third parameter is the number of window lengths. The first window length is fixed to $17$ as in the previous MTSAE experiment at Rodriguez-Fernandez et al.~\cite{rodriguezfernandez2023deepvats}. The remaining window lengths are selected using the Fourier Transform result, which identifies the most relevant frequencies, as window sizes until we reach the total number of sizes desired (through the \texttt{find\_dominant\_window\_sizes} function in \texttt{aeon.segmentation})~\cite{middlehurst2024aeon}. The experiment considers the following number of window sizes: $\{1,\,2,\,4,\,6,\,8,\,10\}$.

The combination of these options (a total of $72$ cases) leads to a sufficiently complete scenario for a preliminary selection of the fine-tune parameters for the visual analysis based on the loss improvement. However, due to computational costs, the scenario is reduced, setting the number of epochs to $20$, dataset percentages to $\{0.15,\,0.2,\,0.3\}$, masked percentages at $\{0.25,\,0.5,\,0.75\}$, and window lengths at $\{1,\,5\}$, resulting in $18$ different use cases.

This setup allows us to compare MOMENT not only against itself but also against MTSAE. However, this comparison is challenging and not entirely fair, as we are evaluating against a pre-trained MTSAE model (replicating the one used in~\cite{rodriguezfernandez2023deepvats}, trained with window sizes from $12$ to $17$). 



The results of the experiments show really similar losses between the different versions of MOMENT, with much more adaptability for MOMENT-small, followed by MOMENT-base, and next, MOMENT-large (See ~\customrefs{fig:moment:losses:small},~\ref{fig:moment:losses:base},~\ref{fig:moment:losses:large}). 

The expectation is that MOMENT will get good interpretable visual embeddings for the same cases as MTSAE, being better as the number of parameters increases - or the loss is reduced. The rest of the statistical analysis allows us to select the best versions and see if any of them are preferable to the others.


To ensure the best comparison metric, we use the original loss function, taking into account the masked part within the computation. For each execution, the model was validated before and after training, obtaining $loss_{first}$ from the previous evaluation and $loss_{final}$ from the post-evaluation (using the best version of the model). Thus, the loss improvement is defined as $\frac{(loss_{first} - loss_{final}) \cdot 100}{loss_{first}}$. The greater the improvement, the larger the reduction in loss. Therefore, this improvement will be used as the optimization objective in analyzing the best combination of parameters.  

The experiment achieves improvements up to $22\%$ in the small model, up to $10\%$ in the base model, and up to $4\%$ in the large model (see~\customrefs{fig:improvements:small},~\ref{fig:improvements:base},~\ref{fig:improvements:large}). This reduction is expected since each larger model contains significantly more parameters than the previous one, allowing it to fit the original time series more effectively. Consequently, achieving further improvement becomes increasingly difficult.  

\begin{table}[!th]
    \centering
    \renewcommand{\arraystretch}{1.2}
    \setlength{\tabcolsep}{8pt}
    \caption{Feature importance scores from SelectKBest, Random Forest, SHAP impact and linear correlation for the MOMENT-small model.}
    \resizebox{1\linewidth}{!}{ 
    \begin{tabular}{ccccc}
    \hline
        \textbf{Feature} & 
        \textbf{R. Forest \%} & 
        \textbf{KBest \%} & 
        \textbf{SHAP \%} & 
        \textbf{Corr. \%} \\
        \hline
        \noalign{\vskip 2pt}
        masked\_percent  & 
        \cellcolor{green!80} 71.47  & 
        \cellcolor{green!75} 69.53  & 
        \cellcolor{green!50} 21.09  & 
        \cellcolor{red!80} -79.21 \\
        \arrayrulecolor{black!40}\hline
        
        best\_epoch  & 
        \cellcolor{green!40} 21.34  & 
        \cellcolor{green!45} 28.89  & 
        \cellcolor{green!30} 10.51  & 
        \cellcolor{green!80} 64.16 \\ 
        \arrayrulecolor{black!40}\hline
        dataset\_percent  & 
        \cellcolor{yellow!40} 7.19  & 
        \cellcolor{yellow!20} 1.58  & 
        \cellcolor{yellow!35} 5.58  & 
        \cellcolor{orange!60} -19.17 \\ 
        \arrayrulecolor{black!40}\hline
        n\_windows  & 
        \cellcolor{yellow!10} 0.00  & 
        \cellcolor{yellow!5} 0.00  & 
        \cellcolor{yellow!7} 0.02  & 
        \cellcolor{gray!10} 0.07 \\ 
        \arrayrulecolor{black}\hline
        
    \end{tabular}
    } 
    
    \label{tab:features_importance:small}
\end{table}

\begin{table}[!th]
    \centering
    \renewcommand{\arraystretch}{1.2}
    \setlength{\tabcolsep}{8pt}
    \caption{Feature importance scores from SelectKBest, Random Forest, SHAP impact and linear correlation for the MOMENT-base model.}
    \resizebox{1\linewidth}{!}{ 
    \begin{tabular}{ccccc}
    \hline
        \textbf{Feature} & 
        \textbf{R. Forest \%} & 
        \textbf{KBest \%} & 
        \textbf{SHAP \%} & 
        \textbf{Corr. \%} \\
        \hline
        \noalign{\vskip 2pt}
        masked\_percent  & 
        \cellcolor{green!70} 51.22  & 
        \cellcolor{green!75} 63.81  & 
        \cellcolor{green!50} 18.64  & 
        \cellcolor{red!80} -76.35 \\ 
        \arrayrulecolor{black!40}\hline
        best\_epoch  & 
        \cellcolor{green!60} 48.21  & 
        \cellcolor{green!65} 31.92  & 
        \cellcolor{green!40} 16.38  & 
        \cellcolor{green!80} 64.12 \\ 
        \arrayrulecolor{black!40}\hline
        dataset\_percent  & 
        \cellcolor{yellow!40} 0.53  & 
        \cellcolor{yellow!20} 4.27  & 
        \cellcolor{yellow!35} 0.67  & 
        \cellcolor{green!50} 29.24 \\
        \arrayrulecolor{black!40}\hline
        n\_windows  & 
        \cellcolor{yellow!10} 0.03  & 
        \cellcolor{yellow!5} 0.00  & 
        \cellcolor{yellow!7} 0.1  & 
        \cellcolor{gray!10} 0.04 \\
        \arrayrulecolor{black}\hline
    
    \end{tabular}
    }
    
    \label{tab:features_importance:base}
\end{table}

\begin{table}[!th]
    \centering
    \renewcommand{\arraystretch}{1.2}
    \setlength{\tabcolsep}{8pt}
    \caption{Feature importance scores from SelectKBest, Random Forest and linear correlation for the MOMENT-large model.}
    \resizebox{1\linewidth}{!}{ 
    \begin{tabular}{ccccc}
    \hline
        \textbf{Feature} & 
        \textbf{R. Forest \%} & 
        \textbf{Best \%} & 
        \textbf{SHAP \%} & 
        \textbf{Corr. \%} \\
        \hline
        \noalign{\vskip 2pt}
        masked\_percent  & 
        \cellcolor{green!85} 56.77  & 
        \cellcolor{green!90} 80.19  & 
        \cellcolor{green!70} 18.11  & 
        \cellcolor{green!95} 97.33 \\
        \arrayrulecolor{black!40}\hline
        best\_epoch  & 
        \cellcolor{green!70} 41.52  & 
        \cellcolor{green!75} 19.76  & 
        \cellcolor{green!50} 12.51  & 
        \cellcolor{green!85} 90.32 \\
        \arrayrulecolor{black!40}\hline
        dataset\_percent  & 
        \cellcolor{yellow!30} 1.39  & 
        \cellcolor{yellow!15} 0.05  & 
        \cellcolor{yellow!20} 1.21  & 
        \cellcolor{red!60} -10.56 \\
        \arrayrulecolor{black!40}\hline
        n\_windows  & 
        \cellcolor{yellow!10} 0.33  & 
        \cellcolor{yellow!5} 0.001  & 
        \cellcolor{yellow!7} 0.2  & 
        \cellcolor{red!50} -1.82 \\ \arrayrulecolor{black}\hline
    \end{tabular}
    }
    
    \label{tab:features_importance:large}
\end{table}

To check the parameters that are more related to the improvement in loss, correlations are checked. First, a linear correlation matrix is displayed to check the relationship between the analyzed values: $time$, $best\_epoch$, $dataset\_percent$, $masked\_percent$, $n\_windows$ and $improvemnt$. Our expectation was to have a high positive correlation between the percentage of the dataset and the improvement, as well as the masked dataset and the improvement. However, the only constant correlation found is the time with the percentage of the dataset used for the training (see~\customrefs{tab:features_importance:small},~\ref{tab:features_importance:base},~\ref{tab:features_importance:large}~\ref{fig:correlations:small},~\ref{fig:correlations:base},\ref{fig:correlations:large}). Thus, we continued to apply other techniques to check non-linear influences. 

    
    
    
%





First, we used different DL-based algorithms to detect the most relevant features and how much was related to the loss improvement (to increase or decrease). Second, we looked for the best parameters within the two most relevant parameters. Finally, for the best values of these parameters, the best combination of the other parameters is selected.

For the first step, we use  \texttt{sklearn.feature\_selection}'s \texttt{SelectKBest} and \texttt{f\_regression} functions and \texttt{sklearn.ensemble}'s \texttt{RandomForestRegressor}. Also, the SHAP impact percent is used as another correlation metric that gives an idea of the ``direction'' of the influence (the larger the best or the smaller the best). The result is constant among the versions of MOMENT: \texttt{masked\_percent} $>$ \texttt{best\_epoch} $>>$ \texttt{dataset\_percent} $>>$ \texttt{n\_windows}. The \texttt{masked\_percent} has a direct correlation with the loss improvement in the MOMENT-large model while maintaining an inverse correlation in the small and base versions. The best epoch index maintains a direct relation in the three cases (see~\customref[2]{tab:features_importance:small},~\ref{tab:features_importance:base},~\ref{tab:features_importance:large}). 

Based on these results, the best combination of parameters is computed using the two more relevant parameters best improvement rows, obtaining the best combination as a start (see~\customref{tab:best_parameters}). 

The next step is to compare MOMENT versus MTSAE. The main challenges arise when constructing the validation dataset: MTSAE does not support batches with varying window sizes, and the masked percentage was predefined during training. To address the first issue, we use a window size of $17$ in validation, since it is the only common size (fixed by hand) in all execution cases. For the second issue, we compute the MSE using the full original input and predictions instead of focusing only on the masked part and execute MOMENT evaluation for the best case in the same case, using the same functions and percentages as in MTSAE for the generation of masks, ensuring equivalent experiment.

\begin{figure}[!htb]
    \centering
    \includegraphics[height=125px]{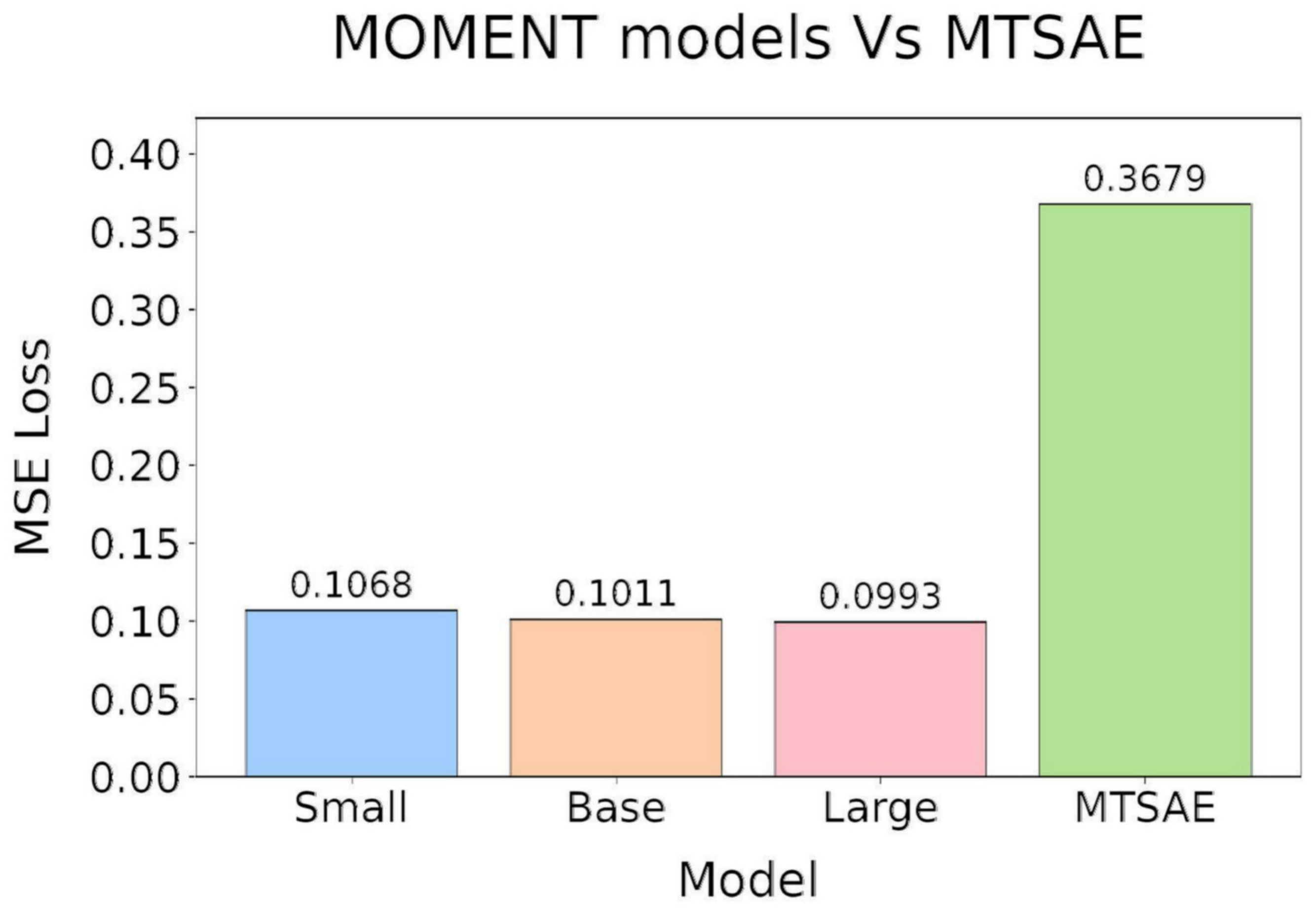}
    \includegraphics[height=125px]{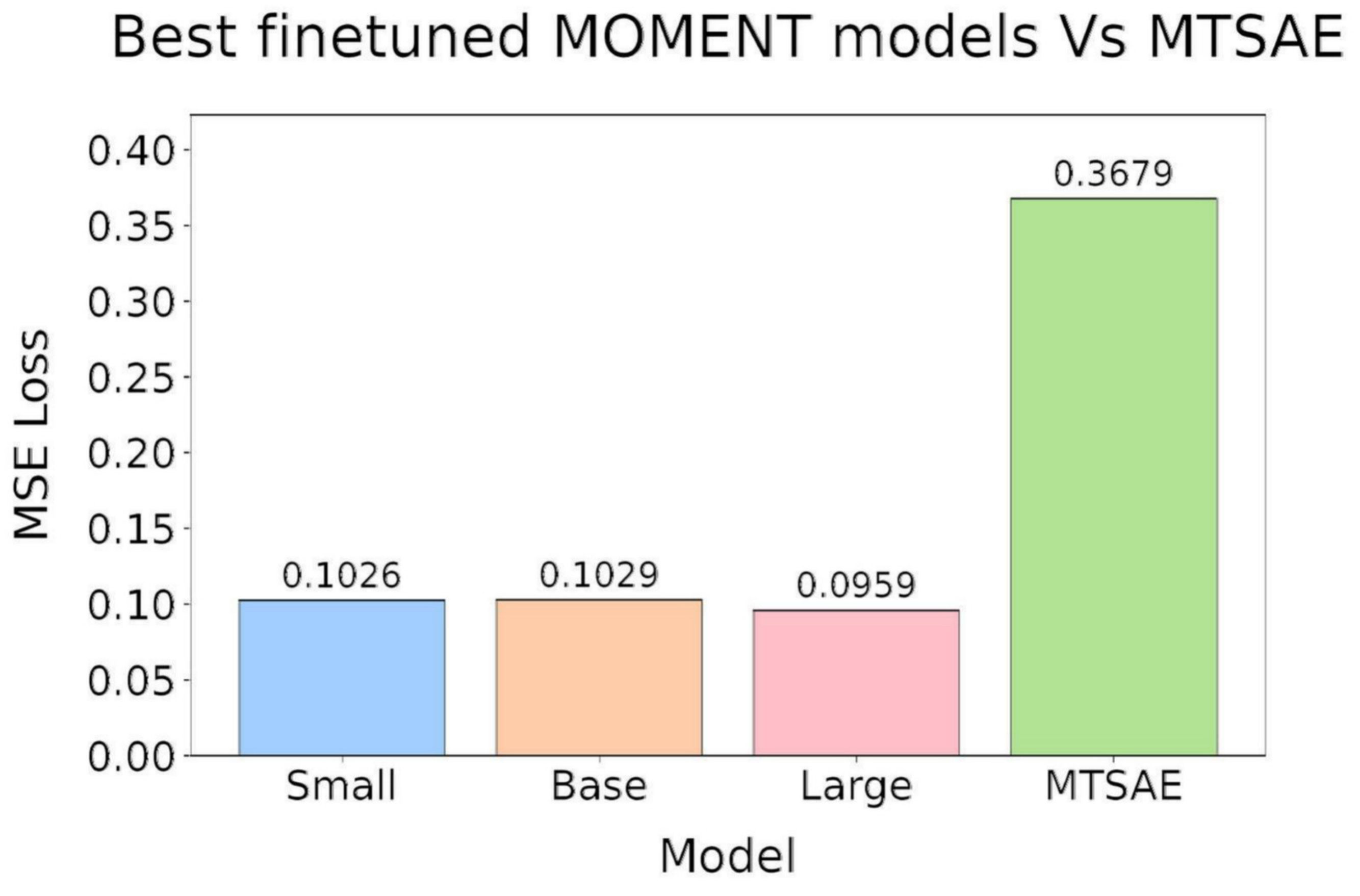}
    \caption{Comparison MOMENT models and the MTSAE model using the mse loss comparing the full prediction to the original batch. At the top, the original version of MOMENT models. At the bottom, a re-training of the best cases. } 
    \label{fig:comparison}
\end{figure}

\begin{figure}[!htb]
    \centering
    \includegraphics[width=0.9\linewidth, pagebox=artbox]{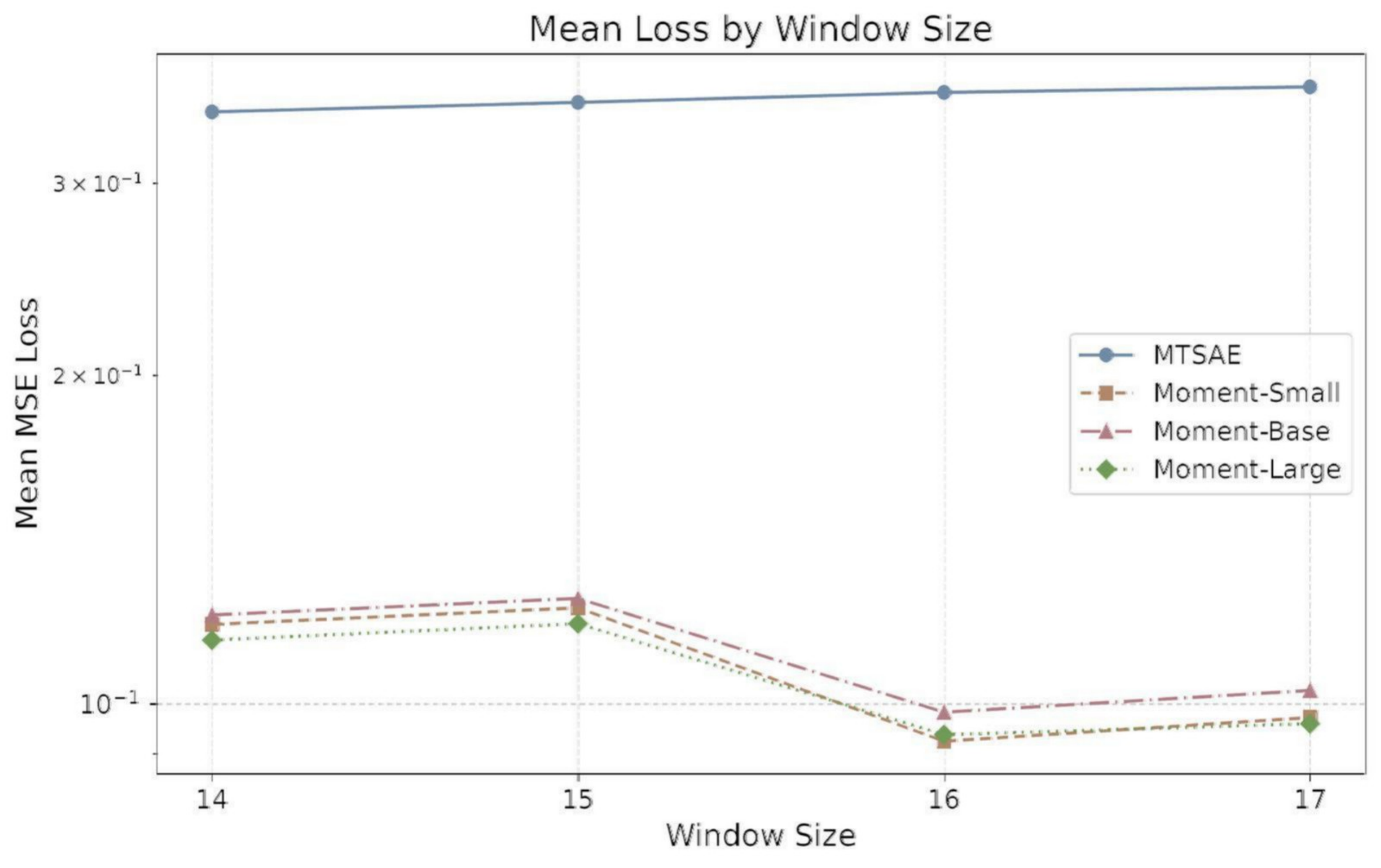}
    \includegraphics[width=0.9\linewidth, pagebox=artbox]{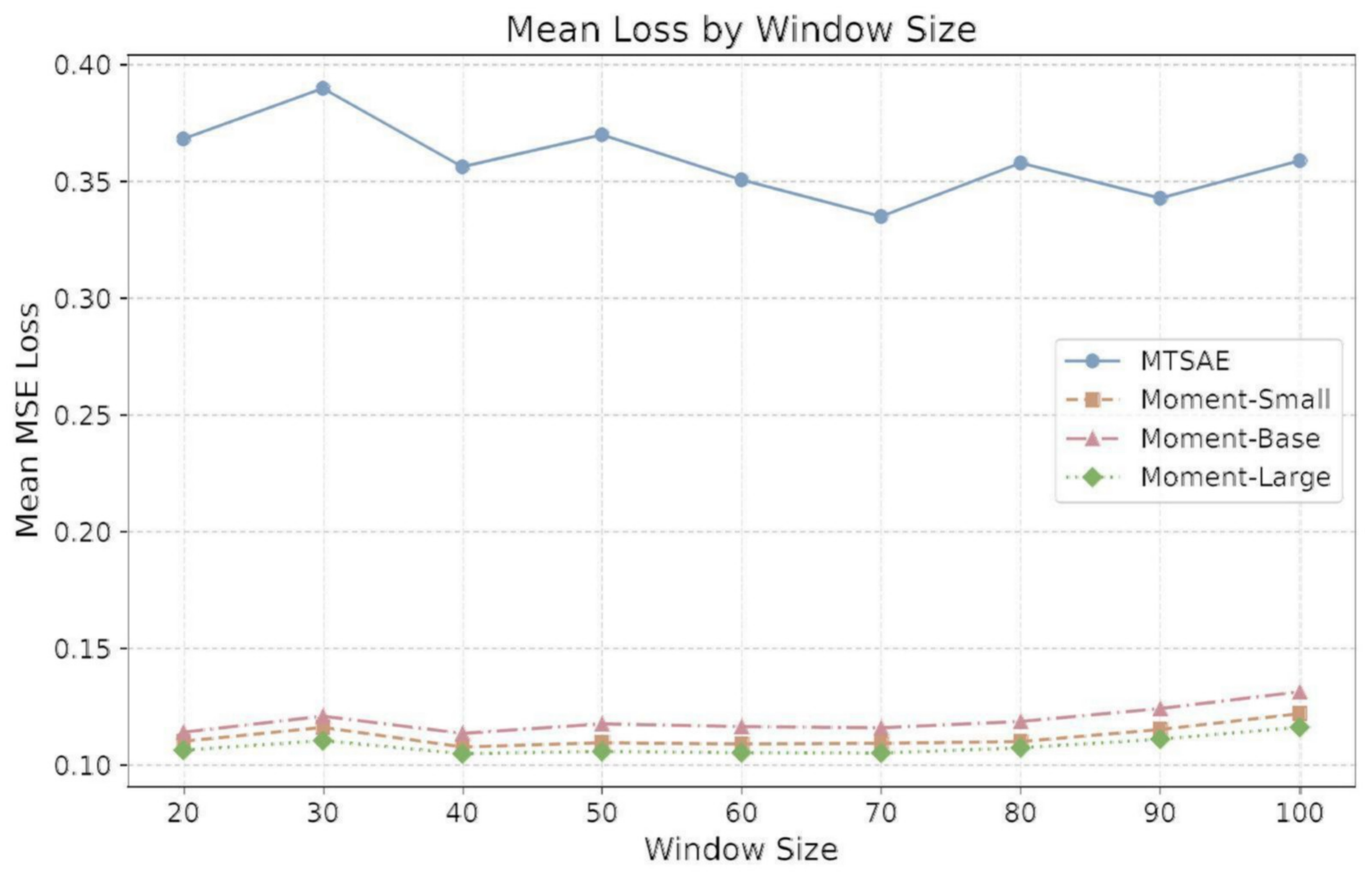}  
    \caption{Comparison MOMENT models in their best version and the MTSAE model using the mse loss comparing the full prediction to the original batch. At the top, using sizes in the range of those used in MTSAE training. At the bottom, usin sizes in the range between $20$ and $100$ in steps of $10$.} 
    \label{fig:comparison2}
\end{figure}

The comparison of the best versions of MOMENT and MTSAE shows how increasing the number of trainable parameters does not necessarily imply an improvement in terms of MSELoss. Also, MTSAE has a loss $70\%$ higher than MOMENT models, regardless of the window length used for the inference (see~\customref{fig:comparison},~\ref{fig:comparison2}). Thus, the expectation is that MOMENT yields really interpretable visual embeddings for the same cases as MTSAE but does not result in too different plots between the three versions of MOMENT (once fine-tuned). The rest of the statistical analysis allows us to select the best versions and see if any of them is preferable to the others.

As the number of epochs may suppose a difference in obtaining the best model in any execution case, its frequencies have been computed to get the largest of the two more repeated values (or the first one if the difference was too big). This analysis resulted in the selection of $17$ epochs for MOMENT-small, $13$ for MOMENT-base and $20$ for MOMENT-large (see~\customrefs{fig:best_epochs:small},~\ref{fig:best_epochs:base},~\ref{fig:best_epochs:large}).

\begin{figure}[!htb]
    \centering
    \includegraphics[width=0.8\linewidth, pagebox=artbox]{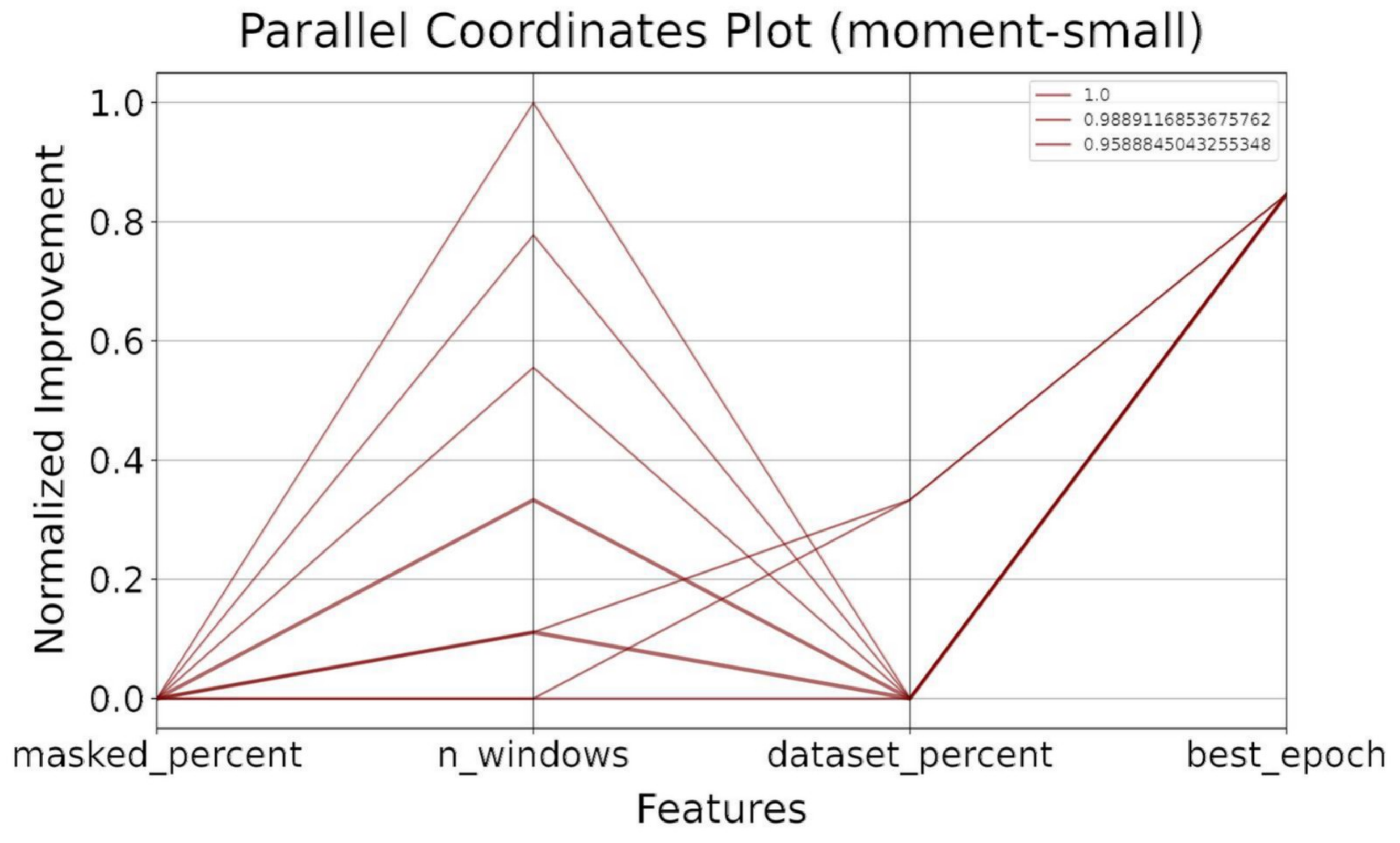}
    \caption{Parallel coordinates plot matrices for MOMENT-small analysis.} 
    \label{fig:parallel_coordinates:small}
\end{figure}

\begin{figure}[!htb]
    \centering
    \includegraphics[width=0.8\linewidth, pagebox=artbox]{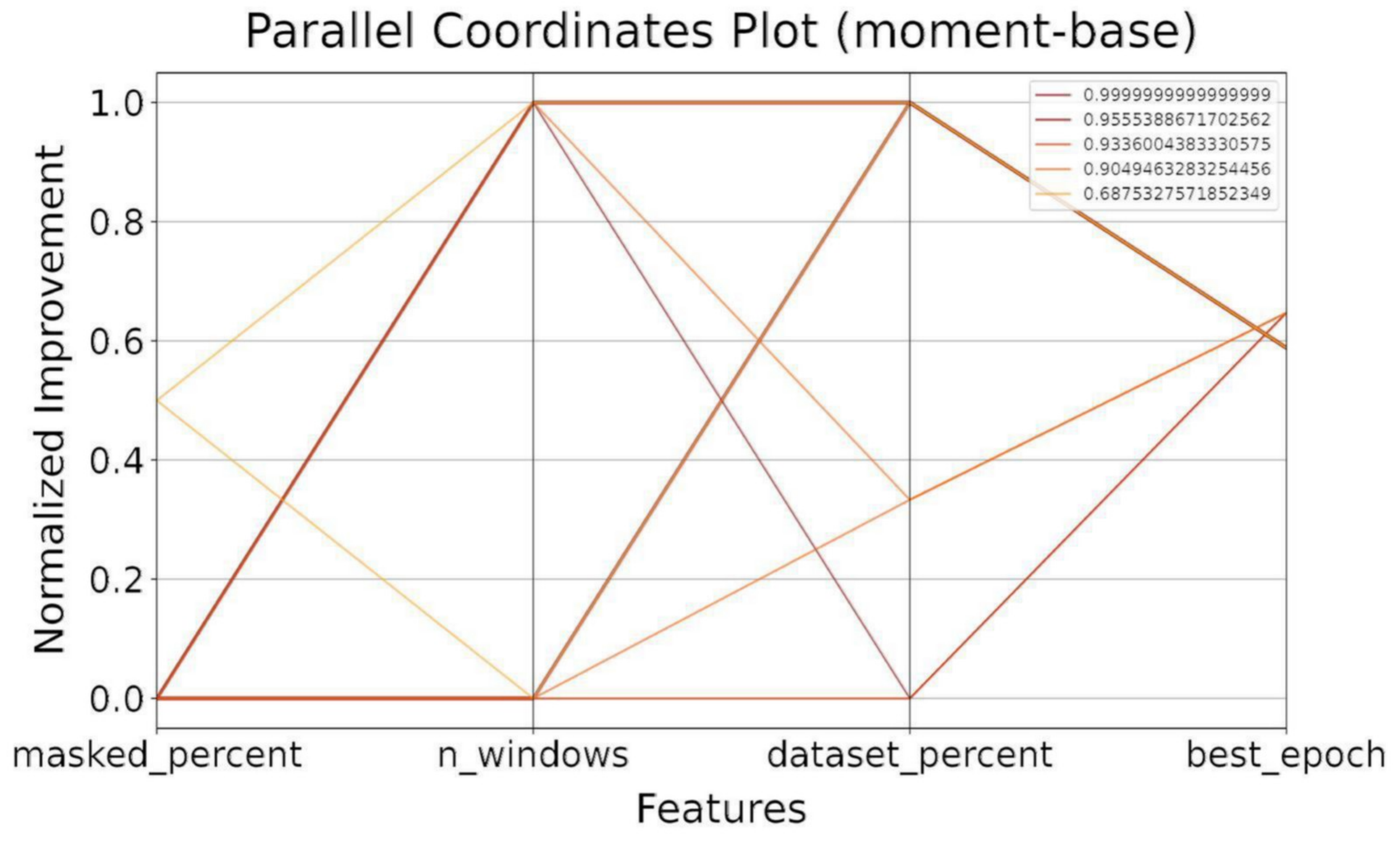}
    \caption{Parallel coordinates plot matrices for MOMENT-base analysis.} 
    \label{fig:parallel_coordinates:base}
\end{figure}

\begin{figure}[!htb]
    \centering
    \includegraphics[width=0.8\linewidth, pagebox=artbox]{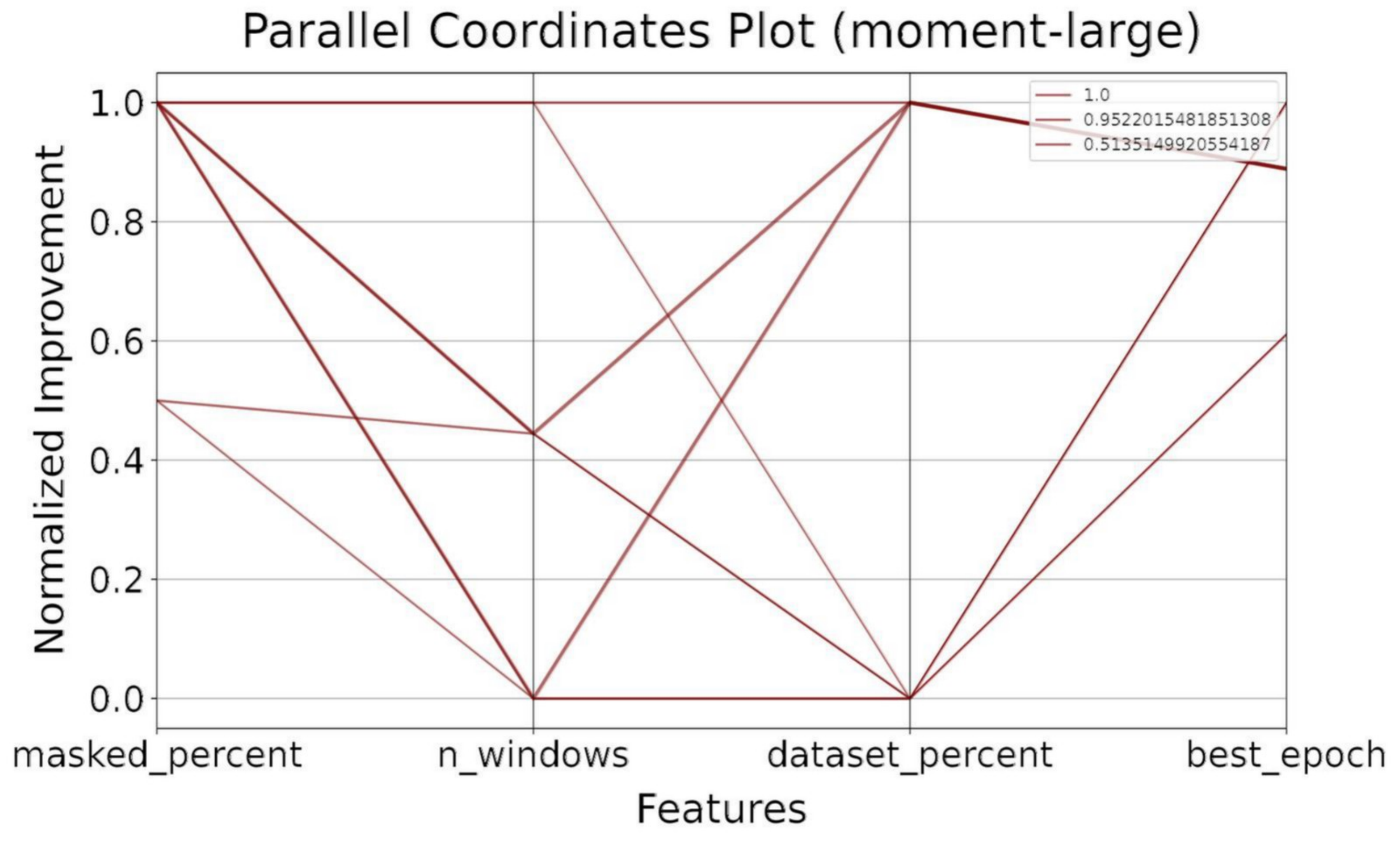}
    \caption{Parallel coordinates plot matrices for MOMENT-large analysis.} 
    \label{fig:parallel_coordinates:large}
\end{figure}

\begin{table}[!htb]
    \centering
    \caption{Best parameter values for Small, Base, and Large models based on feature importance analysis.}
    \label{tab:best_parameters}
    \renewcommand{\arraystretch}{1.3} 
    \setlength{\tabcolsep}{12pt} 
    {
        \begin{tabular}{>{\centering\arraybackslash}c
                        >{\centering\arraybackslash}c
                        >{\centering\arraybackslash}c
                        >{\centering\arraybackslash}c
        }
            \arrayrulecolor{black}\hline
            \textbf{Parameter} & \textbf{Small} & \textbf{Base} & \textbf{Large} \\
            \hline
            \noalign{\vskip 2pt}
            masked\_percent & 25 & 25 & 75 \\ 
            \arrayrulecolor{black!40}\hline
            best\_epoch & 17 & 13 & 17 \\
            \arrayrulecolor{black!40}\hline
            n\_windows & 1 & 5 & 1 \\
            \arrayrulecolor{black!40}\hline
            dataset\_percent & 15 & 15 & 20 \\
            \arrayrulecolor{black}\hline
        \end{tabular}
    }
    
\end{table}

\begin{table}[!htb]
    \centering
    \caption{Final parameter selection for MOMENT-small, MOMENT-base, and MOMENT-large models for the visual experimentation.}
    \label{tab:final_parameters}
    \renewcommand{\arraystretch}{1.3} 
    \setlength{\tabcolsep}{12pt} 
    {
        \begin{tabular}{>{\centering\arraybackslash}c
                        >{\centering\arraybackslash}c
                        >{\centering\arraybackslash}c
                        >{\centering\arraybackslash}c
        }
            \arrayrulecolor{black}\hline
            \textbf{Parameter} & \textbf{Small} & \textbf{Base} & \textbf{Large} \\
            \hline
            \noalign{\vskip 2pt}
            masked\_percent & 25 & 15 & 75 \\ 
            \arrayrulecolor{black!40}\hline
            best\_epoch & 17 & 13 & 10 \\
            \arrayrulecolor{black!40}\hline
            n\_windows & 1 & 5 & 1 \\
            \arrayrulecolor{black!40}\hline
            dataset\_percent & 15 & 25 & 20 \\
            \arrayrulecolor{black}\hline
        \end{tabular}
    }
\end{table}

To confirm that the number of window sizes seems irrelevant, a normalized Parallel Coordinates Plot is computed for the best improvement values. The plots in~\customrefs[2]{fig:parallel_coordinates:small},~\ref{fig:parallel_coordinates:base},~\ref{fig:parallel_coordinates:large} show how, indeed, the value remains irrelevant. This conclusion is given by the sparsity of the best number of windows length value, which does not remain fixed to an option within the best executions. 
Thus, the best option on Tab.~\customref{tab:best_parameters} is selected for each case.

Finally, based on the analysis conducted and looking for a good cost-to-improvement balance, the parameter descripted in \ref{tab:final_parameters} where selected for the visual experimentation.
In all cases, the percentage of the validation dataset was $30\%$. We focused on analyzing the best improvement scenario, but we did not analyze the relation between improvement and execution time - which would also be relevant for interactive applications. 

\subsection{Analysis of S1 using fine-tuned MOMENT-small}

\begin{figure}[!htb]
    \centering
    \includegraphics[width=1\linewidth, pagebox=artbox]{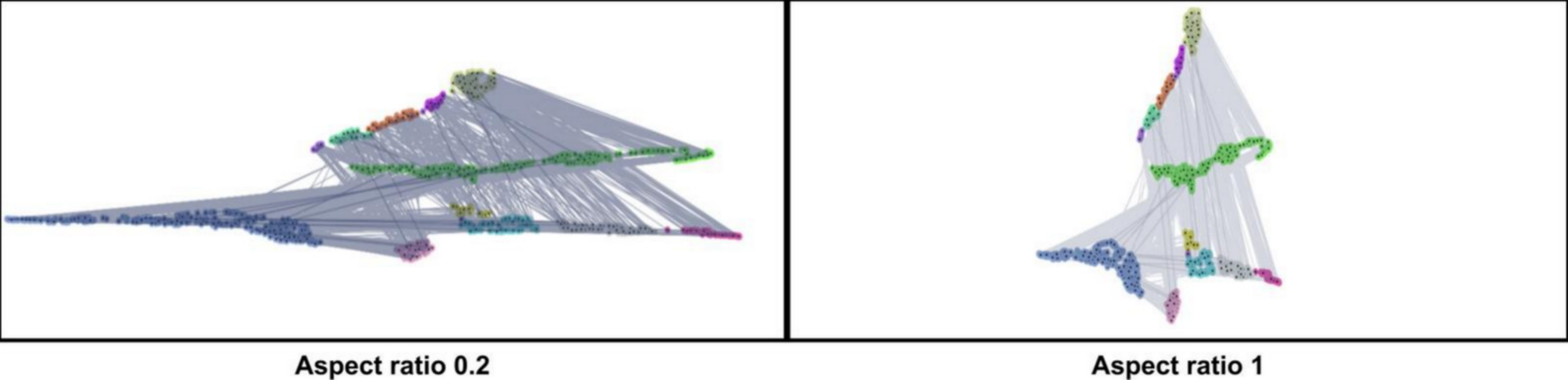}
    \includegraphics[width=1\linewidth, pagebox=artbox]{pic/moment/small/S1-finetuned-global-cluster.pdf}
    \caption{Analysis of the embeddings projection plot of the fine-tuned version of MOMENT-small for \texttt{S1}. } 
    \label{fig:s1:moment:finetune}
\end{figure}

Figure \customref{fig:s1:moment:finetune} shows how MOMENT-small fine-tuning does not represent a visual difference with the zero-shot plot, resulting in the same four clusters shaped as a tree and (somehow) with the only segmentation of the segment $df2$. This is strange as Time Series Foundation Models are supposed to get a really good adaptation to the new time series when fine-tuned. The loss improvement evidences this fact. However, the visual plot of the embedding space does not seem to capture this enhancement. Thus, related to~\ref{rq:moment:1}, this data set does not appear to get a more visually interpretable embedding space when the loss decreases.

To better answer this question, the analysis is performed for the different tasks in DeepVATS.  For segmentation, \texttt{S1} (with the rest of the model versions). For anomaly detection, \texttt{S2} and \texttt{M-Toy}. For trend detection, \texttt{S3}. 

\subsection{Analysis of S1 using MOMENT-base}

\begin{figure}[!htb]
    \centering
    \includegraphics[width=1\linewidth, pagebox=artbox]{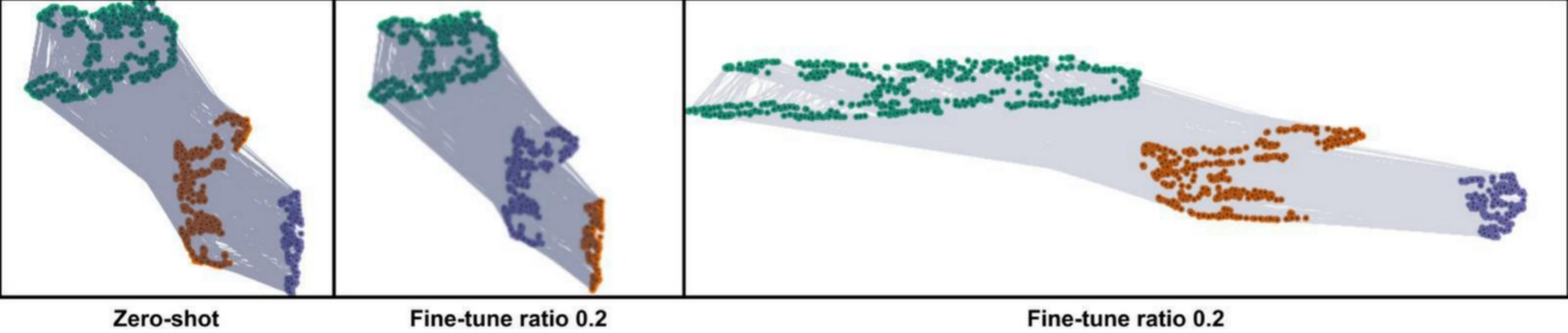}
    \caption{Embeddings proyections plot of the zero-shot and fine-tuned versions of the MOMENT-base model for \texttt{S1}. } 
    \label{fig:s1:moment:base:global}
\end{figure}

The plot of the MOMENT-base-zero-shot and the MOMENT-base-fine-tuned models are practically the same (see Fig. \customref{fig:s1:moment:base:global}), showing no visual enhancement within the projection plot.

 The embedding space is divided into three clusters. The first one seems to detect small patterns, not showing a clear classification of the time series' segments (see~\customref{fig:s1:moment:base:cluster_1}). The second cluster makes a really clear segmentation between $df2$ and the rest of the time series and starts to detect $df4$. However, it is not still as clear as it should be, and the split is not done within the $4$ segments, so the model and embedding projection do not capture the full information we need (see~\customref{fig:s1:moment:base:cluster_2}). However, the third cluster starts to detect four segments really well, but mixed with other patterns (see~\customref{fig:s1:moment:base:cluster_3}). 

At this point, it seems that the base version of the model is capturing much better the segmentation information of \texttt{S1}.

\subsection{Analysis of S1 using MOMENT-large}
\begin{figure}[!htb]
    \centering
    \includegraphics[width=1\linewidth, pagebox=artbox]{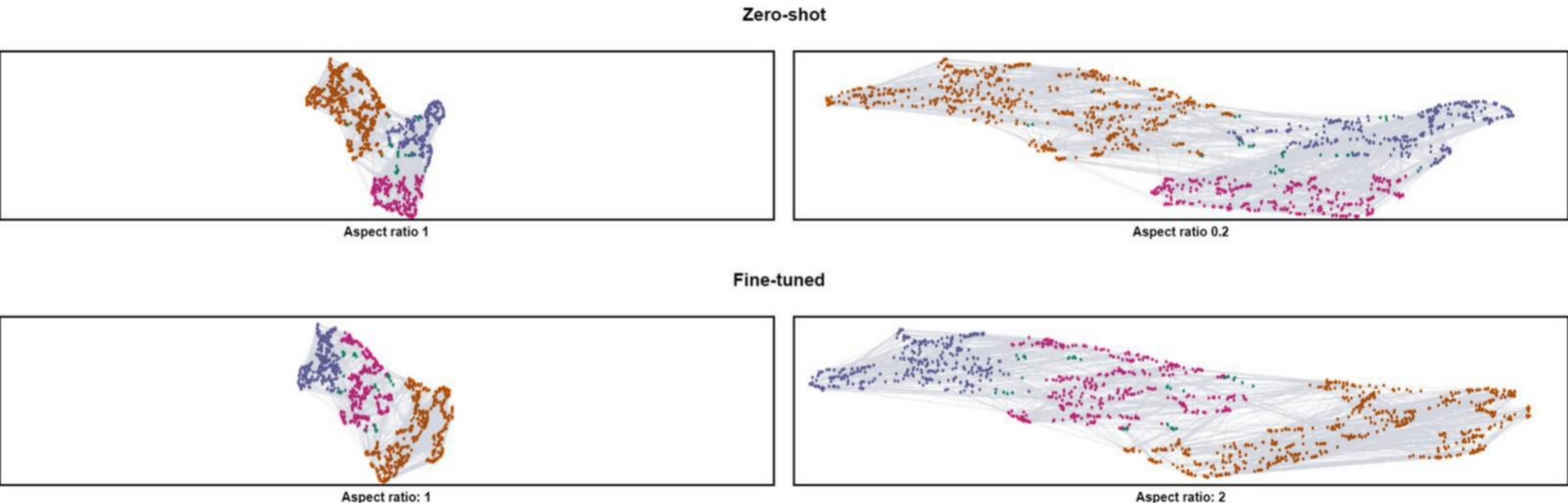}
    \caption{Global view of the embeddings of the zero-shot and the fine-tuned large models for \texttt{S1}.} 
    \label{fig:s1:moment:large:global}
\end{figure}

\begin{figure}[!htb]
    \centering
    \includegraphics[width=1\linewidth, pagebox=artbox]{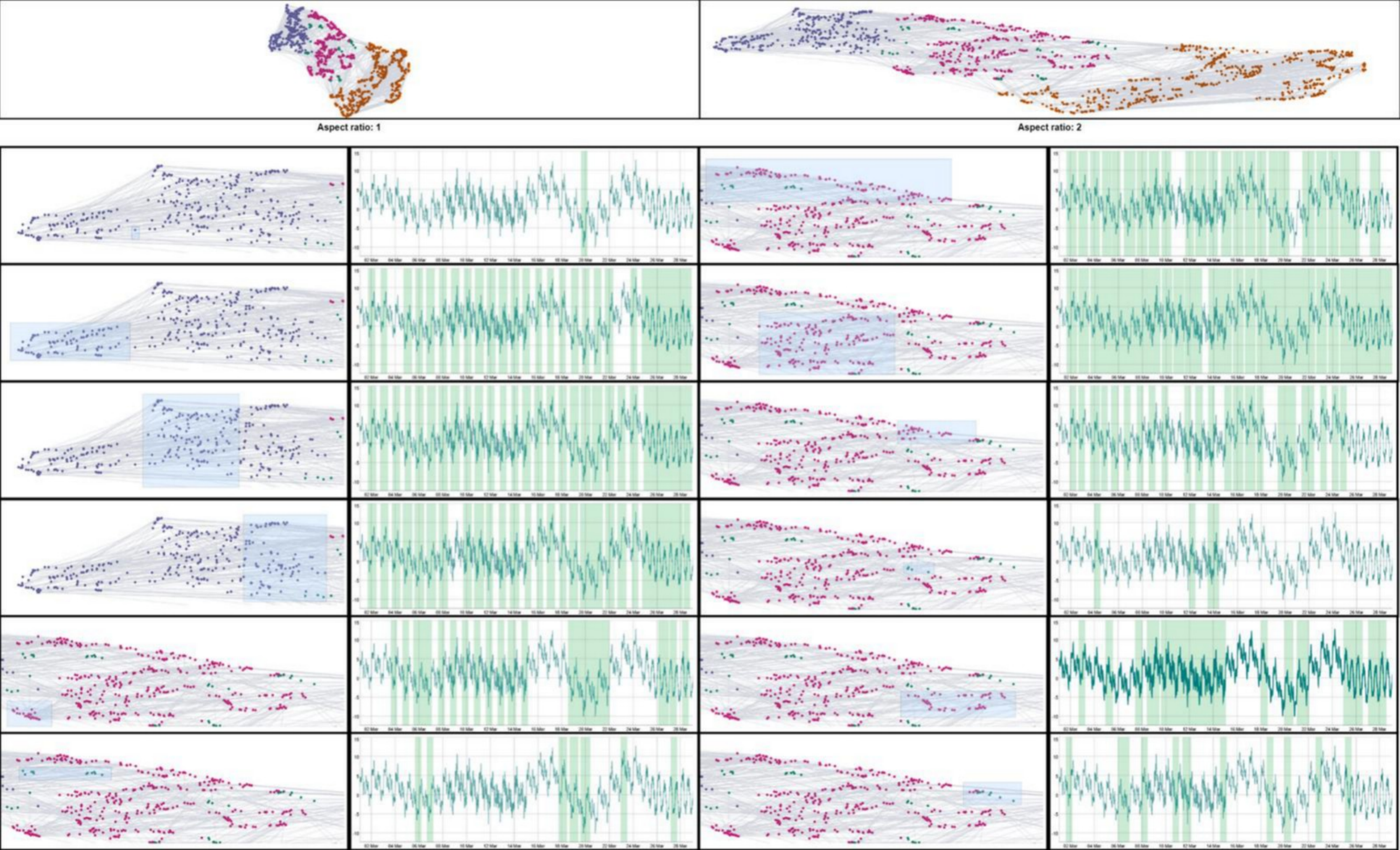}
    \caption{Analisis of \texttt{S1} using the fine-tuned large model.} 
    \label{fig:s1:moment:large:finetuned}
\end{figure}

The plots of the embeddings for \texttt{S1} using the large model in both fine-tuned and zero-shot configurations exhibit a high degree of similarity (see~\customrefs{fig:s1:moment:large:global},\ref{fig:s1:moment:large:finetuned}). Therefore, the analysis will focus primarily on the fine-tuned model, as it is expected to provide a more complete and interpretable representation.

The embedding space initially consists of three main clusters, resembling those observed in the base model embeddings (see~\customrefs{fig:s1:moment:large:global},~\ref{fig:s1:moment:base:global}). However, these clusters appear less distinct and are positioned closer to each other.

\begin{table*}[!htb]
    \centering
    \caption{Advantages and disadvantages of using MOMENT-Small, MOMENT-Base, and MOMENT-Large in zero-shot and fine-tuned configurations for \texttt{S1} dataset (segmentation).}
    \begin{tabular}{>{\centering\arraybackslash}m{3cm} >{\centering\arraybackslash}m{2.5cm} m{10cm}}
        \arrayrulecolor{black}\hline
        \textbf{Model} & \textbf{Training} & \multicolumn{1}{c}{\textbf{Observations}}\\
        \hline
        \noalign{\vskip 3pt}
        \multirow{2}{*}[-.5em]{\textbf{MOMENT-Small}} 
        & 
        zero-shot 
        & 
        {\color{ForestGreen}\ding{51}} Captures general patterns in the data. \newline
        {\color{ForestGreen}\ding{51}} Requires no additional training, reducing computational cost. \newline
        {\color{BrickRed}\ding{55}} Produces poorly defined embeddings ("blurry" clusters). \newline
        {\color{BrickRed}\ding{55}} Struggles to differentiate time series segments effectively. However, it starts to detect $df1, \, df2\,$ and $df3$. \\
        \noalign{\vskip 1pt}
        \arrayrulecolor{black!40}\cline{2-3}
        \noalign{\vskip 1pt}
        & Fine-tuned & 
        {\color{ForestGreen}\ding{51}} Moderate loss reduction ($\sim$22\%), improving embedding quality. \newline
        {\color{ForestGreen}\ding{51}} Some improvement in pattern detection. \newline
        {\color{BrickRed}\ding{55}} Visual embedding projection shows minimal improvement over zero-shot. \newline
        {\color{BrickRed}\ding{55}} Cluster structure remains unclear despite fine-tuning. \\
        \arrayrulecolor{black}\hline
        \noalign{\vskip 2pt}
        \multirow{2}{*}[-.5em]{\textbf{MOMENT-Base}} 
        & 
        zero-shot 
        & 
        {\color{ForestGreen}\ding{51}} Clearly detects $df2$ and is closer to detecting $df1$ and $df3$ than Small. \newline
        {\color{ForestGreen}\ding{51}} Captures some patterns of the time series. \newline
        {\color{BrickRed}\ding{55}} Still lacks precise segmentation of time series into clusters. We have even lost the precision with $df1$ and $df2$. \\
        \noalign{\vskip 1pt}
        \arrayrulecolor{black!40}\cline{2-3}
        \noalign{\vskip 1pt}
        & Fine-tuned & 
        {\color{ForestGreen}\ding{51}} Detects patterns. \newline
        {\color{BrickRed}\ding{55}} Fine-tuning provides only a marginal improvement. \newline
        {\color{BrickRed}\ding{55}} Cluster structure still lacks well-defined segmentation. \\
        \noalign{\vskip 1pt}
        \arrayrulecolor{black}\hline
        \noalign{\vskip 1pt}
        \multirow{2}{*}[-.5em]{\textbf{MOMENT-Large}} 
        
        & 
        zero-shot 
        & 
        {\color{ForestGreen}\ding{51}} Detects patterns. \newline
        {\color{BrickRed}\ding{55}} May be influenced by the time series' noise. \newline
        {\color{BrickRed}\ding{55}} Clusters are not very separated. \newline
        {\color{BrickRed}\ding{55}} Segments are not clear. \\
        \noalign{\vskip 1pt}
        \arrayrulecolor{black!40}\cline{2-3}
        \noalign{\vskip 1pt}
        & Fine-tuned & 
        {\color{BrickRed}\ding{55}} No special advantages. \newline
        {\color{BrickRed}\ding{55}} Fine-tuned embeddings are still highly similar to zero-shot, not getting any different result within the analysis. \\
        \noalign{\vskip 2pt}
        \arrayrulecolor{black}\hline

    \end{tabular}
    
    \label{tab:s1_moment_comparison}
\end{table*}

Table~\ref{tab:s1_moment_comparison} summarizes the analysis for \texttt{S1}. The use of larger versions of MOMENT seems to be a good idea if the computation resources are good enough as the time series are better matched. However, the fine-tuning has not shown large differences with the zero-shot version of the models. In a future analysis for segmentation, some preprocessing should be done within the training dataset, maybe reducing the noise through Simple Moving Average (SMA). Also, we should try other dimensionality reduction methods to check if they end up with more time-ordered clusters. Related to the non-enhancement of the fine-tune for the visual analysis, we should check if there is a better distance for the segmentation when optimizing the loss. Also we could try to fine-tune the model with much larger window lengths and then using it to model the original length to check if it gives a long-term relationship. Training with larger SMA orders should also help in analyzing the global structure of the time series.

\subsection{Analysis of S2}

\begin{figure}[H]
    \centering
    \includegraphics[width=1\linewidth, pagebox=artbox]{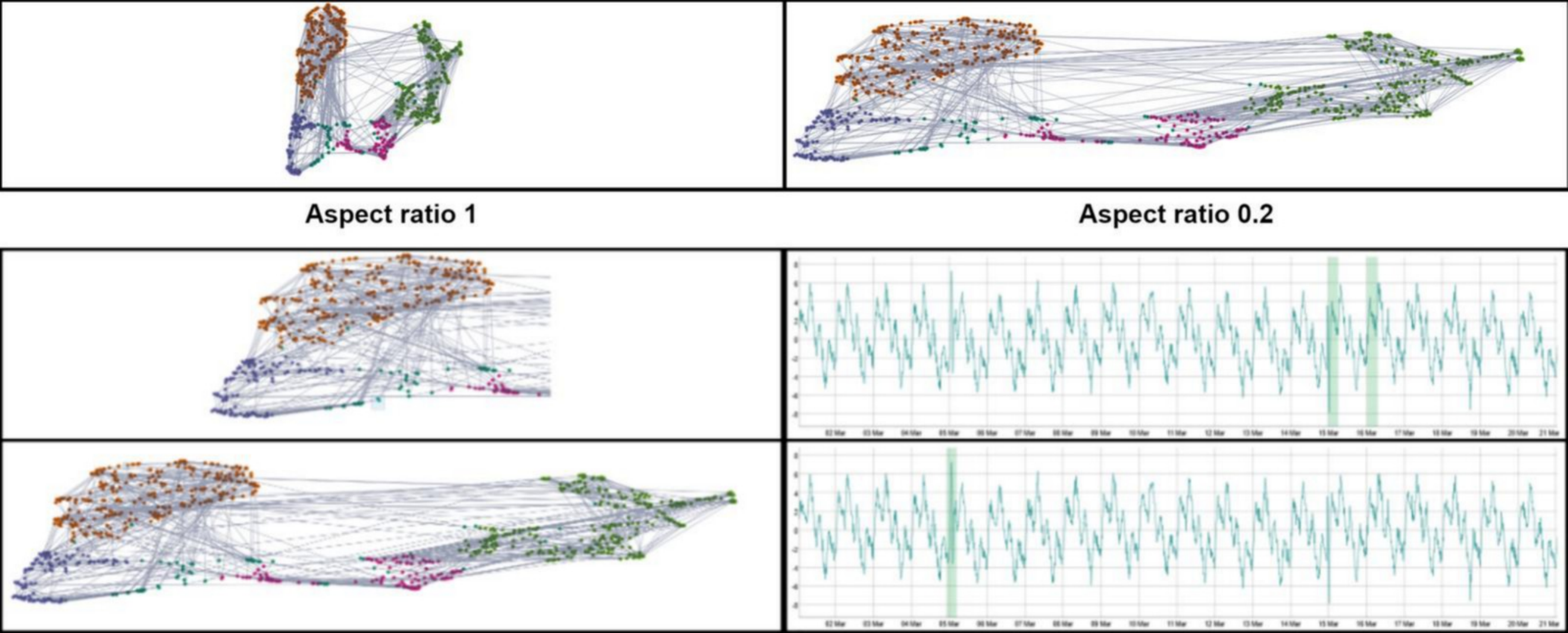}
    \caption{Global view of the embeddings of the zero-shot small model for \texttt{S2}.} 
    \label{fig:s2:moment:small:global}
\end{figure}

The \texttt{S2} dataset contains two point anomalies. In the figure~\customref{fig:s2:moment:small:global} both of them are detected in the moment-small's embedding space, but the interrelations between clusters make it difficult to detect them in a single view, losing interpretability.

If we focus on the left part of the time series and the right one, we observe anomalies in other parts of the embedding space at points that are not joined by time but short shapes. It seems that temporal correlation is being lost and each window is treated like a word in a sentence, making it difficult to detect large parts, so anomalies are easy to check (see~\customref{fig:s2:moment:small:c1}). In this part, we can find both anomalies. Also, if we focus on the right part of the time series, we see more patterns and also the presence of both anomalies of the time series (see~\customref{fig:s2:moment:small:c2}).

\begin{figure}[!htb]
    \centering
    \includegraphics[width=1\linewidth, pagebox=artbox]{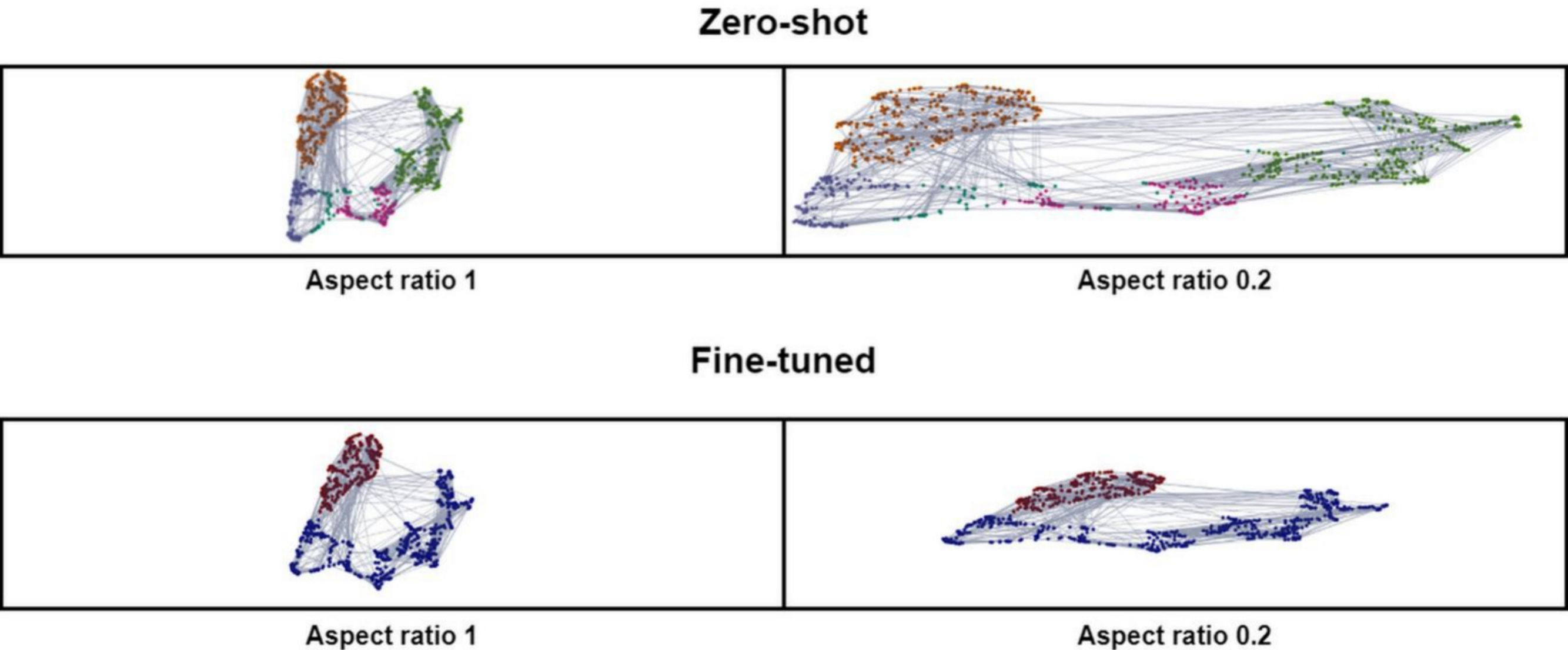}
    \caption{Analysis of the projections plot for the moment-small fine-tuned version of the model for \texttt{S2}.} 
    \label{fig:s2:moment:small:ft}
\end{figure}
\begin{figure}[!htb]
    \centering
    \includegraphics[width=1\linewidth, pagebox=artbox]{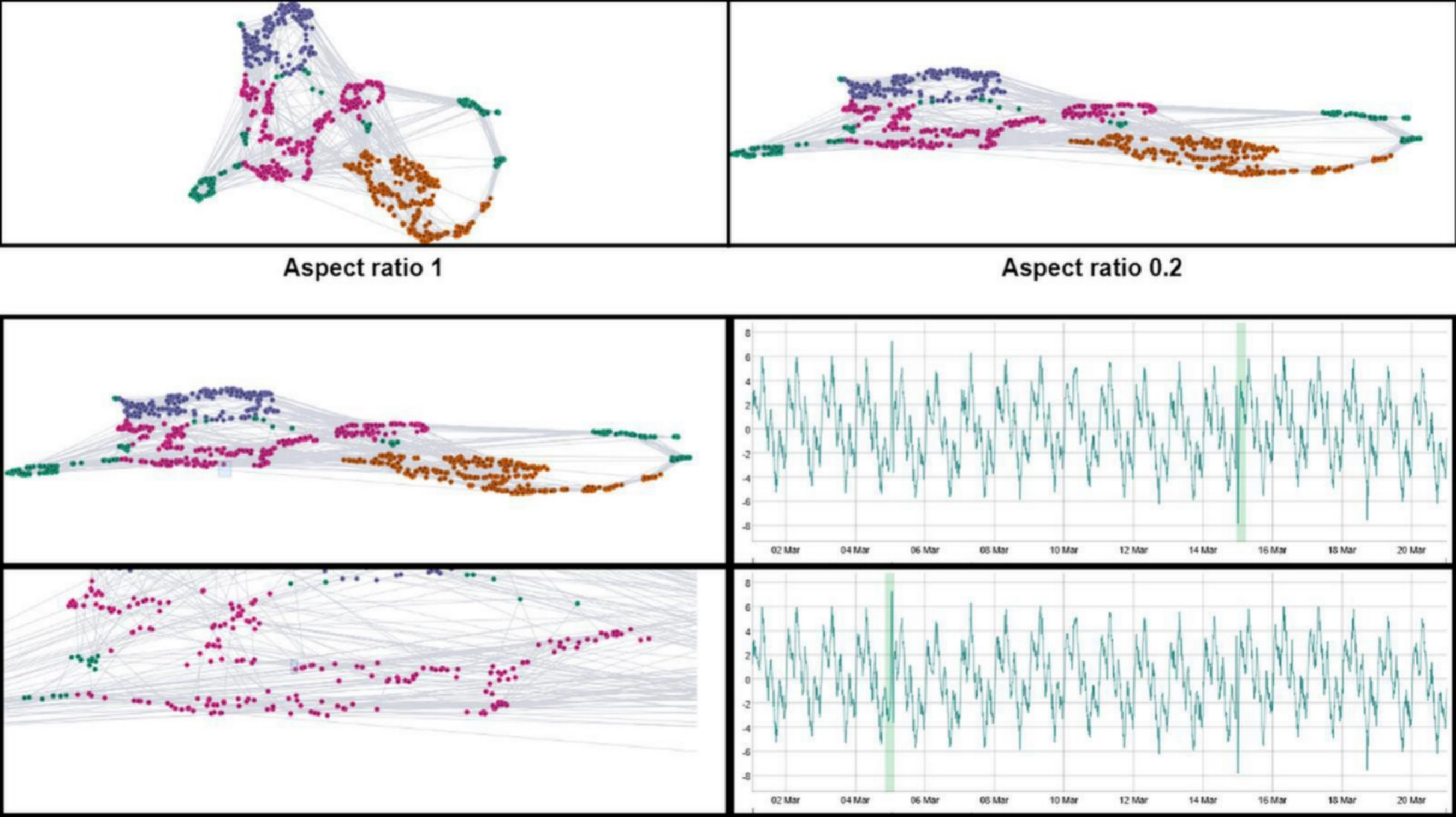}
    \caption{Global view of the embeddings projections of the zero-shot version of MOMENT-base applied to \texttt{S2}.} 
    \label{fig:s2:moment:base:global}
\end{figure}

After fine tuning moment-small, we can see that the embedding space does not change so much, with no improvement of its interpretability. 

\begin{figure}[H]
    \centering
    \includegraphics[width=1\linewidth, pagebox=artbox]{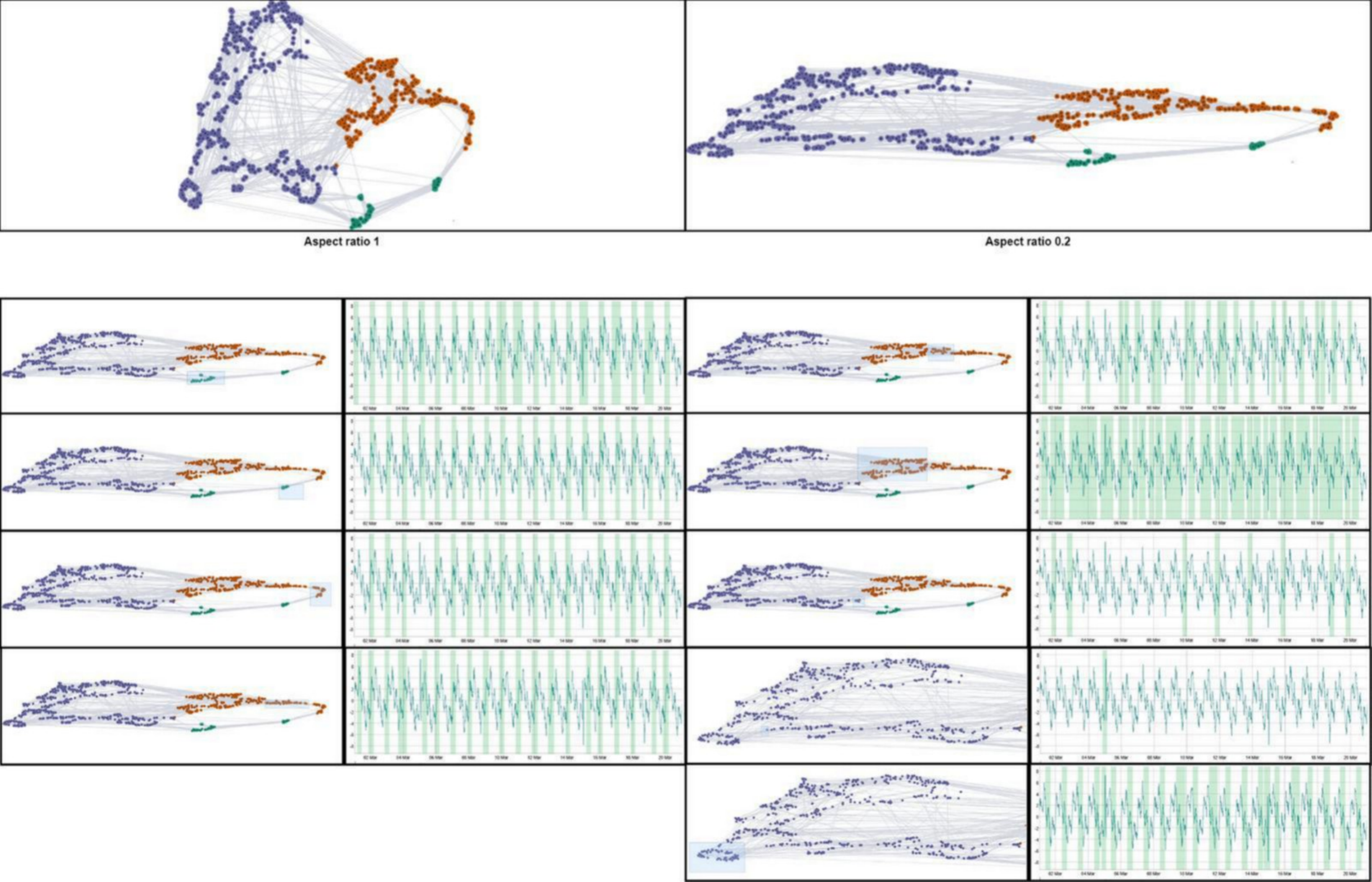}
    \caption{Embeddings projection plot of the fine-tuned version of MOMENT-base applied to \texttt{S2}.} 
    \label{fig:s2:moment:base:ft}
\end{figure}

The MOMENT-base model results on more defined clusters (see~\customrefs{fig:s2:moment:base:global},~\ref{fig:s2:moment:base:ft}). The difference between the zero-shot and the fine-tuned models is still not that much, as the only change is the rotation of positions between the brown and green clusters in the right part. Again, MOMENT seems to correctly detect small patterns but it makes it difficult to check the anomalies as they are not in the appart zones but, at least, the anomaly by the right is at a corner of the pink cluster in an edge point (very similar ``input-output'' directions, like if they were the same line). This makes it easier to detect the anomalies, but not as easy as with the embeddings of MTSAE.

\begin{figure}[!htb]
    \centering
    \includegraphics[width=1\linewidth, pagebox=artbox]{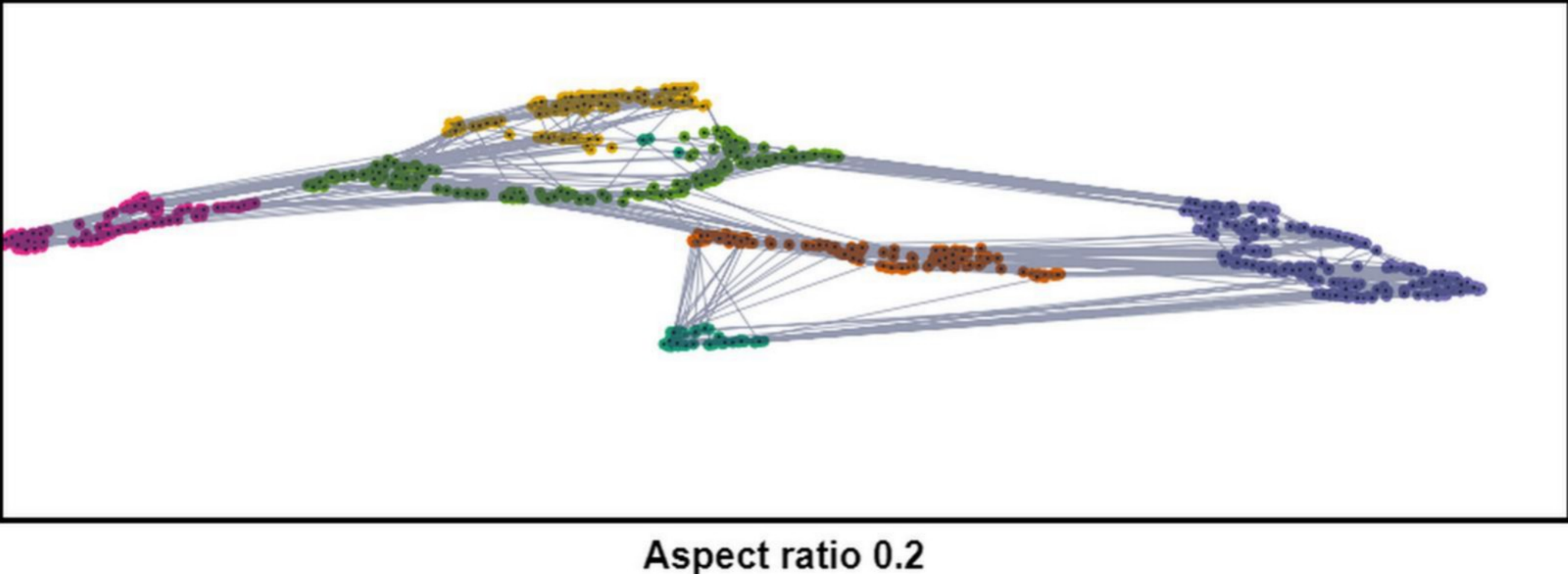}
    \caption{Global view of the embeddings projections of the zero-shot version of MOMENT-large applied to \texttt{S2}.} 
    \label{fig:s2:moment:large:global}
\end{figure}

The embedding space of the MOMENT-large model is similar to the previous ones but seems to be more accurate in the right part with the new purple cluster (see \customref{fig:s2:moment:large:global}). Focusing on the pink cluster we can see something interesting for the first time, both anomalies can be detected in its upper corner, which is a great advance in order to directly check the anomalies (see \customref{fig:s2:moment:large:c1}). 

 Figures~\ref{fig:s2:moment:large:c2},~\ref{fig:s2:moment:large:c3},~\ref{fig:s2:moment:large:c4} and~\ref{fig:s2:moment:large:c5} show the insider view of the clusters. It is interesting that again the anomalies appear in different clusters at the same time (\customref{fig:s2:moment:large:c1}, ~\ref{fig:s2:moment:large:c2}). 
 
 \begin{figure}[H]
    \centering
    \includegraphics[width=1\linewidth, pagebox=artbox]{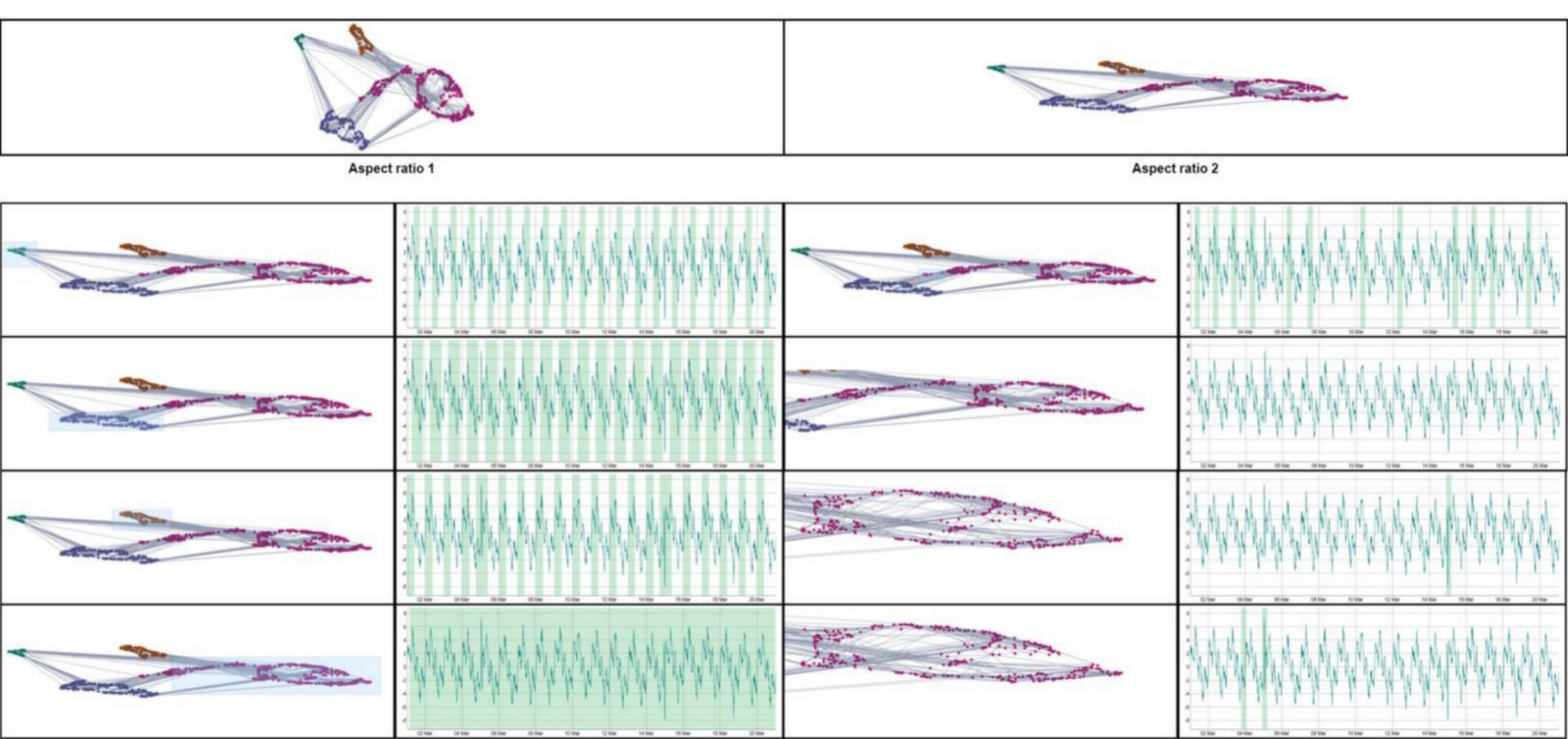}
    \caption{Embeddings projections of the fine-tuned version of MOMENT-large applied to \texttt{S2}.} 
    \label{fig:s2:moment:large:finetuned}
\end{figure}

\begin{table*}[!htb]
    \centering
    \renewcommand{\arraystretch}{1.3}
    \setlength{\tabcolsep}{10pt}
    \caption{Comparison of zero-shot vs. fine-tune for \texttt{S2} (Anomaly detection) using MOMENT-Small, MOMENT-Base, and MOMENT-Large.}
    \label{tab:s2_moment_comparison}
    \resizebox{\linewidth}{!}{
    \begin{tabular}{>{\centering\arraybackslash}m{3cm} >{\centering\arraybackslash}m{2.5cm} m{12cm}}
        \arrayrulecolor{black}\hline
        \textbf{Model} & \textbf{Training Type} & \multicolumn{1}{c}{\textbf{Observations}} \\
        \hline
        \noalign{\vskip 3pt}
        \multirow{2}{*}[-.5em]{\textbf{MOMENT-Small}} 
        & 
        Zero-shot 
        & 
        {\color{ForestGreen}\ding{51}} Clearly defined clusters. \newline
        {\color{ForestGreen}\ding{51}} Captures general patterns in \texttt{S2}. \newline
        {\color{BrickRed}\ding{55}} Clusters are highly intertwined. \newline
        {\color{BrickRed}\ding{55}} Anomalies detected but hard to isolate. \\
        \arrayrulecolor{black!40}\cline{2-3}
        & Fine-tuned & 
        {\color{ForestGreen}\ding{51}} No significant improvement. \newline
        {\color{BrickRed}\ding{55}} Anomalies remain hard to isolate. \newline
        {\color{BrickRed}\ding{55}} Cluster structure unchanged. \\
        \arrayrulecolor{black}\hline
        \noalign{\vskip 2pt}
        \multirow{2}{*}[-.5em]{\textbf{MOMENT-Small}} 
        & 
        Zero-shot 
        & 
        {\color{ForestGreen}\ding{51}} One anomaly detected at a cluster edge. \newline
        {\color{BrickRed}\ding{55}} The anomalies appear in different clusters at the same time. \\
        \arrayrulecolor{black!40}\cline{2-3}
        & Fine-tuned & 
        {\color{ForestGreen}\ding{51}} Minor improvement in anomaly detection. \newline
        {\color{BrickRed}\ding{55}} Clusters rotated but similar. \\
        \arrayrulecolor{black}\hline
        \noalign{\vskip 2pt}
        \multirow{2}{*}[-.5em]{\textbf{MOMENT-Large}} 
        & 
        Zero-shot 
        & 
        {\color{ForestGreen}\ding{51}} Most defined clusters. \newline
        {\color{ForestGreen}\ding{51}} Anomalies detected at cluster edges. \newline
        {\color{BrickRed}\ding{55}} Anomalies still appear in multiple clusters. \\
        \arrayrulecolor{black!40}\cline{2-3}
        & Fine-tuned & 
        {\color{BrickRed}\ding{55}} No significant improvement. \newline
        {\color{BrickRed}\ding{55}} Cluster structure remains nearly identical. \newline
        {\color{BrickRed}\ding{55}} The anomalies move to the middle of a cluster, making them more difficult to detect in a visual inspection. \\
        \noalign{\vskip 2pt}
        \arrayrulecolor{black}\hline
    \end{tabular}
    }
    
\end{table*}
 
 The fine-tuned version has a very similar shape. In this case, the anomalies are in the middle of the pink cluster~\customref{fig:s2:moment:large:finetuned}, difficult to detect.

Thus, in summary, anomalies are not that easy to detect. Table~\ref{tab:s2_moment_comparison} shows an overview of the analysis for each version of the dataset.

\subsection{Analysis of S3 using MOMENT}

\begin{figure*}[!htb]
    \centering
    \includegraphics[width=1\linewidth, pagebox=artbox]{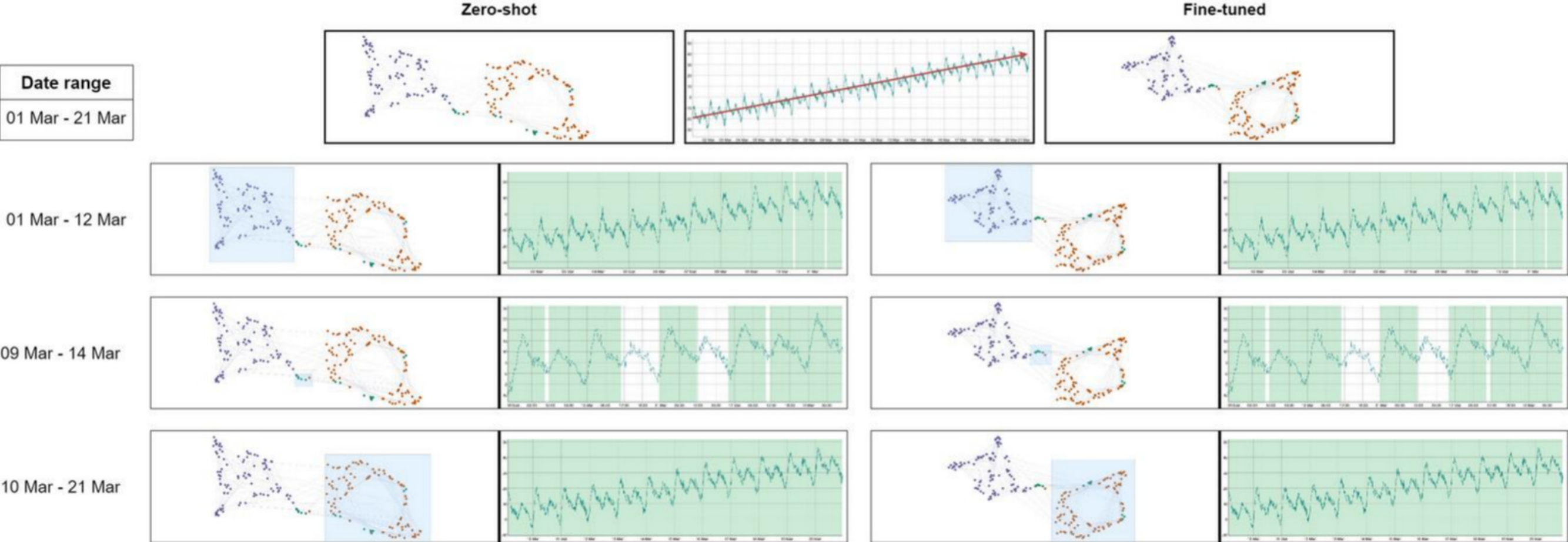}
    \caption{Embeddings projections of MOMENT-small applied to \texttt{S3}.} 
    \label{fig:s3:moment:small:zero-shot}
\end{figure*}

This dataset is used to check the interpretation of trends through the visual projection of the embedding space. The zero-shot and fine-tuned versions of MOMENT-small have very similar projections, with small rotations. Taking a loop from left to right to the different clusters, no trend is detected, showing a spring effect (see~\customref{fig:s3:moment:small:zero-shot}).

The base projections are very similar to the small ones in both the zero-shot and the fine-tuned version. Although it seems to be getting the trend when selecting the two clusters as a full object, getting inside the blue part and going through the lines, we still have the same spring effect, showing that the trend is not related to the position of the points in the embedding space (see Figs.~\ref{fig:s3:moment:base:zero-shot},~\ref{fig:s3:moment:base:finetuned}).

  The large version also shows really few differences between the zero-shot and fine-tuned embedding projections. This time, the trend seems to be detected, but it still has a spring effect in the blue cluster (see~\customref{fig:s3:moment:large}).

\begin{table*}[!htbp]
    \centering
    \renewcommand{\arraystretch}{1.4}
    \setlength{\tabcolsep}{6pt}
    \Large
    \caption{Comparison of zero-shot vs. fine-tuned for \texttt{S3} (Trends) using MOMENT-Small, MOMENT-Base, and MOMENT-Large.}
    \label{tab:s3_moment_comparison}
    \resizebox{0.75\linewidth}{!}{
    \begin{tabular}{>{\centering\arraybackslash}m{4cm} >{\centering\arraybackslash}m{3cm} m{12cm}}
        \arrayrulecolor{black}\hline
        \textbf{Model} & \textbf{Training Type} & \multicolumn{1}{c}{\textbf{Observations}} \\
        \hline
        \noalign{\vskip 3pt}
        \multirow{2}{*}[-.5em]{\textbf{MOMENT-Small}} 
        &
        Zero-shot 
        & 
        {\color{ForestGreen}\ding{51}} Defined but interconnected clusters. \newline
        {\color{BrickRed}\ding{55}} No clear trend detected, with a spring effect. \\
        \arrayrulecolor{black!40}\cline{2-3}
        & Fine-tuned 
        & {\color{BrickRed}\ding{55}} No significant changes. \newline
        {\color{BrickRed}\ding{55}} Spring effect remains in cluster visualization. \\
        \arrayrulecolor{black}\hline
        \noalign{\vskip 2pt}
        \multirow{2}{*}[-.5em]{\textbf{MOMENT-Base}} 
        &
        Zero-shot 
        &
        {\color{ForestGreen}\ding{51}} Some structures suggest trend detection. \newline
        {\color{BrickRed}\ding{55}} Still no fully defined trends. \\
        \arrayrulecolor{black!40}\cline{2-3}
        & Fine-tuned 
        & {\color{ForestGreen}\ding{51}} Slightly clearer structure. \newline
        {\color{BrickRed}\ding{55}} Trend still mixed within clusters. \\
        \arrayrulecolor{black}\hline
        \noalign{\vskip 2pt}
        \multirow{2}{*}[-.5em]{\textbf{MOMENT-Large}} 
        & 
        Zero-shot 
        & 
        {\color{ForestGreen}\ding{51}} Best model for detecting trend-like structures. \newline
        {\color{BrickRed}\ding{55}} Spring effect still present. \\
        \arrayrulecolor{black!40}\cline{2-3}
        & Fine-tuned 
        {\color{BrickRed}\ding{55}} No significant differences with the zero-shot. \\
        \noalign{\vskip 2pt}
        \arrayrulecolor{black}\hline
    \end{tabular}
    }
\end{table*}

\subsection{Analysis of Kohl's time series using MOMENT}

\begin{figure*}[!htb]
    \centering
    \includegraphics[width=1\linewidth, pagebox=artbox]{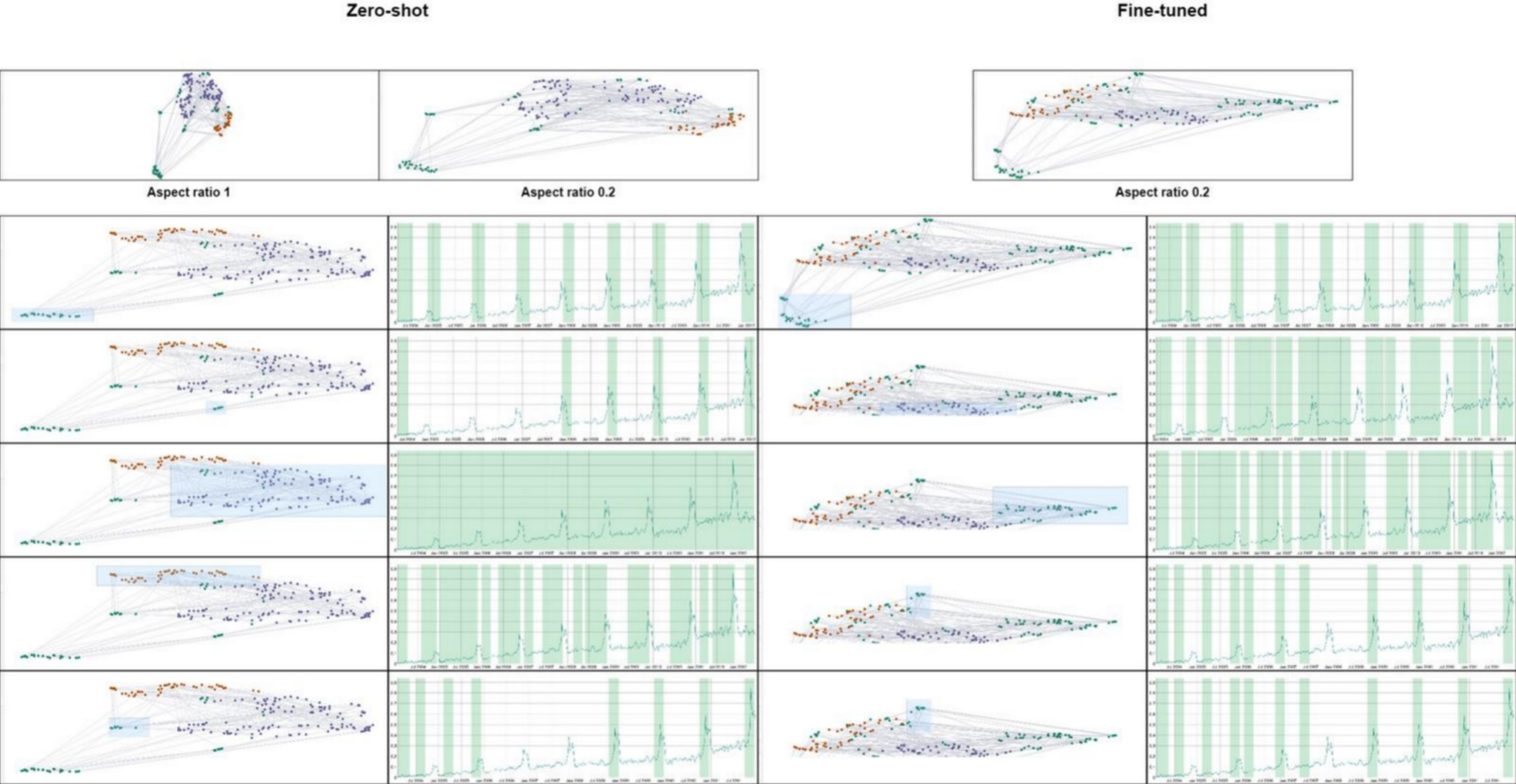}
    \caption{Embeddings projections of MOMENT-small applied to Kohls.} 
    \label{fig:kohls:moment:small}
\end{figure*}

The Kohl's time series is used for analyzing both segmentation and trend. The model is supposed to fit both the upper trend and the segmentation between peaks and plains. 

The MOMENT-small-zero-shot versions show two distinct clusters, however. The green one (in the bottom) is near to detect all the peaks - valleys - (with some temporal shift to the next timesteps)) and the first plain zone. Those peaks seems to be detected if using the other two small green clusters. However, when we select the rest of the embedding space, we see that the segmentation is not that clear as the blue part contains the full time series. Also, the ``linear'' pattern of the trend within the embedding space is not detected. The fine-tuned version has more defined clusters. The ``plains'' are nearly learned, in the green part. However, some parts are shared between all the clusters, like the peaks and the first plain. It is easier to detect, but there is no clear trend or segmentation pattern in the plot (see ~\customref{fig:kohls:moment:small}). The MOMENT-base model seems to be clearly detecting the peaks but not the plains, correctly detecting them as a pattern more than a segmentation. In fact, the blue clusters gets the full time series again instead of just the plain parts. Again, no trend visual pattern is found (see ~\customref{fig:kohls:moment:base}). For the MOMENT-large version, see that the zero-shot and the fine-tuned model are really similar, with a rotation in the right part (and a subtle more definition). This time, the fine-tune and the zero-shot models are more different (see~\customrefs{fig:kohls:moment:large:zero-shot},~\ref{fig:kohls:moment:large:finetuned}). The zero-shot model shows clear segmentation between peaks and plains (except for the first plain), showing a great performance in detecting the plains while detecting both patterns and intermediates. The fine-tuned version seemed to be distinct, but the results are really similar.

\begin{table*}[!htbp]
    \centering
    \caption{Comparison of zero-shot vs. fine-tuned for Kohl’s using MOMENT-Small, MOMENT-Base, and MOMENT-Large.}
    \label{tab:kohls_moment_comparison}
    \renewcommand{\arraystretch}{1.4}
    \setlength{\tabcolsep}{6pt}
    \Large
    \resizebox{0.75\linewidth}{!}{
    \begin{tabular}{>{\centering\arraybackslash}m{4cm} >{\centering\arraybackslash}m{3cm} m{12cm}}
        \arrayrulecolor{black}\hline
        \textbf{Model} & \textbf{Training Type} & \multicolumn{1}{c}{\textbf{Observations}} \\
        \hline
        \noalign{\vskip 3pt}
        \multirow{2}{*}[-.5em]{\textbf{MOMENT-Small}} 
        & 
        Zero-shot 
        &
        {\color{ForestGreen}\ding{51}} Clear separation of peaks. \newline
        {\color{BrickRed}\ding{55}} Trend not fully detected. \newline
        {\color{BrickRed}\ding{55}} First plain is mixed within clusters. \newline
        {\color{BrickRed}\ding{55}} Not clear separation of the plains. \\
        \arrayrulecolor{black!40}\cline{2-3}
        & Fine-tuned 
        & {\color{ForestGreen}\ding{51}} More defined clusters. \newline
        {\color{ForestGreen}\ding{51}} Plains are closer to being detected. \newline
        {\color{BrickRed}\ding{55}} First plain shared between all the clusters. \newline
        {\color{BrickRed}\ding{55}} Plain section still containing peaks. \newline
        {\color{BrickRed}\ding{55}} Still no clear trend detection. \\
        \arrayrulecolor{black}\hline
        \noalign{\vskip 2pt}
        \multirow{2}{*}[-.5em]{\textbf{MOMENT-Base}} 
        &
        Zero-shot 
        &
        {\color{ForestGreen}\ding{51}} Peaks are well defined. \newline
        {\color{BrickRed}\ding{55}} Plains are not clearly detected. \\
        \arrayrulecolor{black!40}\cline{2-3}
        & Fine-tuned 
        & {\color{ForestGreen}\ding{51}} Slight improvement in segmentation. \newline
        {\color{BrickRed}\ding{55}} No clear improvement in trend detection. \\
        \arrayrulecolor{black}\hline
        \noalign{\vskip 2pt}
        \multirow{2}{*}[-.5em]{\textbf{MOMENT-Large}} 
        &
        Zero-shot 
        &
        {\color{ForestGreen}\ding{51}} Best model for segmentation. \newline
        {\color{ForestGreen}\ding{51}} Peaks and plains clusters are visually distinct. \newline
        {\color{BrickRed}\ding{55}} Trend is still not fully learned. \\
        \arrayrulecolor{black!40}\cline{2-3}
        & Fine-tuned 
        & {\color{BrickRed}\ding{55}} No major improvement. \newline
        {\color{BrickRed}\ding{55}} Trend detection remains similar to zero-shot. \\
        \noalign{\vskip 2pt}
        \arrayrulecolor{black}\hline
    \end{tabular}
    }
    
\end{table*}

\subsection{Analysis of Toy time series}
\begin{figure*}[!htb]
    \centering
    \includegraphics[width=1\linewidth, pagebox=artbox]{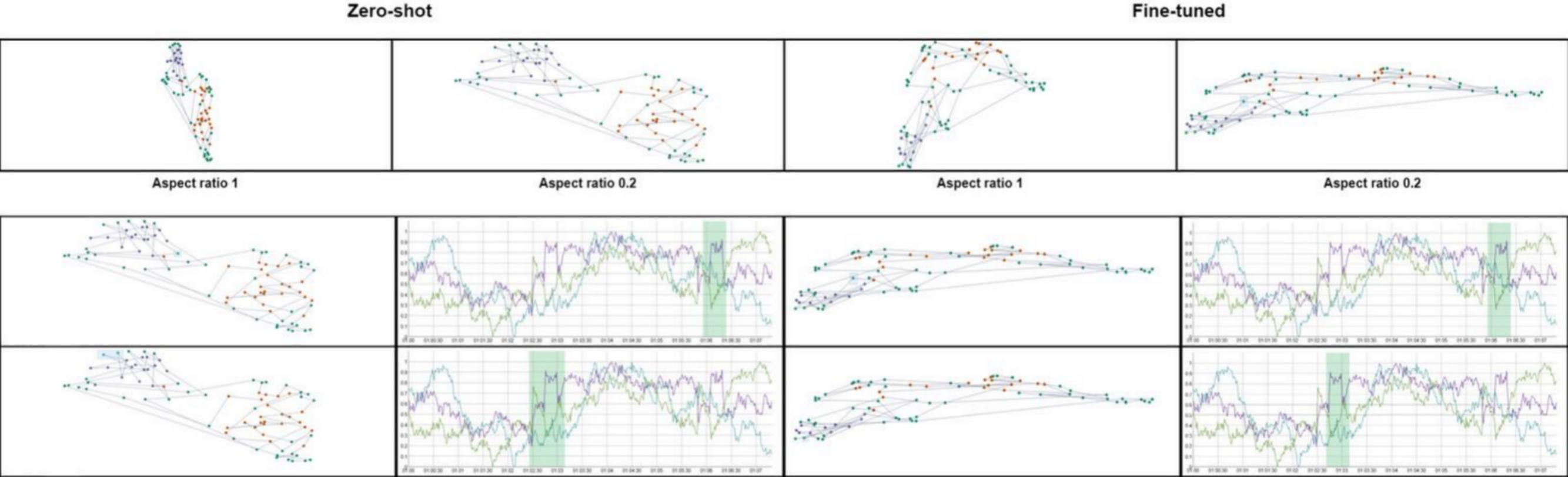}
    \caption{Embeddings projections of MOMENT-small applied to \texttt{M-Toy}.} 
    \label{fig:toy:moment:small}
\end{figure*}

The \texttt{M-Toy} dataset is proposed for analyzing sequence anomalies in multivariate time series. The small versions can detect the two anomalies of the dataset. In both the fine-tuned and the zero-shot version, the anomaly by the left is really easy to check as an edge point, but the right anomaly is detected in the middle of a cluster, making it more likely a transition point and thus difficultying the detection (see~\customref{fig:toy:moment:small}). The base-zero-shot version is more difficult to analyze. However, when applying the fine-tune, the anomalies get easier to detect and get really near between them, showing a relation between both patterns (see Fig~\customref{fig:toy:moment:base}). The large zero-shot version is more difficult to read, especially for the anomaly by the left, as it gets mixed with intermediate patterns (which may also be interesting). The large-finetune version results with similar plot with a better interpretability of the anomaly on the right. However, the anomaly by the left is even more difficult to detect within the embedding space.

\begin{table*}[H]
    \centering
    \renewcommand{\arraystretch}{1.4}
    \setlength{\tabcolsep}{6pt}
    \Large
    \resizebox{0.65\linewidth}{!}{
    \begin{tabular}{>{\centering\arraybackslash}m{4cm} >{\centering\arraybackslash}m{3cm} m{12cm}}
        \arrayrulecolor{black}\hline
        \textbf{Model} & \textbf{Training Type} & \multicolumn{1}{c}{\textbf{Observations}} \\
        \hline
        \noalign{\vskip 3pt}
        \multirow{2}{*}[-.5em]{\textbf{MOMENT-Small}} 
        &
        Zero-shot 
        & 
        {\color{ForestGreen}\ding{51}} Detects the left anomaly as an edge point. \newline
        {\color{BrickRed}\ding{55}} Right anomaly is mixed within clusters. \\
        \arrayrulecolor{black!40}\cline{2-3}
        & Fine-tuned 
        & {\color{ForestGreen}\ding{51}} Left anomaly remains easy to detect. \newline
        {\color{BrickRed}\ding{55}} No major improvement in right anomaly visibility. \\
        \arrayrulecolor{black}\hline
        \noalign{\vskip 2pt}
        \multirow{2}{*}[-.5em]{\textbf{MOMENT-Base}} 
        &
        Zero-shot 
        &
        {\color{ForestGreen}\ding{51}} Anomalies are more structured than in Small. \newline
        {\color{BrickRed}\ding{55}} Right anomaly is still not well-separated. \\
        \arrayrulecolor{black!40}\cline{2-3}
        & Fine-tuned 
        & {\color{ForestGreen}\ding{51}} Right anomaly becomes slightly more distinguishable. \newline
        {\color{BrickRed}\ding{55}} Left anomaly detection remains unchanged. \\
        \arrayrulecolor{black}\hline
        \noalign{\vskip 2pt}
        \multirow{2}{*}[-.5em]{\textbf{MOMENT-Large}} 
        &
        Zero-shot 
        &
        {\color{ForestGreen}\ding{51}} Best structured clusters. \newline
        {\color{ForestGreen}\ding{51}} Left anomaly still visible. \newline
        {\color{BrickRed}\ding{55}} Right anomaly is harder to isolate. \\
        \arrayrulecolor{black!40}\cline{2-3}
        & Fine-tuned 
        & {\color{BrickRed}\ding{55}} No significant improvement. \newline
        {\color{BrickRed}\ding{55}} Right anomaly remains difficult to distinguish. \\
        \noalign{\vskip 2pt}
        \arrayrulecolor{black}\hline
    \end{tabular}
    }
    \caption{Comparison of zero-shot vs. fine-tuned for Toy using MOMENT-Small, MOMENT-Base, and MOMENT-Large.}
    \label{tab:toy_moment_comparison}
\end{table*}

\section{Conclusions and future work \label{sec:conclusions}}

\begin{table*}[!ht]
    \centering
    \renewcommand{\arraystretch}{1.4}
    \setlength{\tabcolsep}{6pt}
    
    \begin{tabular}{>{\centering\arraybackslash}m{1.2cm} 
                    >{\centering\arraybackslash}m{3.5cm} 
                    >{\centering\arraybackslash}m{3.5cm} 
                    >{\centering\arraybackslash}m{3.5cm} 
                    >{\centering\arraybackslash}m{3.5cm}}
        \arrayrulecolor{black}\hline
        \textbf{Model} & 
        \multicolumn{1}{c}{\textbf{Anomaly Detection}} & 
        \multicolumn{1}{c}{\textbf{Pattern Detection}} & 
        \multicolumn{1}{c}{\textbf{Segmentation}} & 
        \multicolumn{1}{c}{\textbf{Trend Detection}} \\
        \hline
        \noalign{\vskip 3pt}
        \textbf{MOMENT-Small} & 
        {\color{BrickRed}\ding{55}} Anomalies detected but difficult to isolate. & 
        {\color{ForestGreen}\ding{51}} Some patterns detected. & 
        {\color{BrickRed}\ding{55}} No segmentation. & 
        {\color{BrickRed}\ding{55}} No visible trend detection. \\
        \arrayrulecolor{black!40}\hline
        \arrayrulecolor{black!40}\noalign{\vskip 2pt}
        \textbf{MOMENT-Base} & 
        {\color{ForestGreen}\ding{51}} Minor improvements in detecting anomalies at edges. & 
        {\color{ForestGreen}\ding{51}} Patterns are clearer than using Small. & 
        {\color{BrickRed}\ding{55}} Some, but incomplete. & 
        {\color{BrickRed}\ding{55}} Trend detection is suggested but unclear. \\
        \arrayrulecolor{black!40}\hline
        \textbf{MOMENT-Large} & 
        {\color{ForestGreen}\ding{51}} Anomalies more difficult to detect than with base. & 
        {\color{ForestGreen}\ding{51}} Most structured pattern detection. & 
        {\color{BrickRed}\ding{55}}  No clear segments. & 
        {\color{BrickRed}\ding{55}} Still lacks clear trend separation. \\
        \noalign{\vskip 2pt}
        \arrayrulecolor{black}\hline
    \end{tabular}
    \caption{Summary of model performance across different tasks (anomaly detection, pattern detection, segmentation, trend detection).}
    \label{tab:moment_task_comparison}
\end{table*}

Table~\ref{tab:moment_task_comparison} shows how MOMENT's embeddings are not that easy to check visually in order to detect the timeseries characteristics. The best performance is obtained for short pattern detection. Also, the clusters are really interconnected, showing that the time correlation may not be correctly detected. 

Focusing on the research questions~\rqref{rq:moment:1} and~\rqref{rq:moment:2}, the fine-tuned loss performance is directly related to the percentage of the dataset used. However, after selecting the best cases within the statistical analysis, the model is not likely showing a great difference within the UMAP followed by PCA projections. Thus, it seems that the improvement in terms of loss is definitely not related to the fine-tune level (in fine-tune percentages), but we open the question whether MOMENT is able to detect the same patterns than MTSAE if trained with the full training dataset. However, this is against the idea of having a faster interaction due to the less need of training (still, we could check whether a full training of MOMENT in any of its versions is faster than MTSAE or not). 

In conclusion, the analysis suggests that a loss improvement for a moment model version is not directly related to the embeddings precision, so the small version should show good results in future analysis, resulting in lower time and memory cost within the analysis. Thus, the integration of foundation models with few parameters into visual analytics tools is a great idea to save waiting time. Thus, the integration of MOMENT into deepVATS - with the option of changing it to any other foundation model - is a great contribution that offers a flexible and efficient way for visually analyzing large time series data. 

To refine this fine-tune approach, different paths for improvement are identified: exploring alternative loss functions - such as soft-DTW~\cite{cuturi2017soft} - could produce more refined embeddings, while testing different projection methods (e.g. PCA without UMAP and t-SNE) would help determine the impact of the chosen technique on the results. In addition, incorporating an option to preprocess the data prior to the fine-tune could result in clearer visualizations, and analyzing the effect of fine-tuning on each model later, including selective freexing of certain layers, could improve the embedding space projection interpretability. Either of these options would represent a viable strategy to further improve the application. 

\appenAcknow

\clearpage

\FloatBarrier

\section{Appendix}
\FloatBarrier
\subsection{Statistical analysis auxiliar plots}
\label{ap:statistics}

This section contains the auxiliar plots for the statistical analysis section.

\begin{figure}[!htb]
    \centering
    \includegraphics[width=1\linewidth, pagebox=artbox]{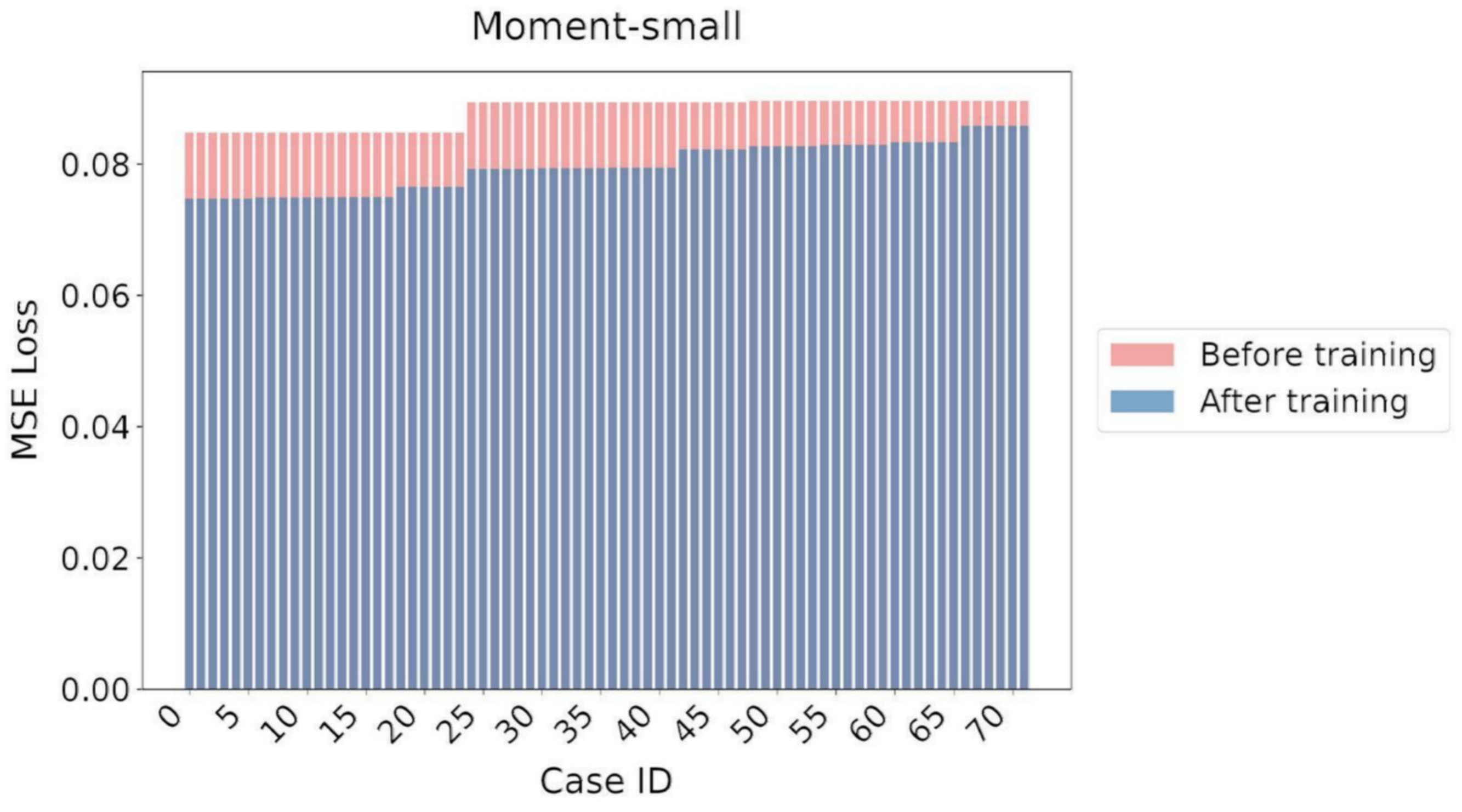}
    \caption{MOMENT-Small. MSE Losses for all the cases in the experimentation. The red line is used to show the MSE Loss for the original model used for Kohl's series. Each bar represent the MSE Loss before (red) and after (blue) the fine-tuning.}
    \label{fig:moment:losses:small}
\end{figure}

\begin{figure}[!htb]
    \centering
    \includegraphics[width=1\linewidth, pagebox=artbox]{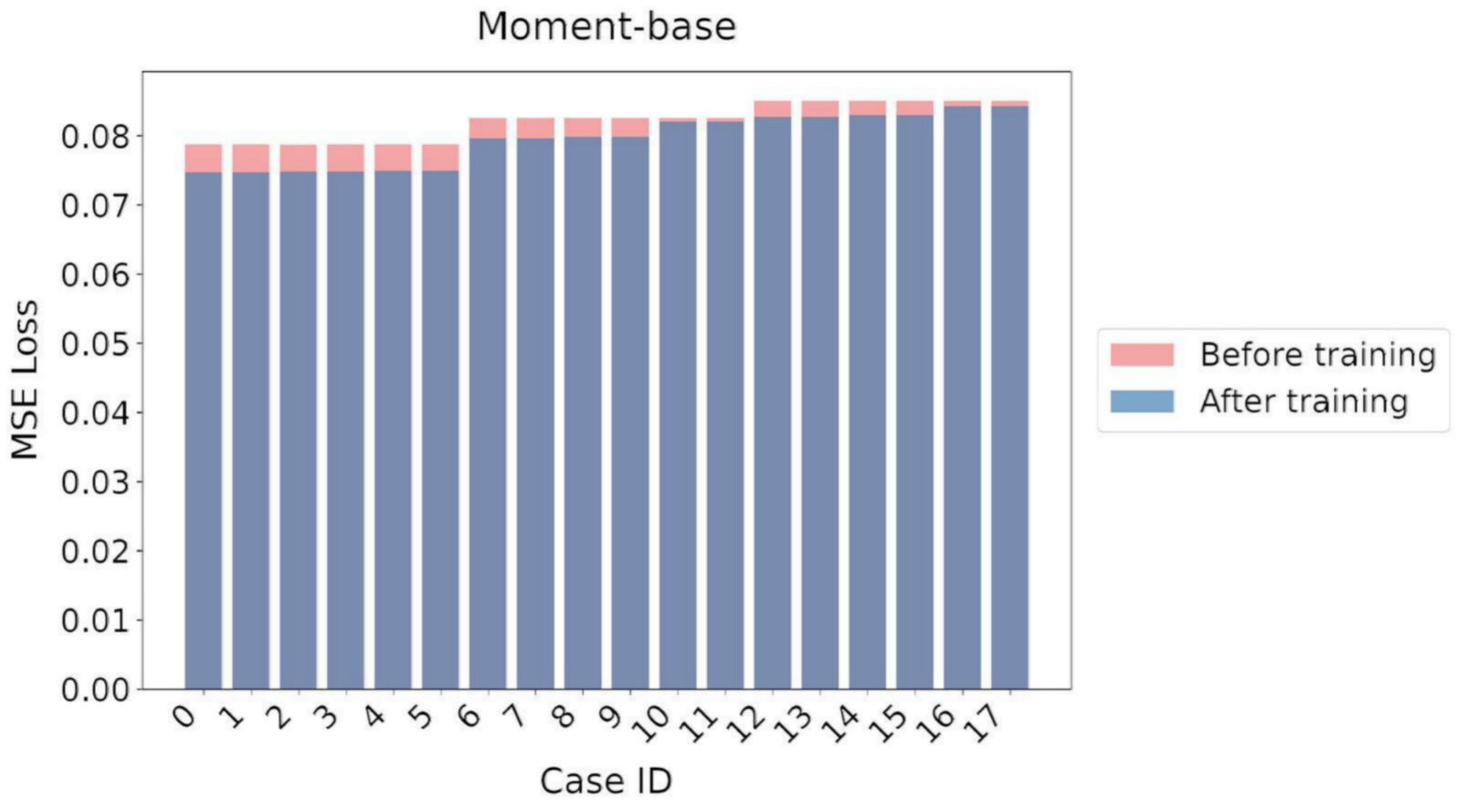}
    \caption{MOMENT-Base. MSE Losses for all the cases in the experimentation. The red line is used to show the MSE Loss for the original model used for Kohl's series. Each bar represent the MSE Loss before (red) and after (blue) the fine-tuning.}
    \label{fig:moment:losses:base}
\end{figure}

\begin{figure}[!htb]
    \centering
    \includegraphics[width=1\linewidth, pagebox=artbox]{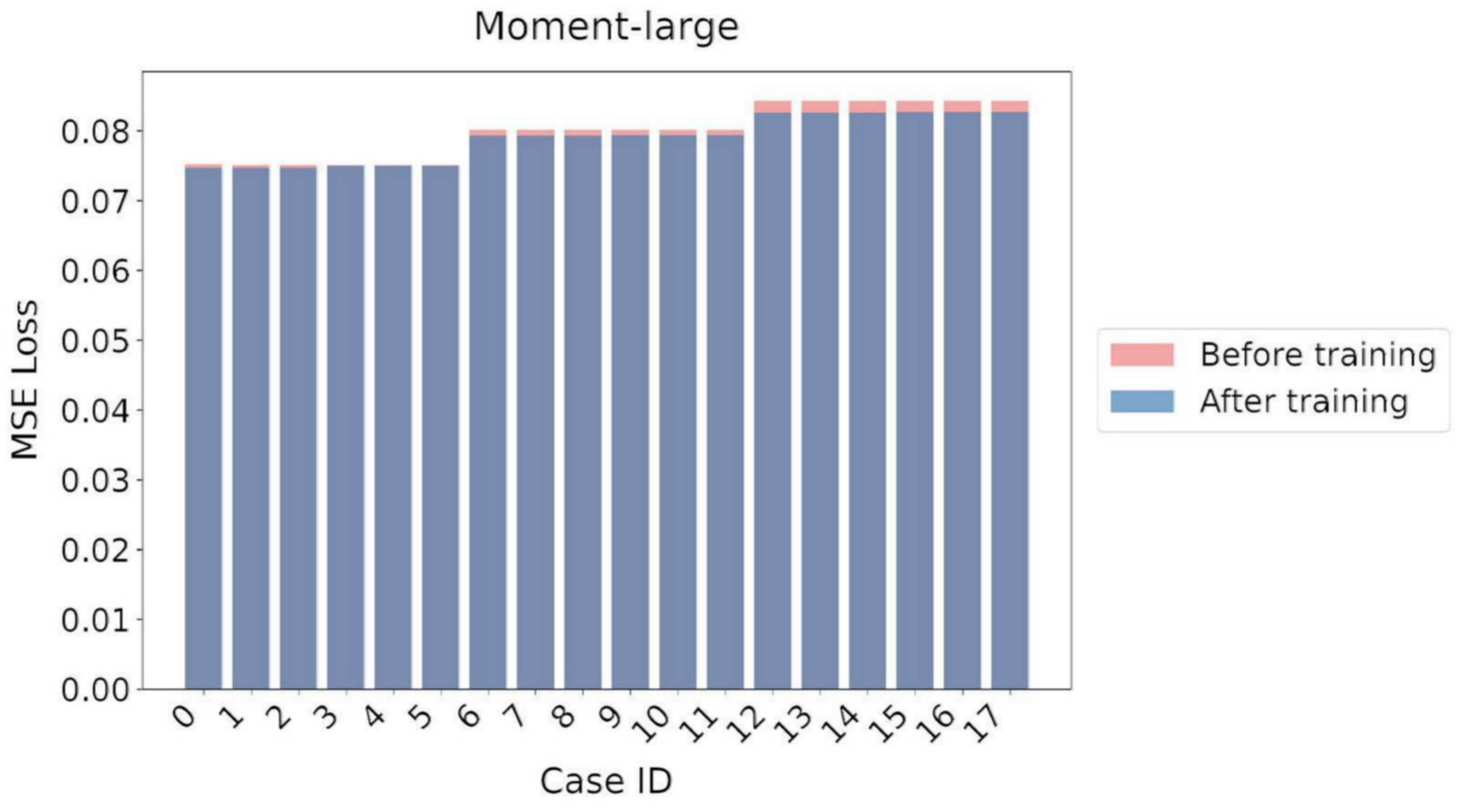}
    \caption{MOMENT-Large. MSE Losses for all the cases in the experimentation. The red line is used to show the MSE Loss for the original model used for Kohl's series. Each bar represent the MSE Loss before (red) and after (blue) the fine-tuning.}
    \label{fig:moment:losses:large}
\end{figure}

First, the loss comparison (before and after the fine-tuning)~\customrefs{fig:moment:losses:small},~\ref{fig:moment:losses:base},~\ref{fig:moment:losses:large} using $MSE$ as loss function. This plots allow to get an idea of the evolution of the loss within the execution cases.

\begin{figure}[!htb]
    \centering
    \includegraphics[width=1\linewidth, pagebox=artbox]{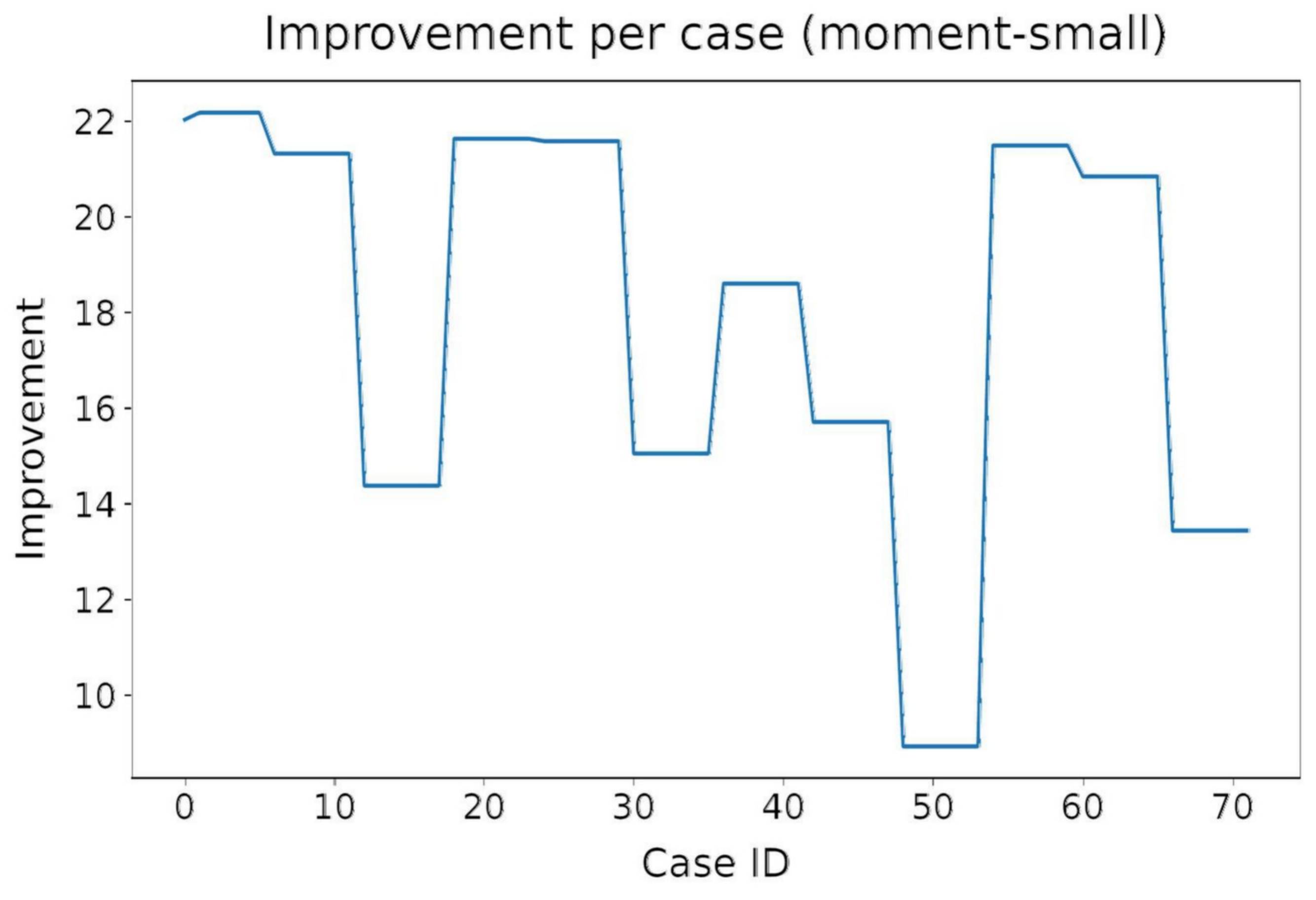}
    \caption{Loss improvement for MOMENT-small across the experimentation.} 
    \label{fig:improvements:small}
\end{figure}

\begin{figure}[!htb]
    \centering
    \includegraphics[width=1\linewidth, pagebox=artbox]{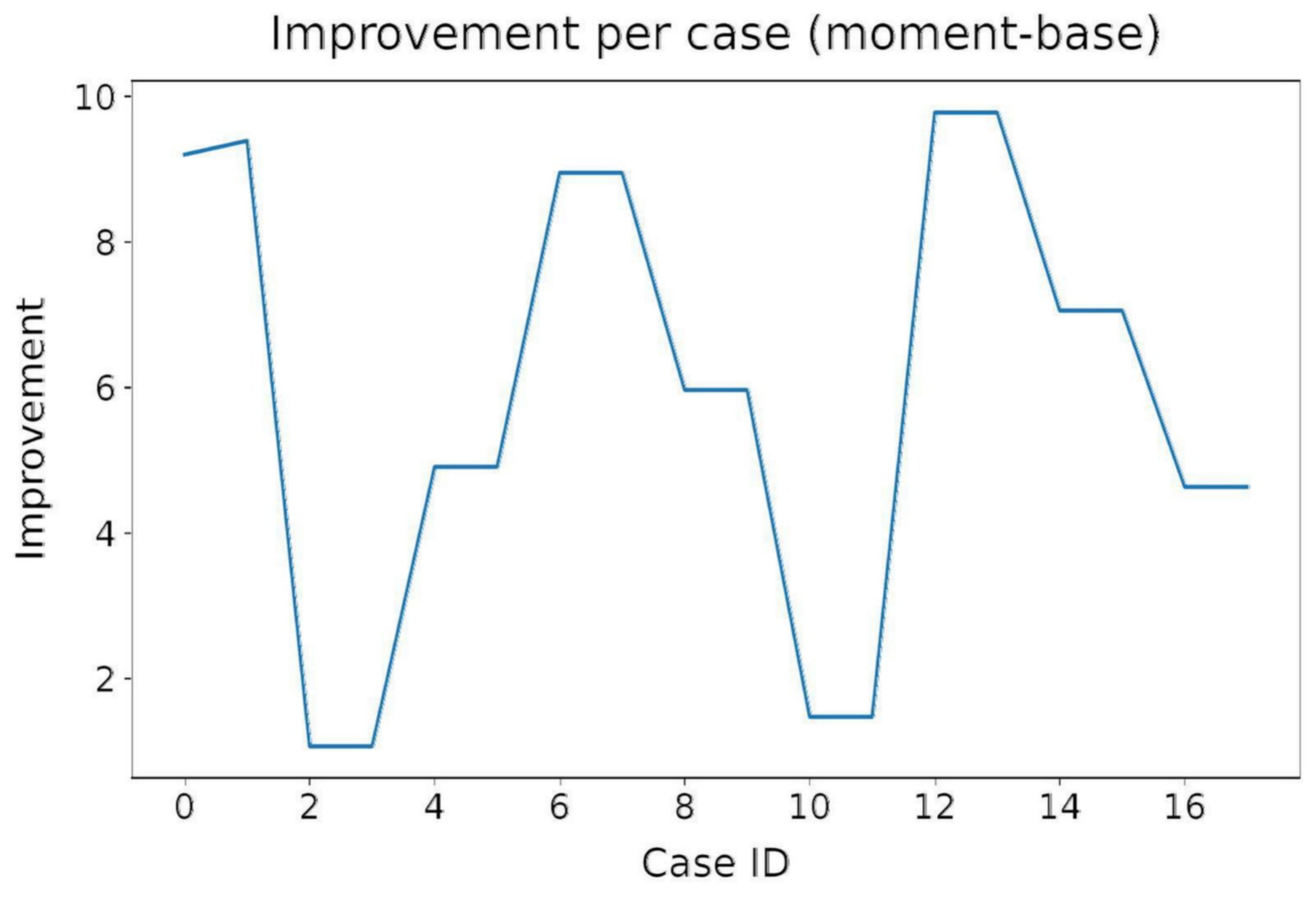}
    \caption{Loss improvement for MOMENT-base across the experimentation.} 
    \label{fig:improvements:base}
\end{figure}

\begin{figure}[!!htb]
    \centering
    \includegraphics[width=1\linewidth, pagebox=artbox]{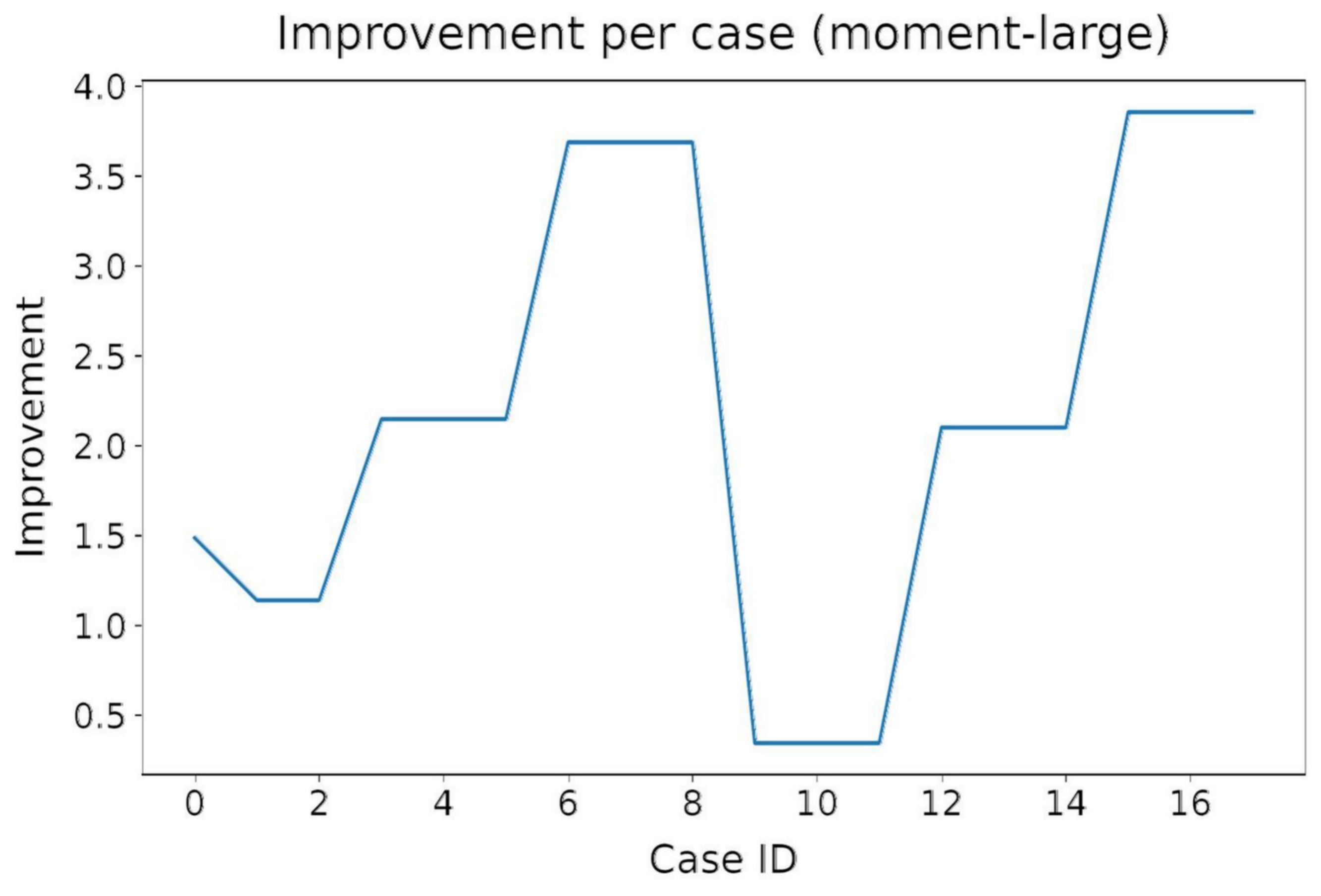}
    \caption{Loss improvement for MOMENT-large across the experimentation.} 
    \label{fig:improvements:large}
\end{figure}

Second, the loss improvement per analyzed case~\customref{fig:improvements:small},~\ref{fig:improvements:base},~\ref{fig:improvements:large} offers a better view in terms of percentage on how much we can improve the loss within the trainings, showing how the small changes on the previous plot really make a significant difference.

\begin{figure}[!htb]
    \centering
    \includegraphics[width=0.8\linewidth, pagebox=artbox]{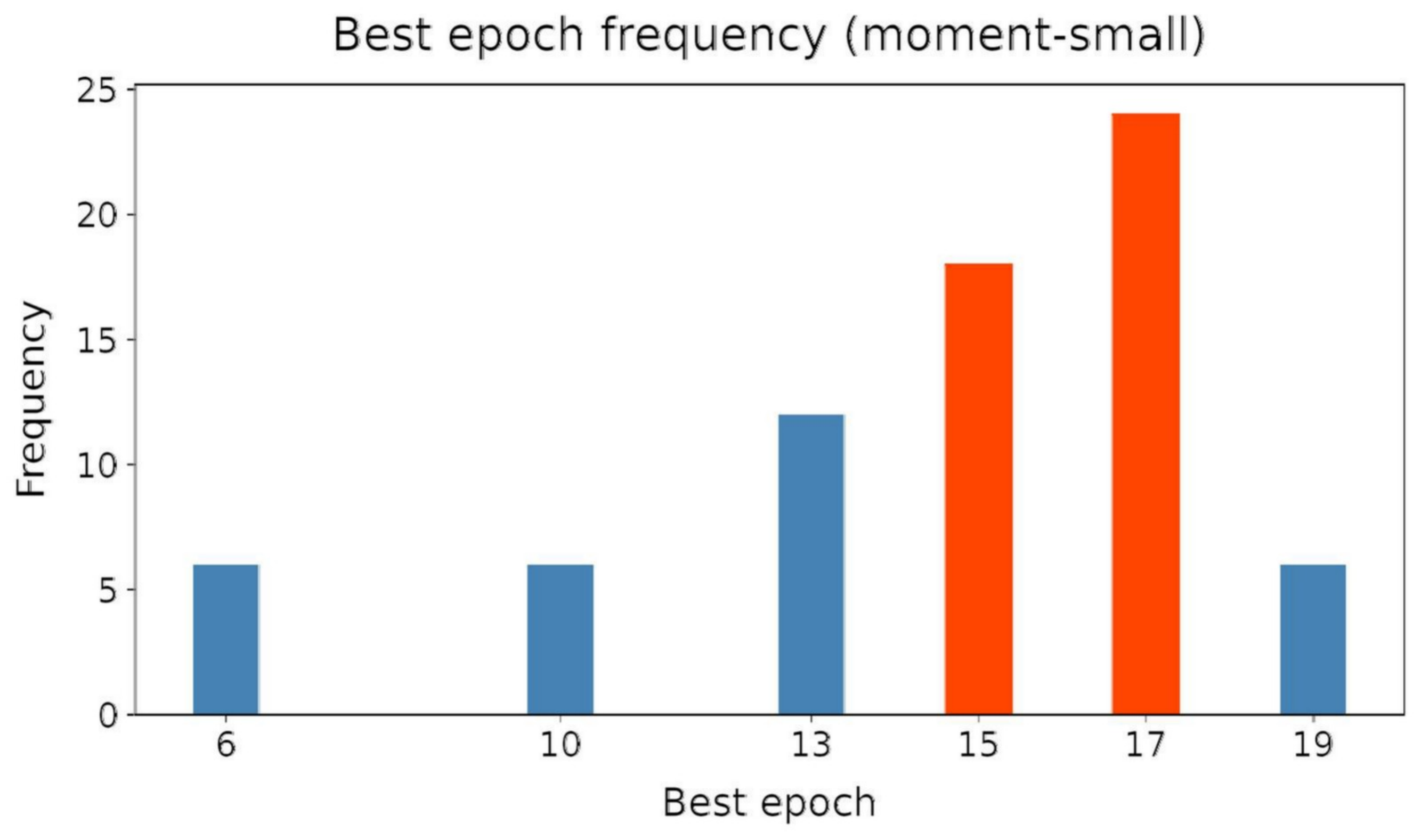}
    \caption{From top to botton, the linear correlation matrices for MOMENT-small.} 
    \label{fig:best_epochs:small}
\end{figure}

\begin{figure}[!htb]
    \centering
    \includegraphics[width=0.8\linewidth, pagebox=artbox]{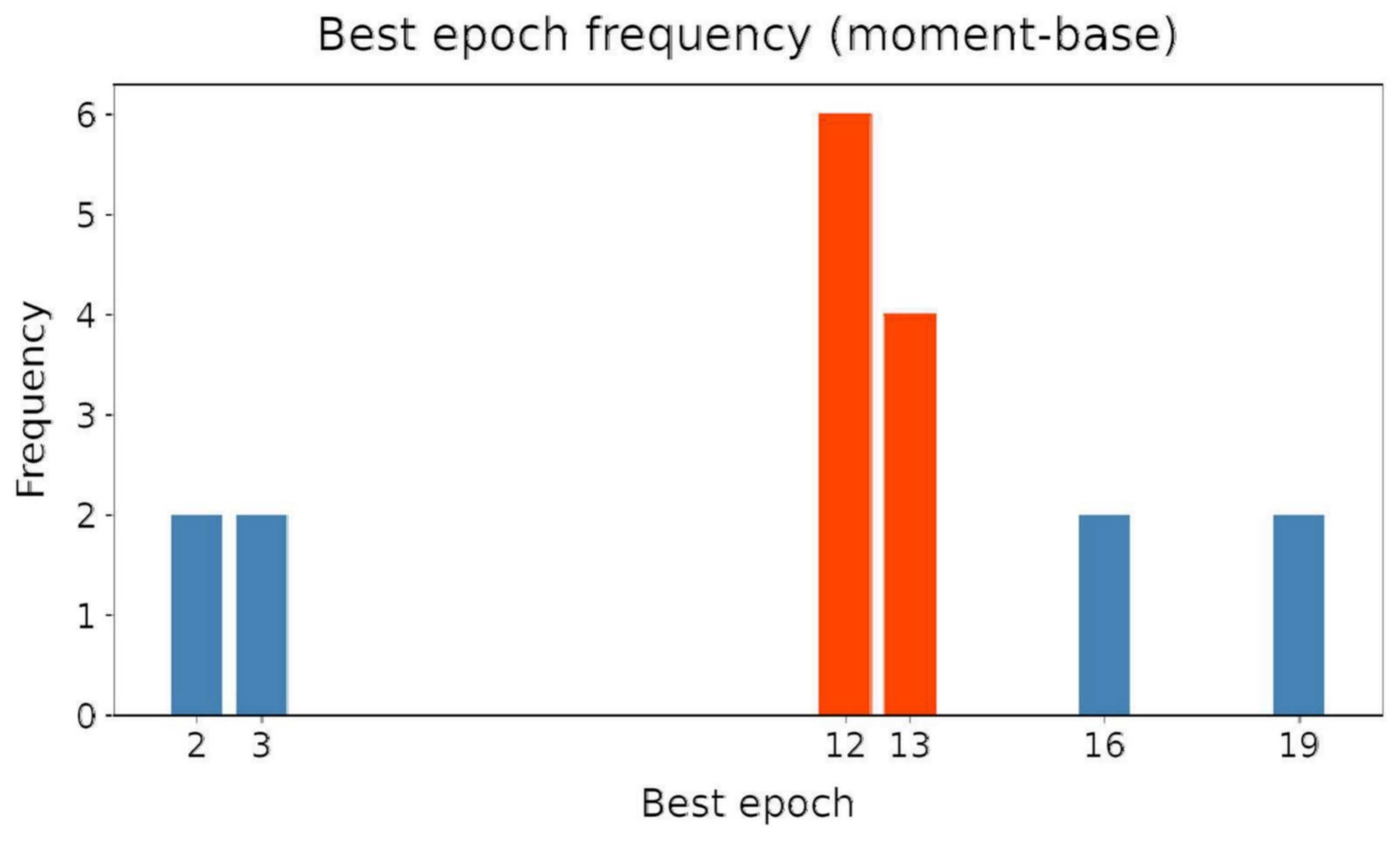}
    \caption{From top to botton, the linear correlation matrices for MOMENT-base} 
    \label{fig:best_epochs:base}
\end{figure}

\begin{figure}[!htb]
    \centering
    \includegraphics[width=0.8\linewidth, pagebox=artbox]{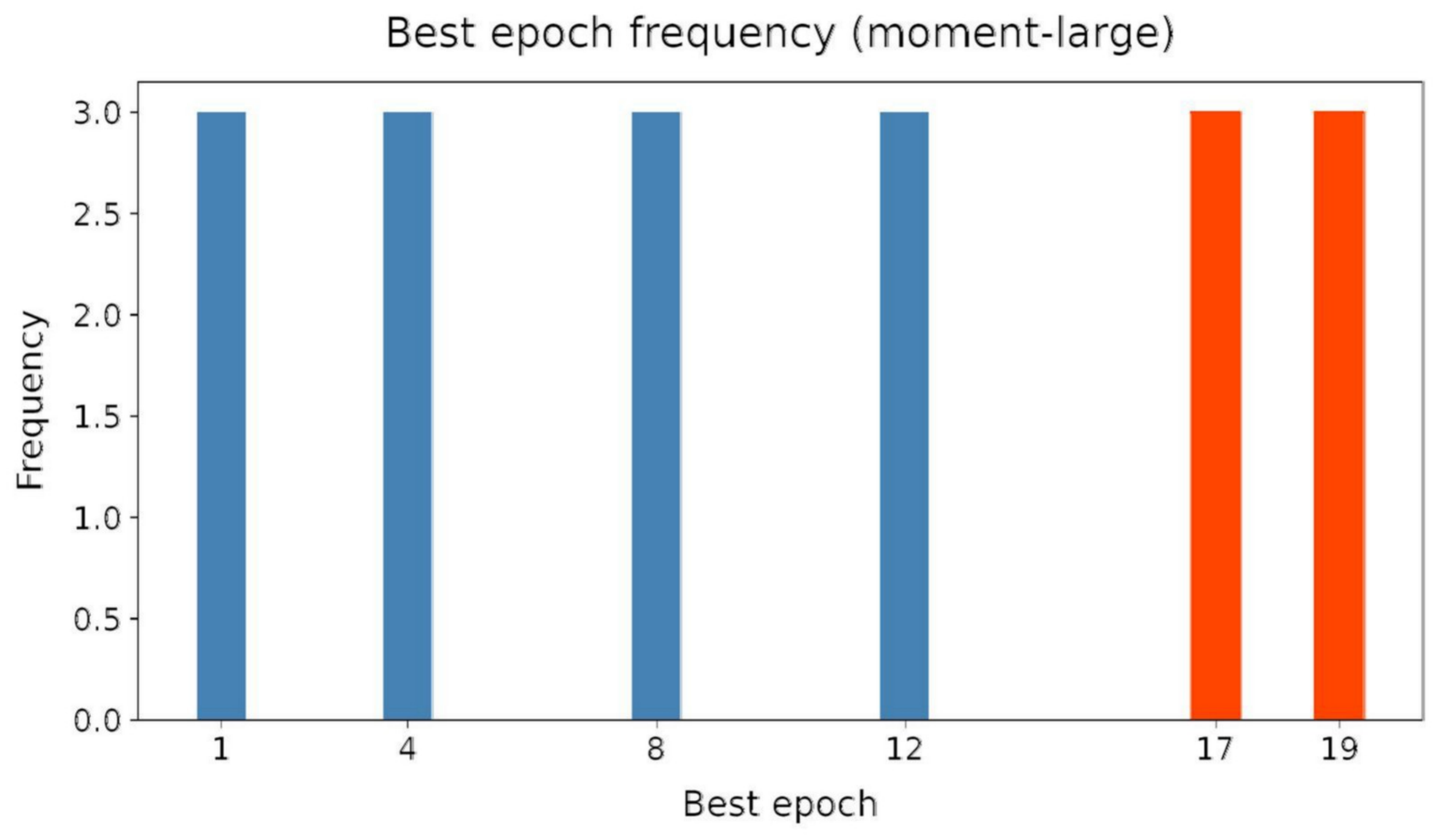}
    \caption{From top to botton, the linear correlation matrices for MOMENT-large model.} 
    \label{fig:best_epochs:large}
\end{figure}

Third, the frequency of the best epoch within the fine-tune loop for each version of the model (see~\customref{fig:best_epochs:small},~\ref{fig:best_epochs:base},~\ref{fig:best_epochs:large}) . This plot aids in the selection of a logical value for the number of epochs to ensure the best improvements with the lowest computation and time cost. This plot resalts in orange the two more repeated values, giving preference to the largest ones as they are the ones that give more assurance on getting the best model for each case configuration. 

\begin{figure}[!htb]
    \centering
    \includegraphics[width=0.8\linewidth, pagebox=artbox]{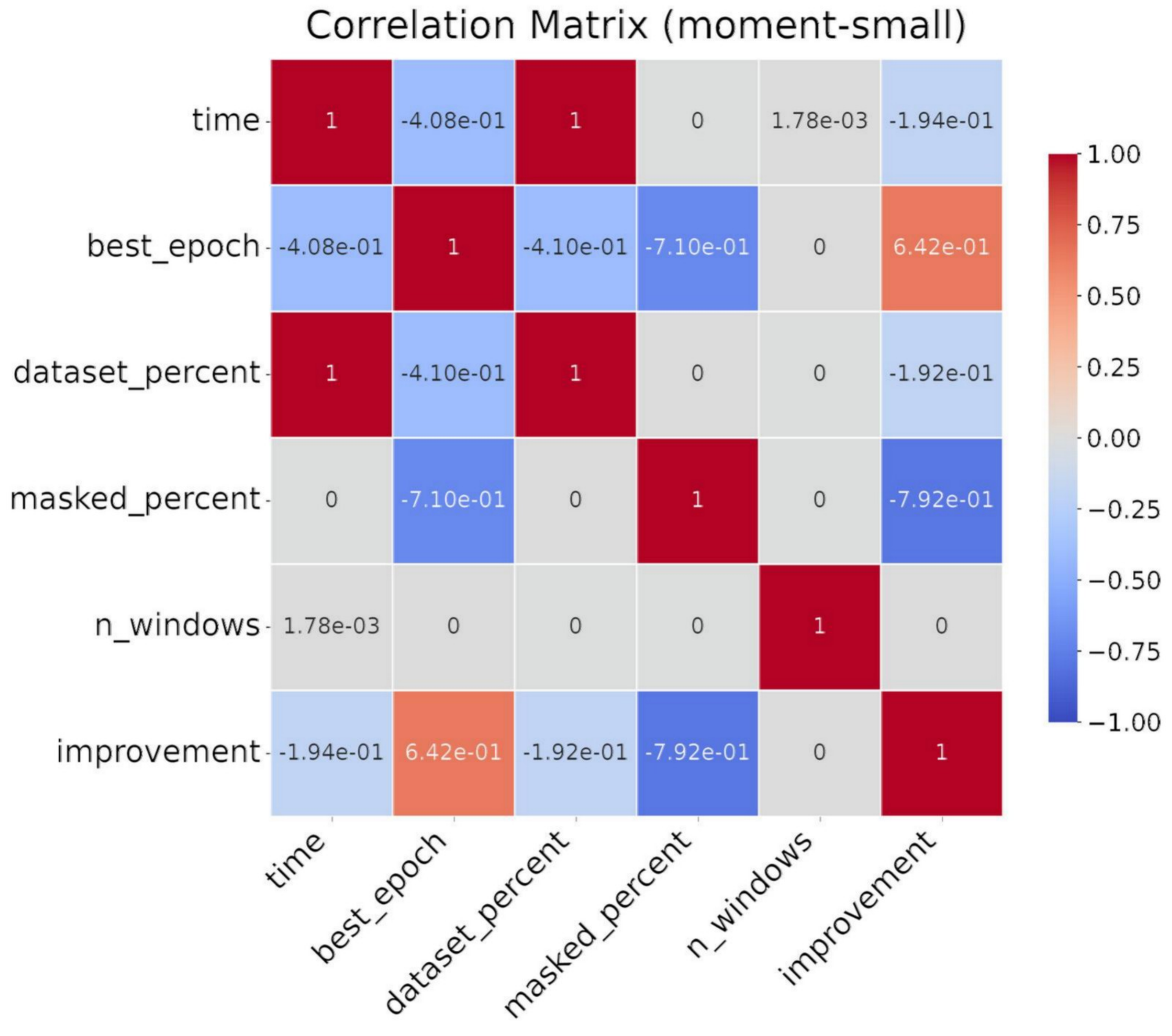}
    \caption{Experimentation parametres correlation matrix for MOMENT-small} 
    \label{fig:correlations:small}
\end{figure}

\begin{figure}[!htb]
    \centering    
    \includegraphics[width=0.8\linewidth, pagebox=artbox]{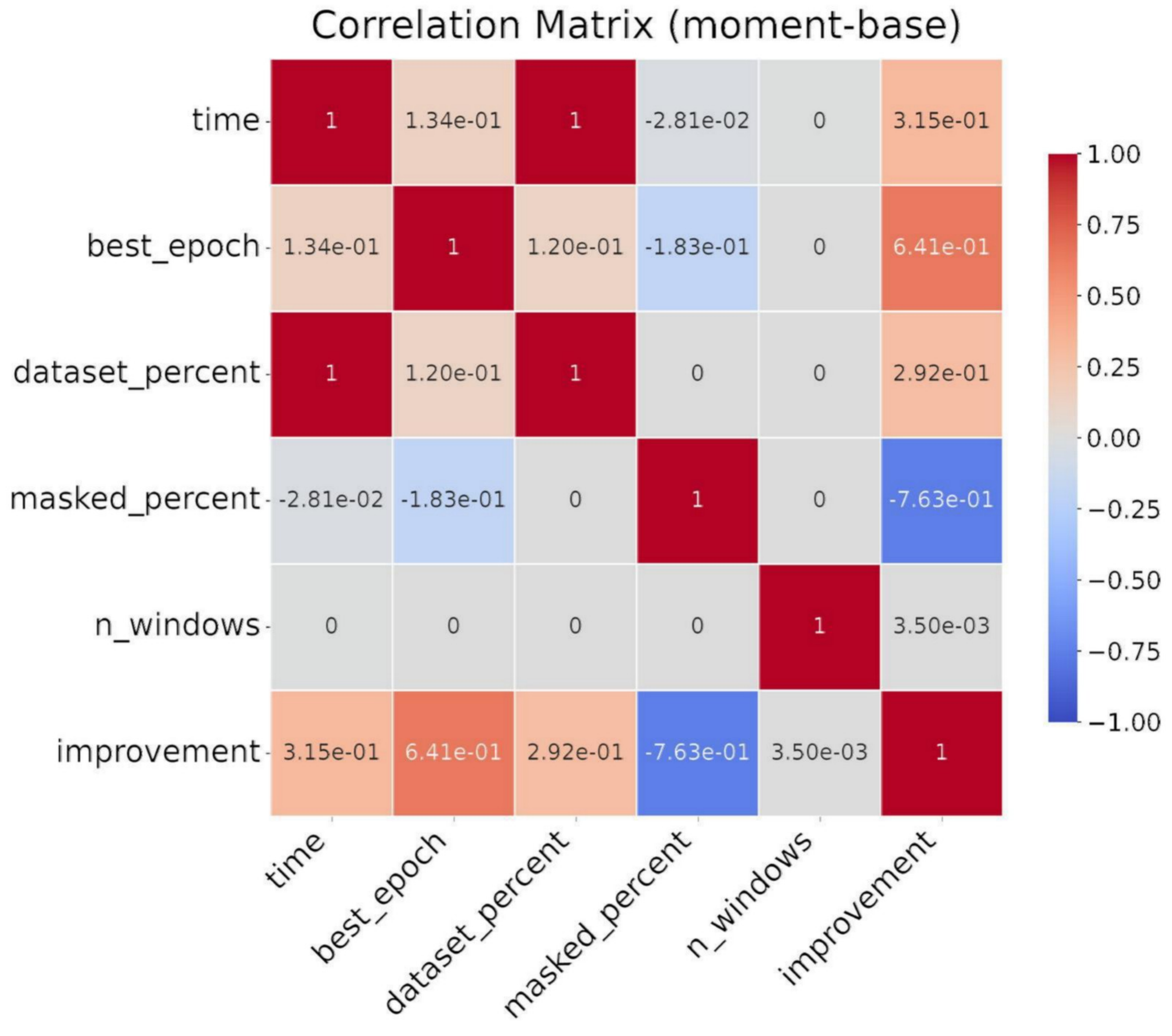}
    \caption{Experimentation parametres correlation matrix for MOMENT-base.} 
    \label{fig:correlations:base}
\end{figure}

\begin{figure}[!htb]
    \centering
    \includegraphics[width=0.8\linewidth, pagebox=artbox]{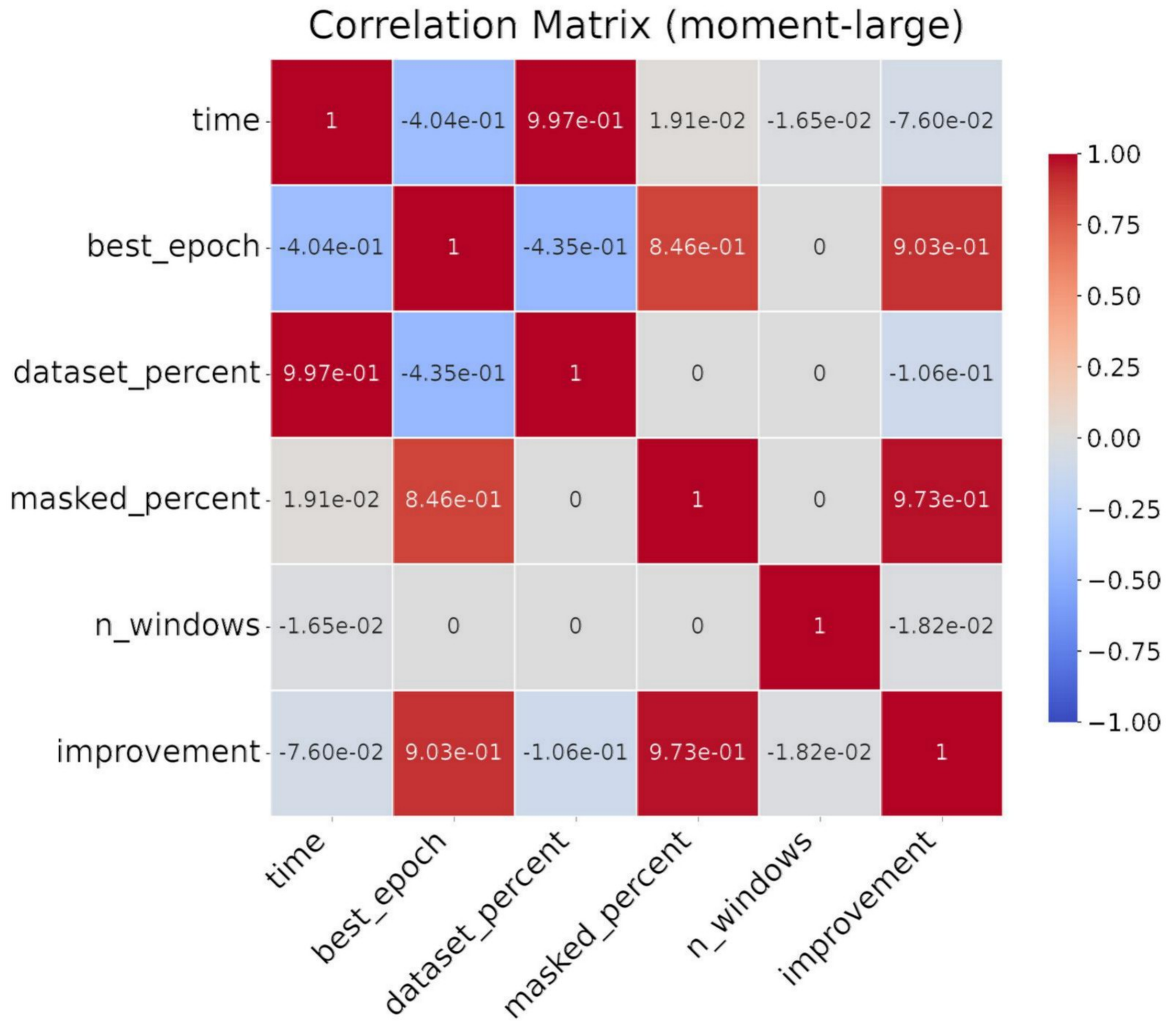}
    \caption{Experimentation parametres correlation matrix for MOMENT-large.} 
    \label{fig:correlations:large}
\end{figure}

Fourth, starting to interconnect all the parameters, the linear correlation matrices between them and the loss improvement~\customref{fig:correlations:small},~\customref{fig:correlations:base},~\customref{fig:correlations:large} for the three versions of MOMENT using tasks.


\FloatBarrier
\subsection{Interactive exploration auxiliar plot}

This section includes auxiliary plots that support the interactive analysis of the datasets.

\subsubsection{\texttt{S1} auxiliar plots}

First, the clusters for \texttt{S1} analysis, zooming in on each of them looking for the different segments in the time series.

\begin{figure}[H]
    \centering
    \includegraphics[width=0.9\linewidth, pagebox=artbox]{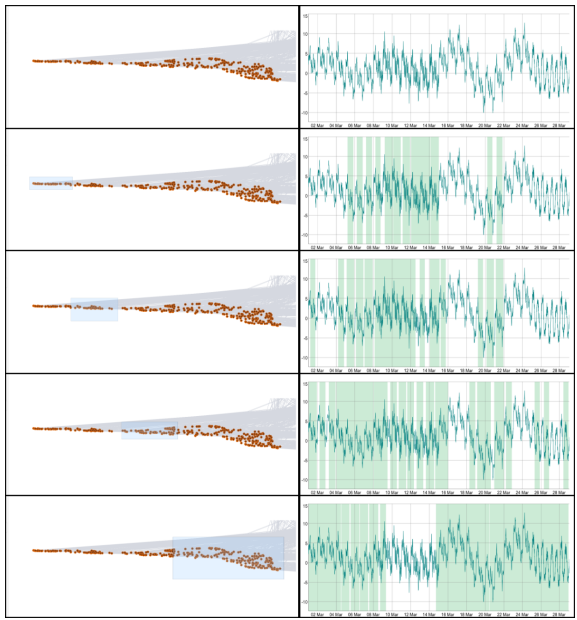}
    \caption{Cluster I. Execution of MOMENT-small for \texttt{S1}.}
    \label{fig:s1:moment:zeroshot:cluster_1}
\end{figure}

\begin{figure}[H]
    \centering
    \includegraphics[width=0.9\linewidth, pagebox=artbox]{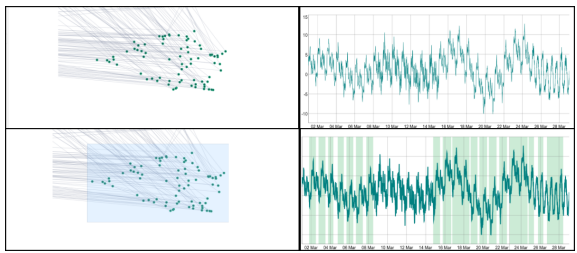}
    \caption{Cluster II. Execution of MOMENT-small  for \texttt{S1}.} 
    \label{fig:s1:moment:zeroshot:cluster_2}
\end{figure}

\begin{figure}[H]
    \centering
    \includegraphics[width=0.9\linewidth, pagebox=artbox]{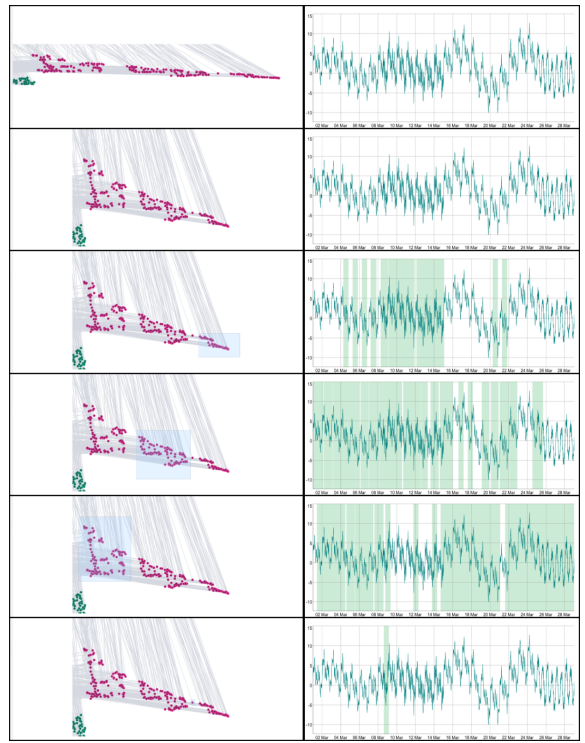}
    \caption{Cluster III. Execution of MOMENT-small for \texttt{S1}.} 
    \label{fig:s1:moment:zeroshot:cluster_3}
\end{figure}

\begin{figure}[H]
    \centering
    \includegraphics[width=1\linewidth, pagebox=artbox]{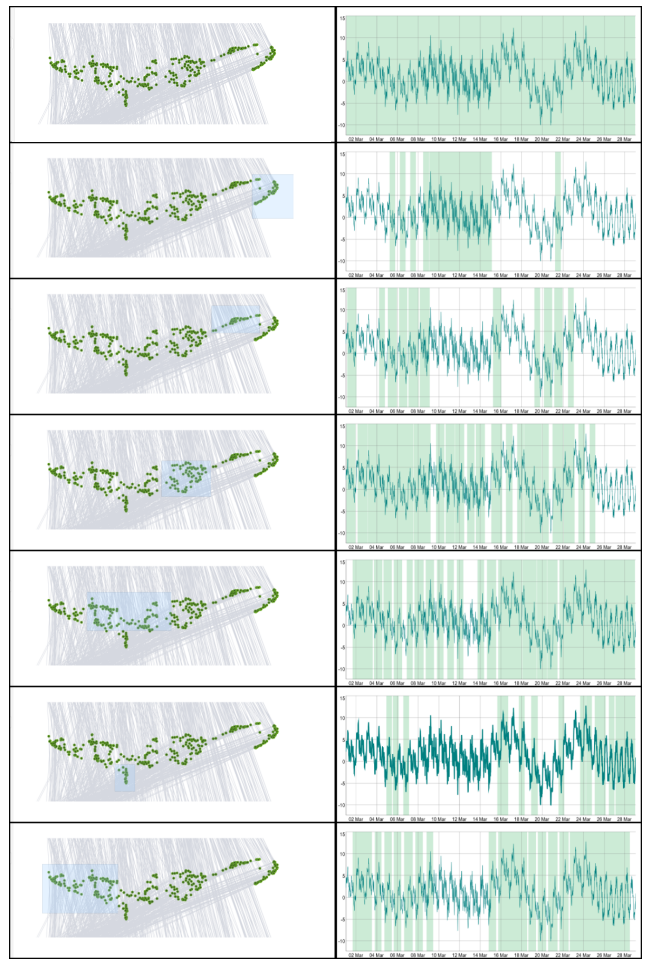}
    \caption{Cluster IV. Execution of MOMENT-small for \texttt{S1}.} 
    \label{fig:s1:moment:zeroshot:cluster_4}
\end{figure}

\begin{figure}[H]
    \centering
    \includegraphics[width=1\linewidth, pagebox=artbox]{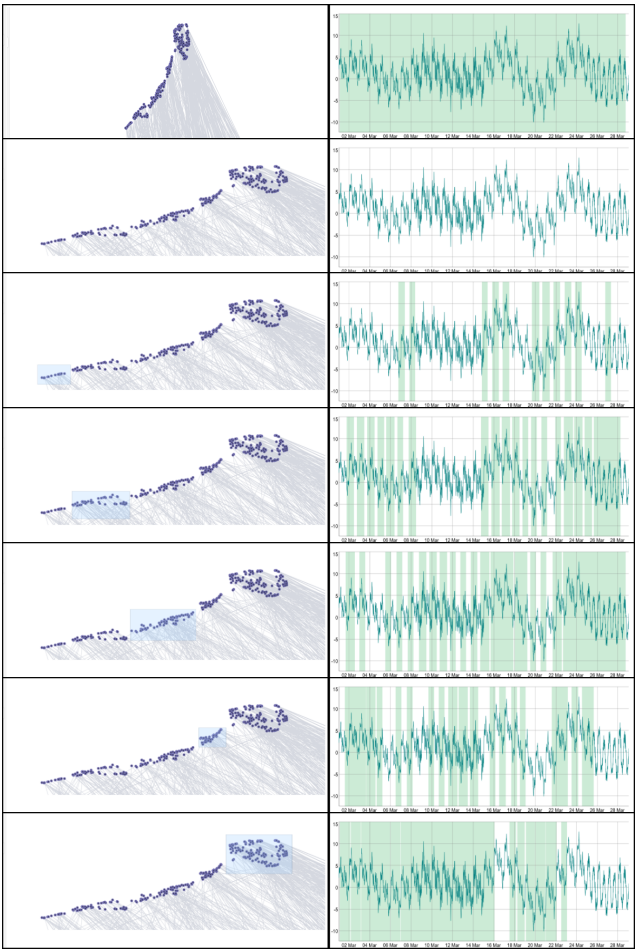}
    \caption{Cluster V. Execution of MOMENT-small for \texttt{S1}.} 
    \label{fig:s1:moment:zeroshot:cluster_5}
\end{figure}

Figures~\ref{fig:s1:moment:zeroshot:cluster_1}, ~\ref{fig:s1:moment:zeroshot:cluster_2},~\ref{fig:s1:moment:zeroshot:cluster_3},~\ref{fig:s1:moment:zeroshot:cluster_4}, and~\ref{fig:s1:moment:zeroshot:cluster_5} show the visualizations for the different clusters of the embedding space when executing MOMENT-small for \texttt{S1}.

\begin{figure}[!htb]
    \centering
    \includegraphics[width=1\linewidth, pagebox=artbox]{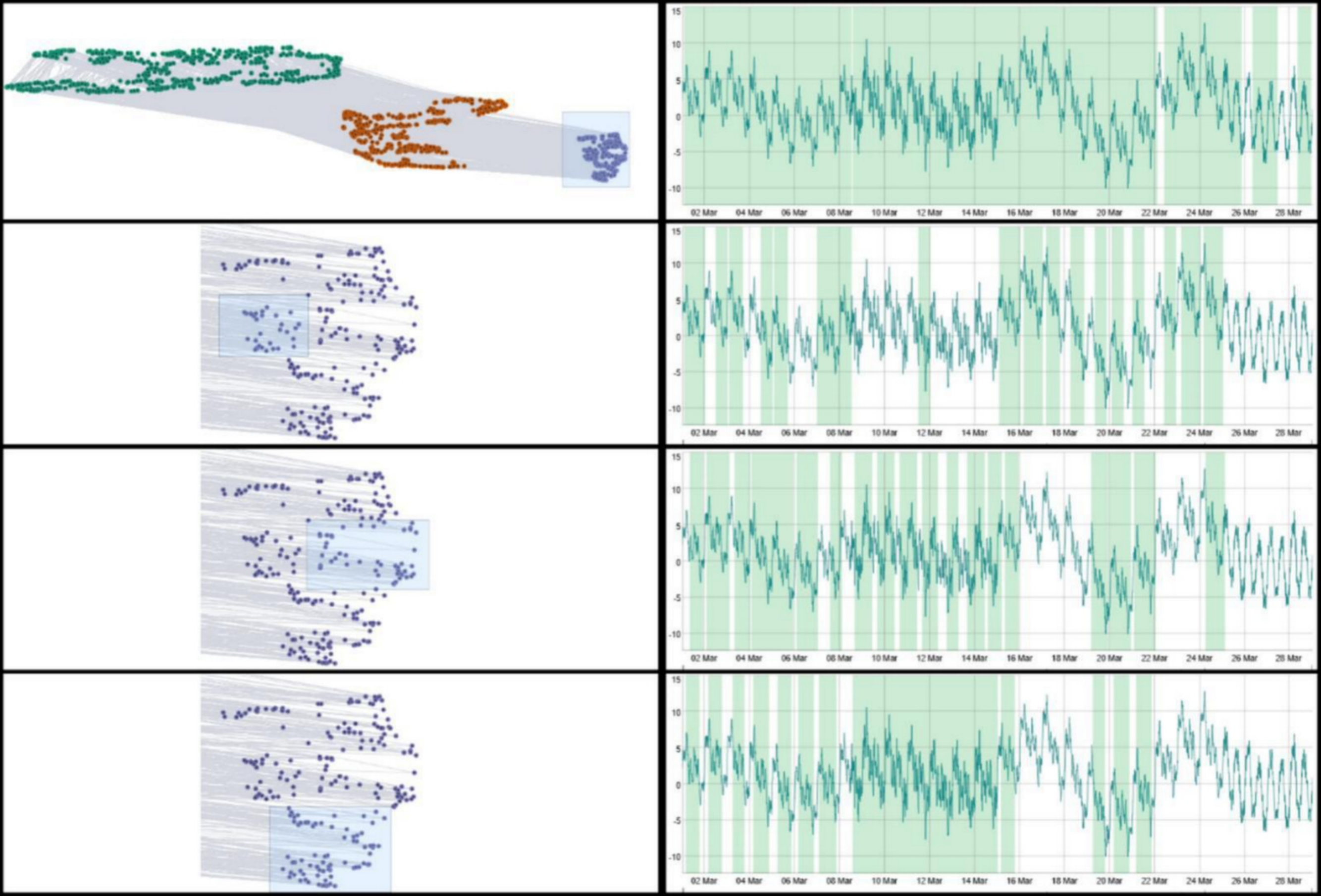}
    \caption{Cluster I: execution of MOMENT-base for \texttt{S1}.} 
    \label{fig:s1:moment:base:cluster_1}
\end{figure}

\begin{figure}[!htb]
    \centering
    \includegraphics[width=1\linewidth, pagebox=artbox]{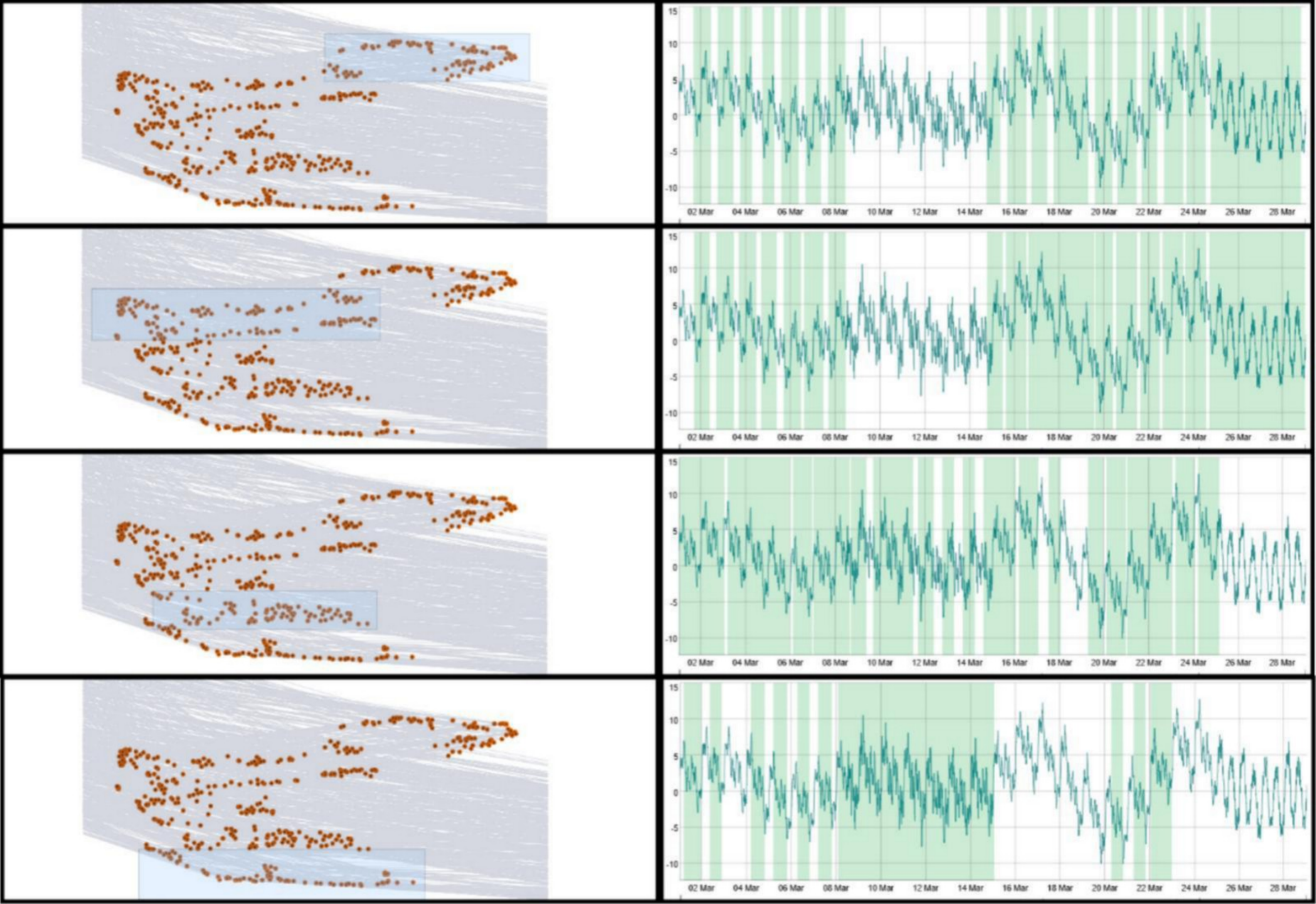}
    \caption{Cluster II: execution of MOMENT-base for \texttt{S1}.} 
    \label{fig:s1:moment:base:cluster_2}
\end{figure}

\begin{figure}[!htb]
    \centering
    \includegraphics[width=1\linewidth, pagebox=artbox]{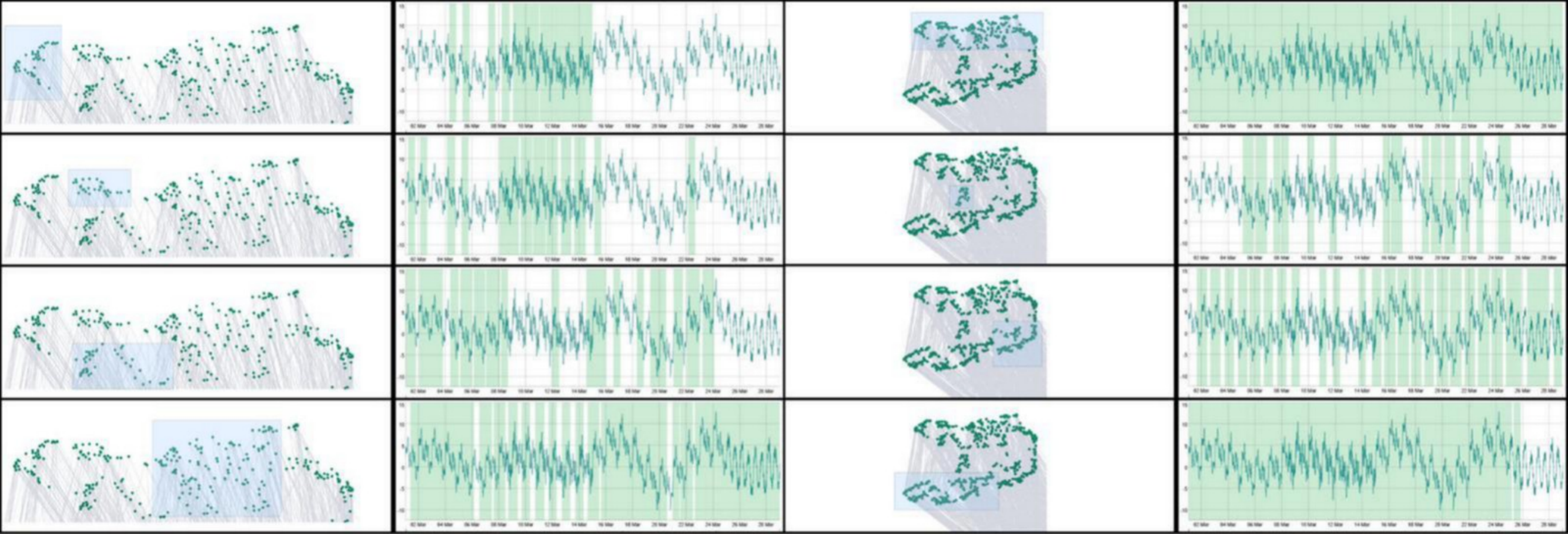}
    \caption{Cluster III: execution of MOMENT-base for \texttt{S1}.} 
    \label{fig:s1:moment:base:cluster_3}
\end{figure}

Figures~\ref{fig:s1:moment:base:cluster_1}, ~\ref{fig:s1:moment:base:cluster_2},~\ref{fig:s1:moment:base:cluster_3}, show the visualizations for the different clusters of the embedding space when executing MOMENT-small for \texttt{S1}.

\FloatBarrier
\subsubsection{\texttt{S2} auxiliar plots}

\begin{figure}[H]
    \centering
    \includegraphics[width=1\linewidth, pagebox=artbox]{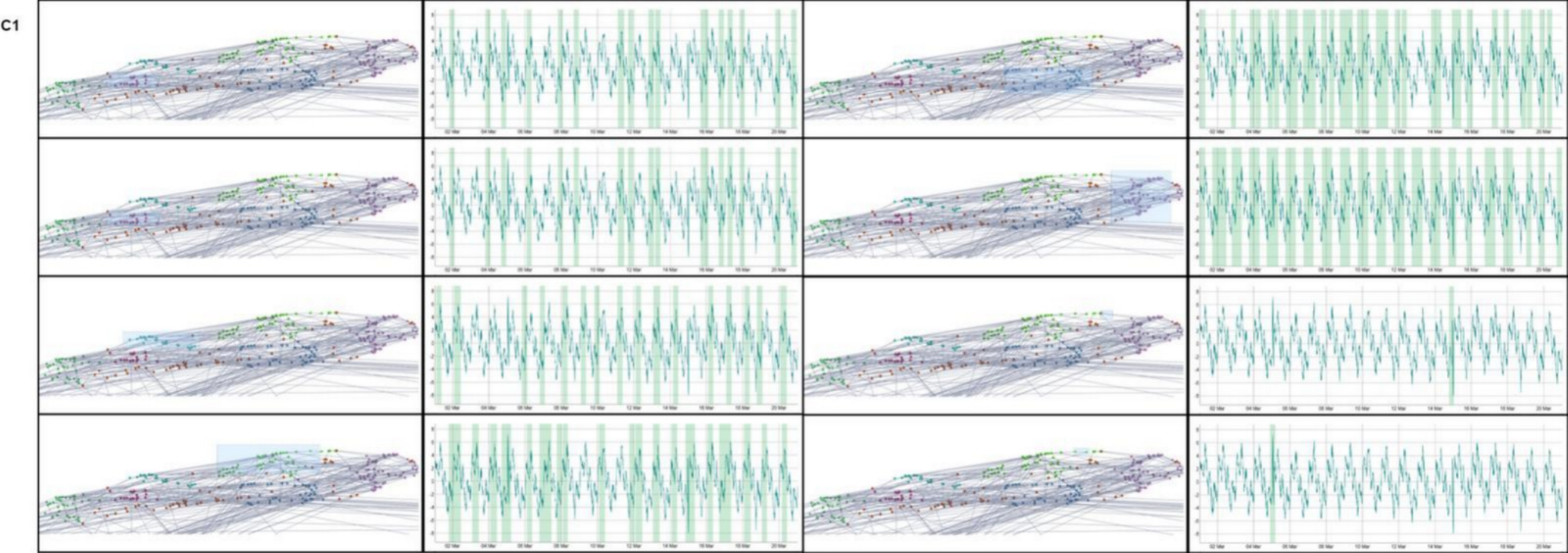}
    \caption{Cluster I. Analysis of the first cluster of the projections plot for the MOMENT-small zero-shot version of the model for \texttt{S2}.} 
    \label{fig:s2:moment:small:c1}
\end{figure}

\begin{figure}[H]
    \centering
    \includegraphics[width=1\linewidth, pagebox=artbox]{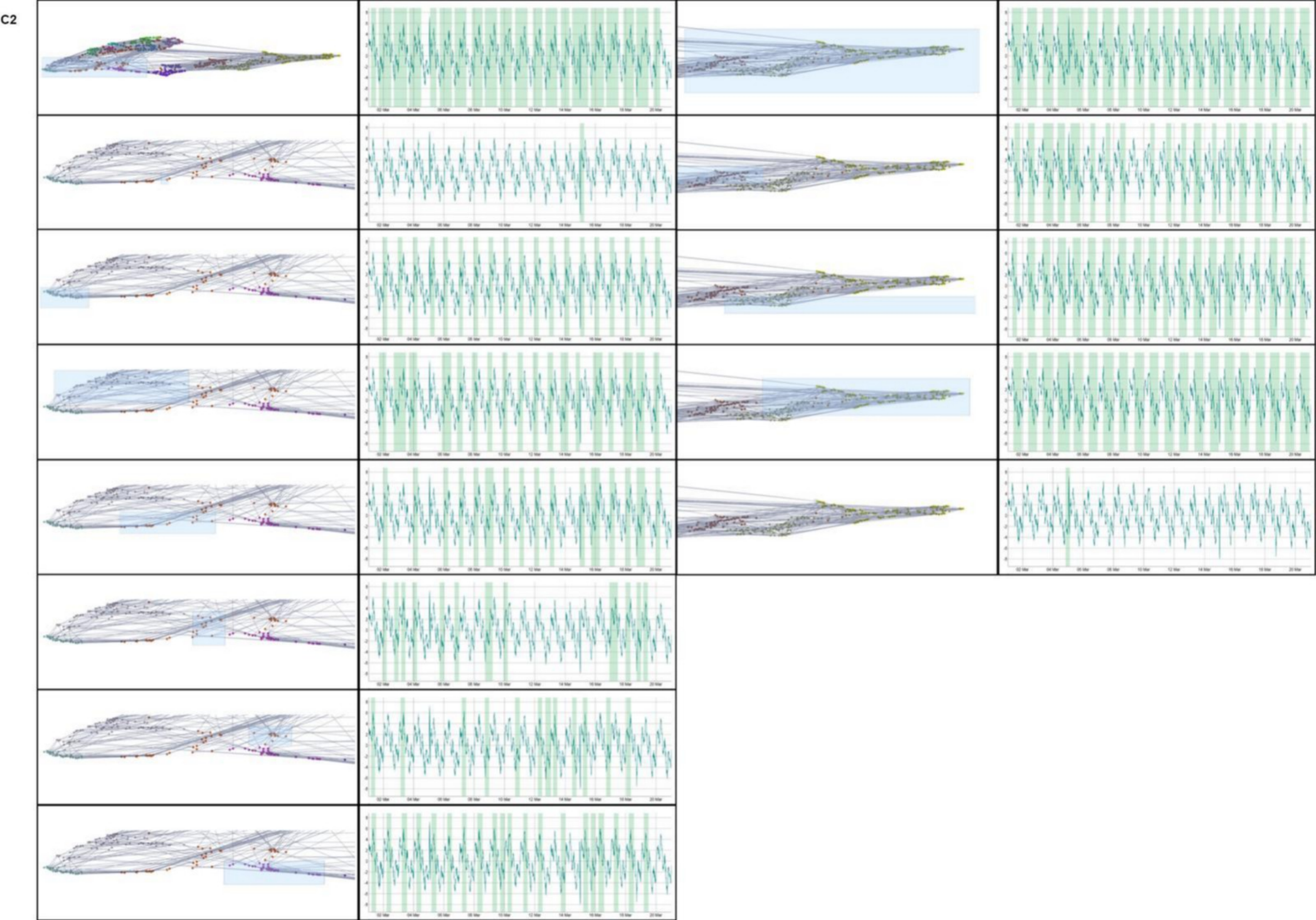}
    \caption{Cluster II. Analysis of the second cluster of the projections plot for the MOMENT-small zero-shot version of the model for \texttt{S2}.} 
    \label{fig:s2:moment:small:c2}
\end{figure}

Figures~\ref{fig:s2:moment:small:c1} and ~\ref{fig:s2:moment:small:c2} show the two clusters of the execution of MOMENT-small in the zero-shot verison

\begin{figure}[H]
    \centering
    \includegraphics[width=1\linewidth, pagebox=artbox]{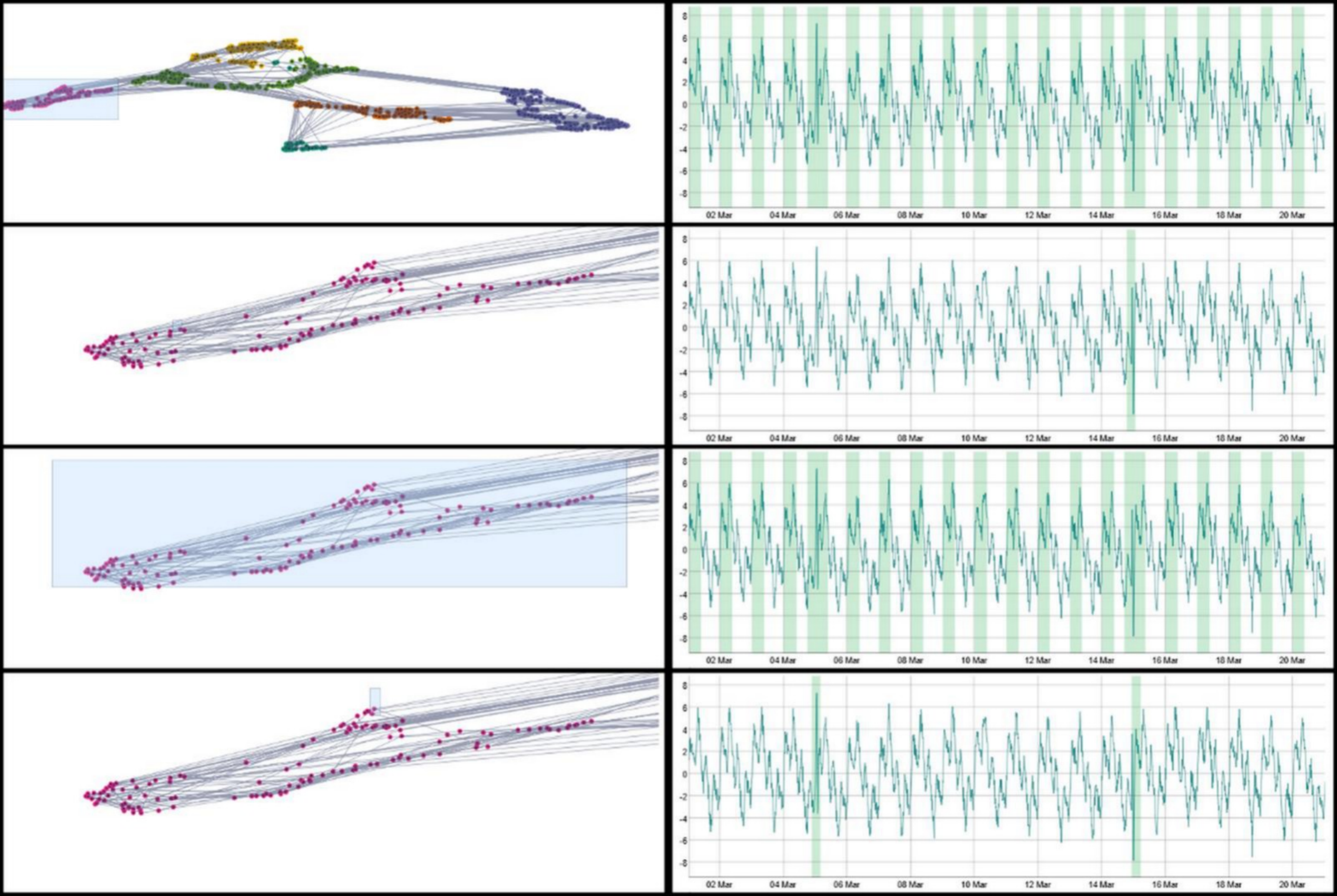}
    \caption{Cluster I. Analysis of the zero-shot version of MOMENT-large applied to \texttt{S2}. Contains both anomalies.} 
    \label{fig:s2:moment:large:c1}
\end{figure}

\begin{figure}[H]
    \centering
    \includegraphics[width=1\linewidth, pagebox=artbox]{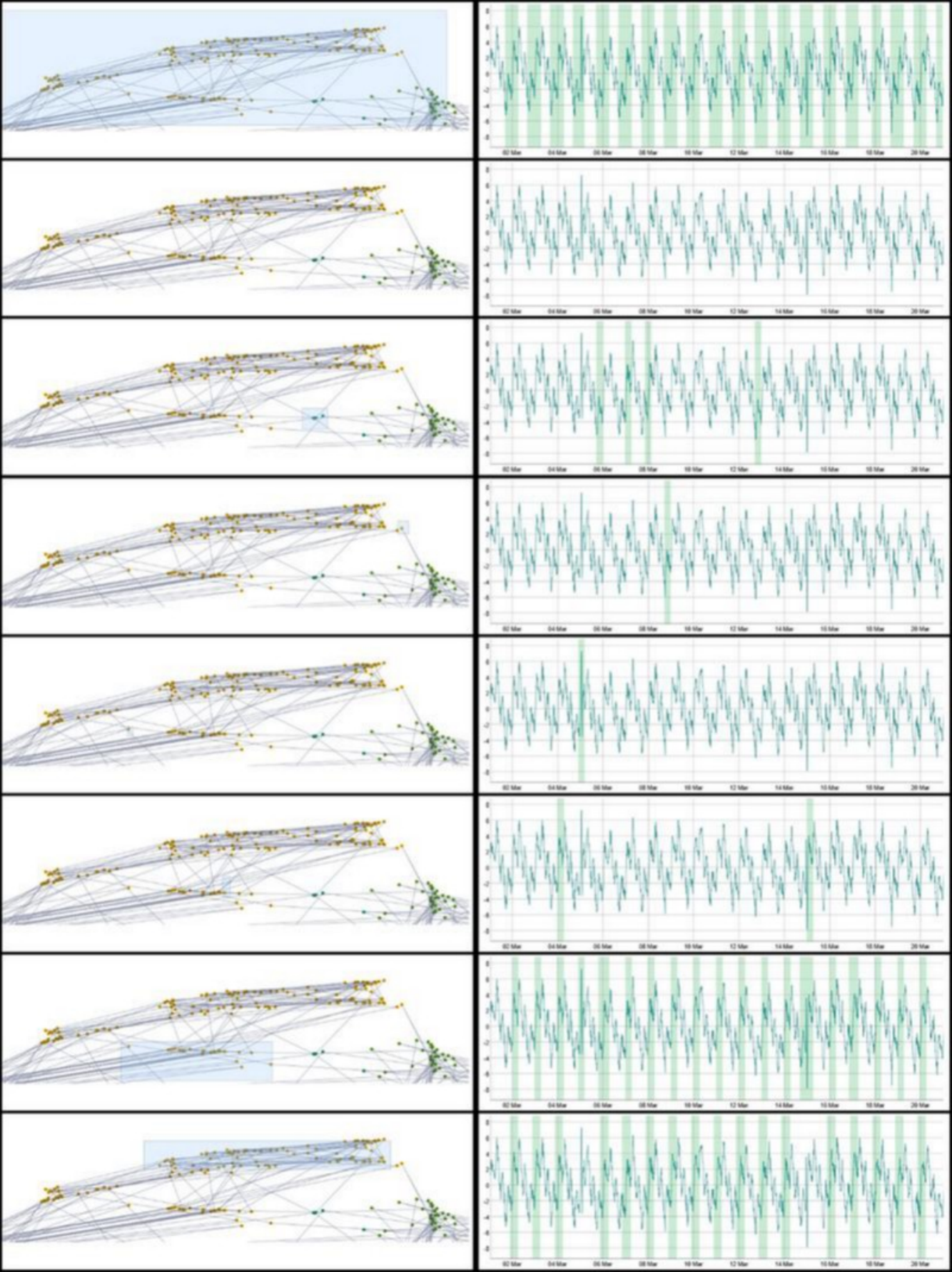}
    \caption{Cluster II. Analysis of the the embeddings projections of the zero-shot version of MOMENT-large applied to \texttt{S2}. Contains both anomalies.} 
    \label{fig:s2:moment:large:c2}
\end{figure}

\begin{figure}[H]
    \centering
    \includegraphics[width=1\linewidth, pagebox=artbox]{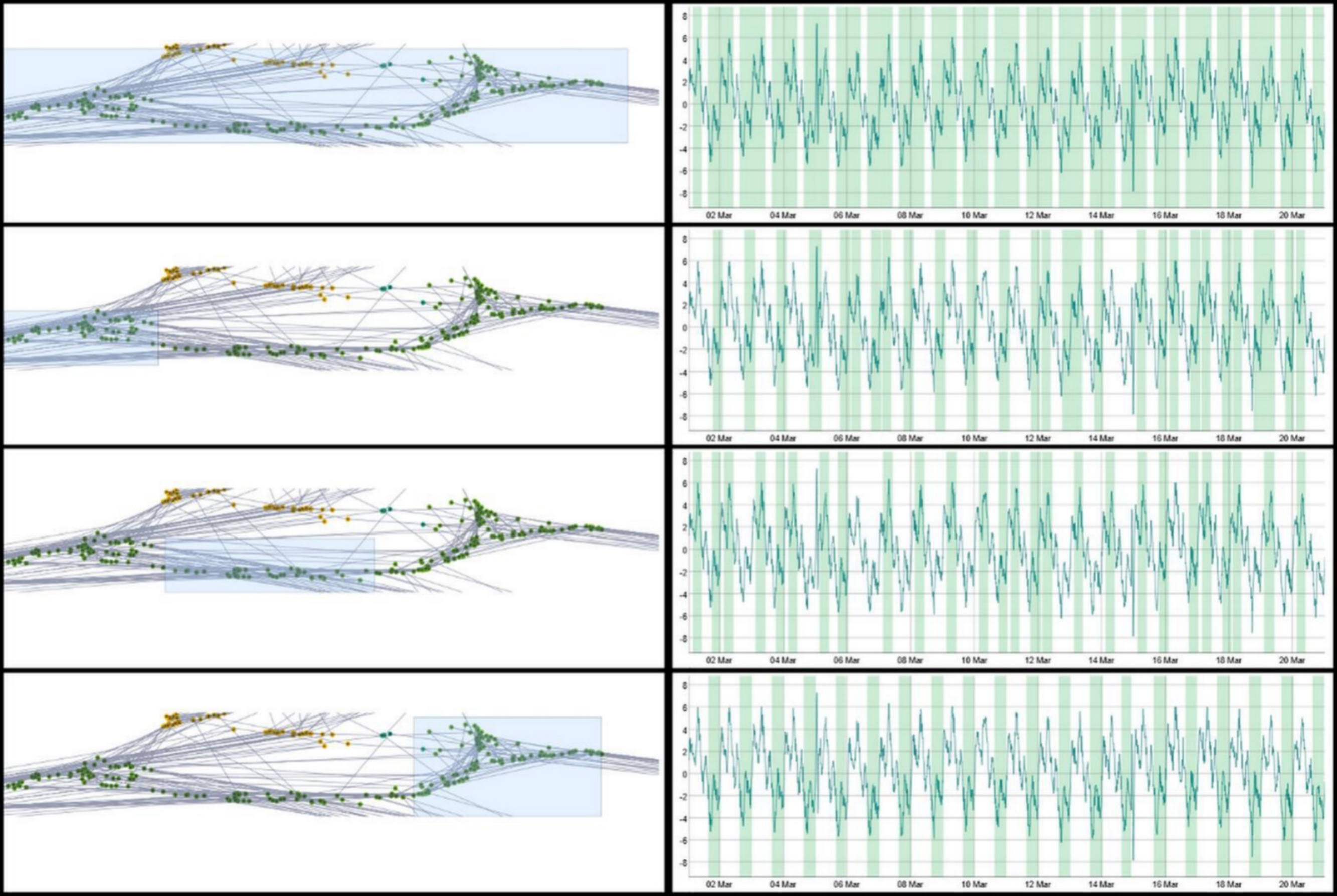}
    \caption{Cluster III. Analysis of the embeddings projections of the zero-shot version of MOMENT-large applied to \texttt{S2}.} 
    \label{fig:s2:moment:large:c3}
\end{figure}

\begin{figure}[H]
    \centering
    \includegraphics[width=1\linewidth, pagebox=artbox]{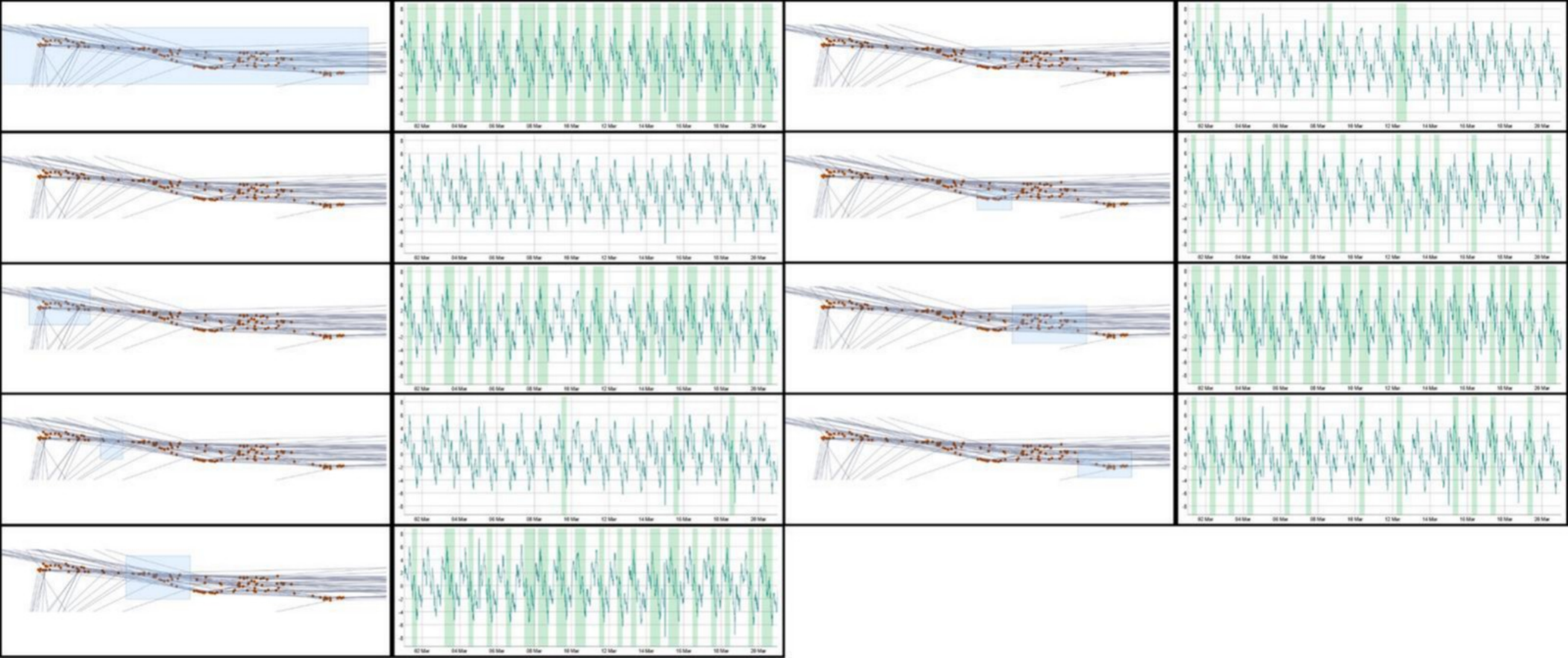}
    \caption{Cluster IV. Analysis of the embeddings projections of the zero-shot version of MOMENT-large applied to \texttt{S2}.} 
    \label{fig:s2:moment:large:c4}
\end{figure}

\begin{figure}[H]
    \centering
    \includegraphics[width=1\linewidth, pagebox=artbox]{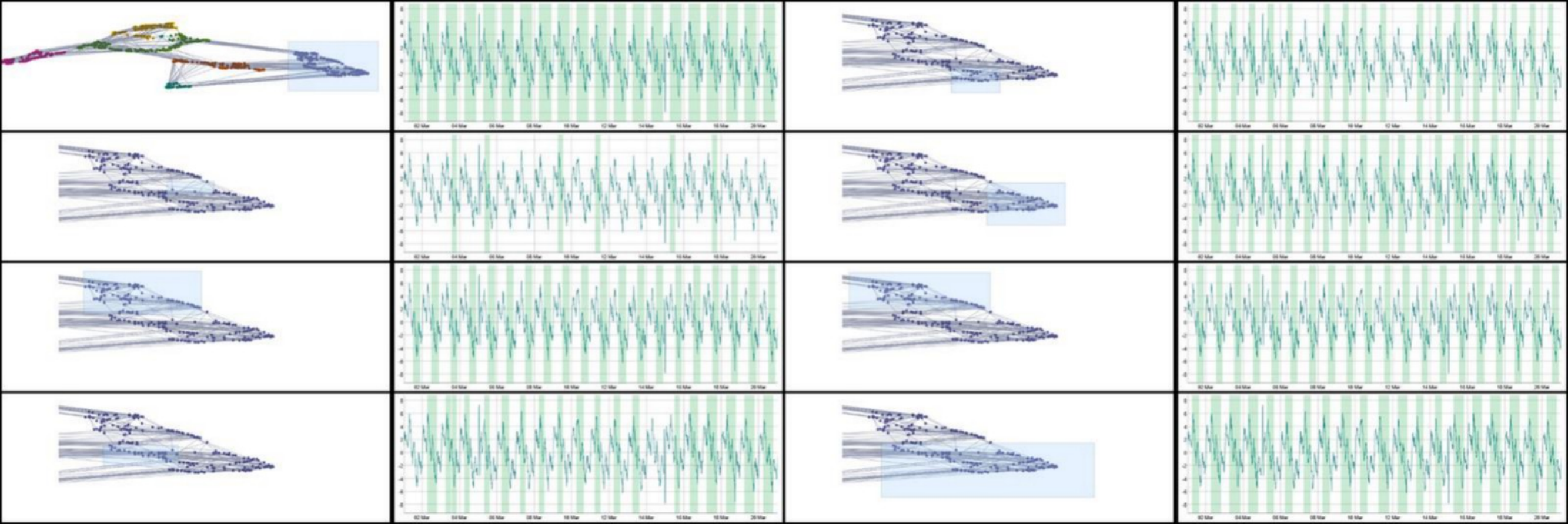}
    \caption{Cluster V. Analysis of the embeddings projections of the zero-shot version of MOMENT-large applied to \texttt{S2}.} 
    \label{fig:s2:moment:large:c5}
\end{figure}

Figures~\ref{fig:s2:moment:large:c1},~\ref{fig:s2:moment:large:c2},~\ref{fig:s2:moment:large:c4},~\ref{fig:s2:moment:large:c3},~\ref{fig:s2:moment:large:c5} show the five clusters of the execution of MOMENT-large in the zero-shot version.


\FloatBarrier
\subsubsection{\texttt{S3} auxiliar plots}
Auxiliar plots for the visual analysis of \texttt{S3} using the diferent versions of MOMENT. 

\begin{figure}[H] 
    \centering
    \includegraphics[width=1\linewidth, pagebox=artbox]{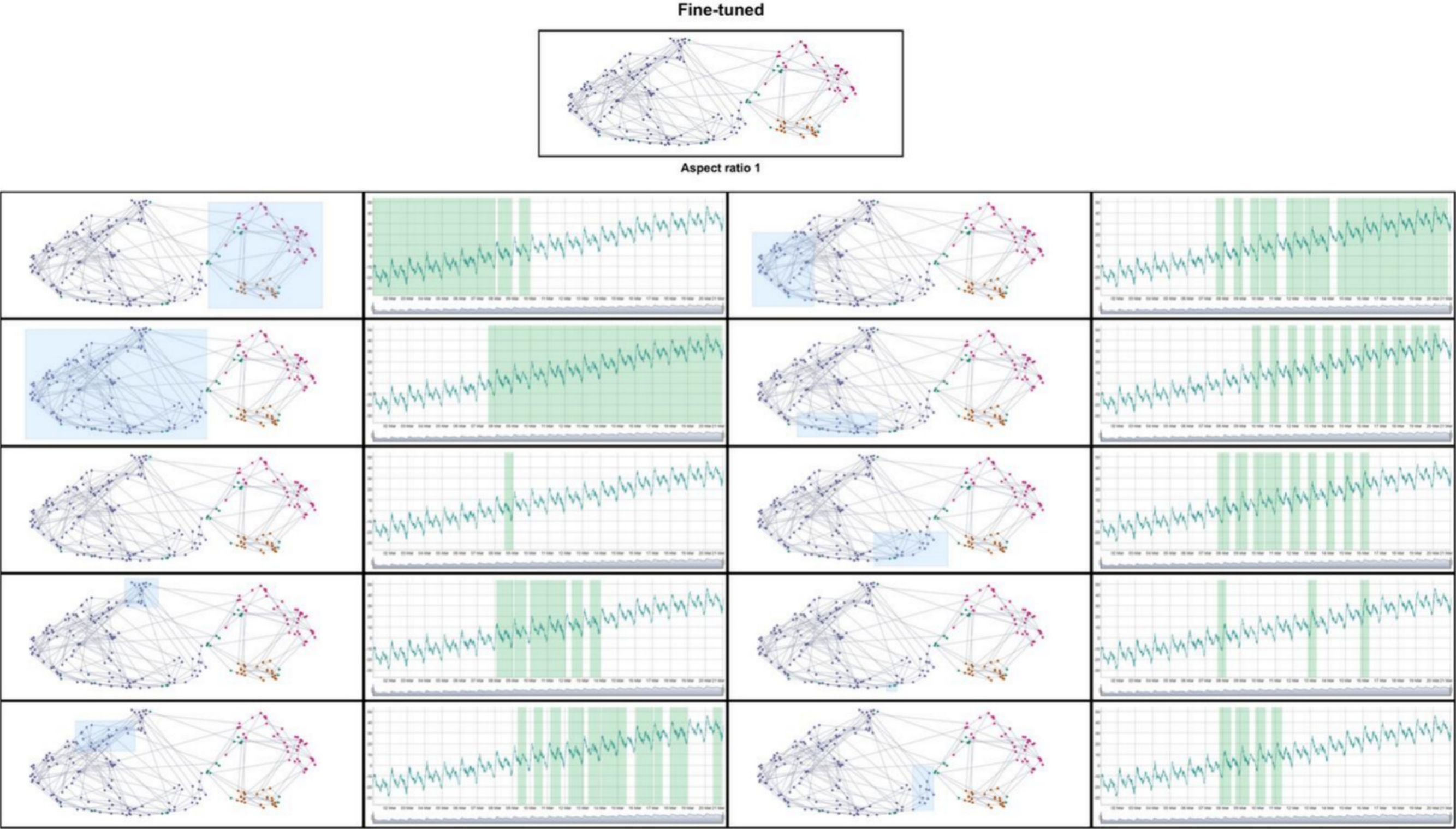}
    \caption{Embeddings projections of the zero-shot version of MOMENT-base applied to \texttt{S3}.} 
    \label{fig:s3:moment:base:zero-shot}
\end{figure}

\begin{figure}[H]
    \centering
    \includegraphics[width=1\linewidth, pagebox=artbox]{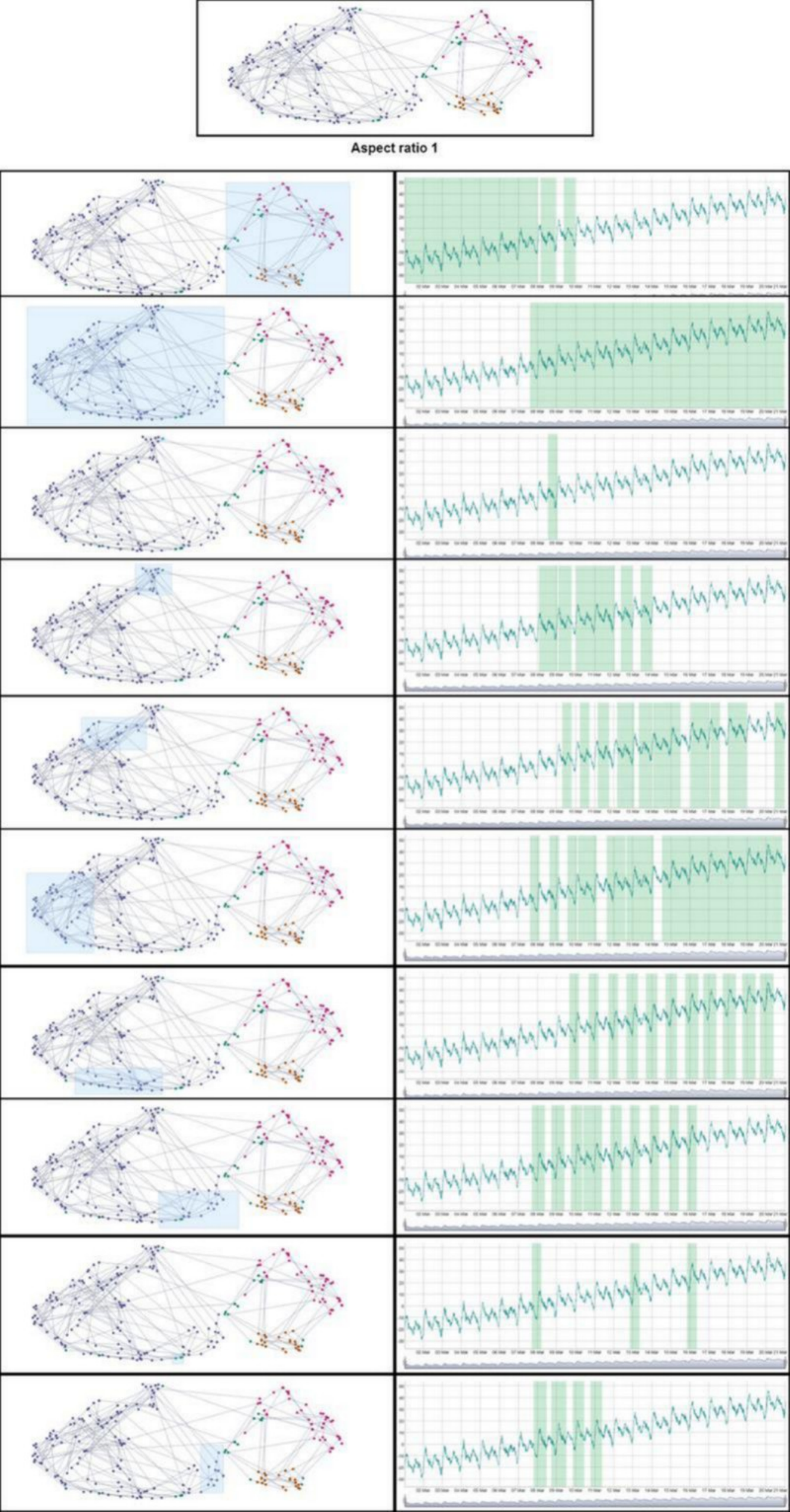}
    \caption{Embeddings projections of the fine-tuned version of MOMENT-base applied to \texttt{S3}.} 
    \label{fig:s3:moment:base:finetuned}
\end{figure}

\begin{figure}[H]
    \centering
    \includegraphics[width=1\linewidth, pagebox=artbox]{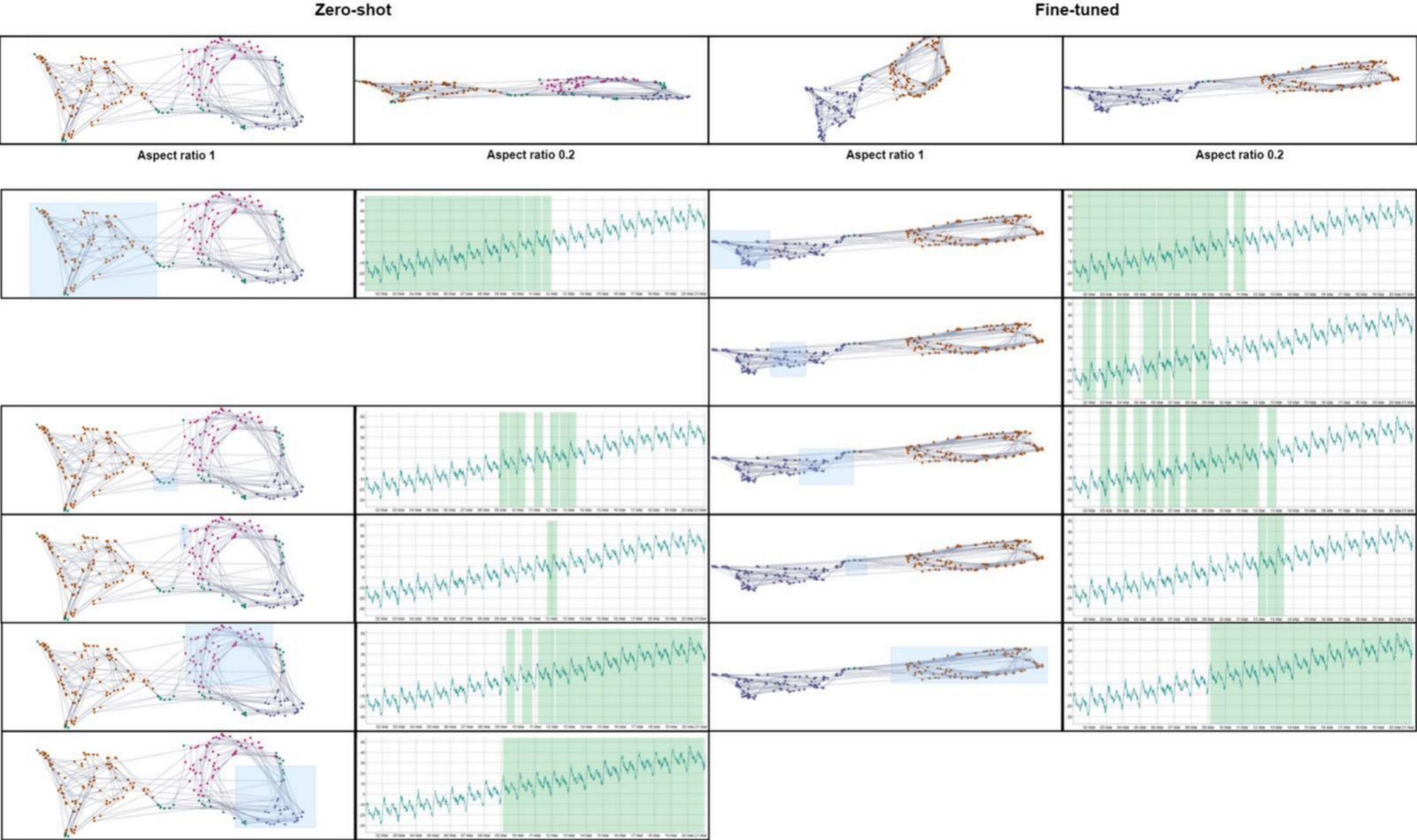}
    \caption{Embeddings projections of the fine-tuned version of MOMENT-large applied to \texttt{S3}.} 
    \label{fig:s3:moment:large}
\end{figure}

Figures~\ref{fig:s3:moment:base:zero-shot} and~\ref{fig:s3:moment:base:finetuned} show the visualization resulting from the inference with zero-shot and fine-tuned versions of MOMENT-base while Fig.~\ref{fig:s3:moment:large} contains the plot for the MOMENT-large version.

\FloatBarrier
\subsubsection{\texttt{M-Toy} auxiliar plots}
\begin{figure}[H]
    \centering
    \includegraphics[width=1\linewidth, pagebox=artbox]{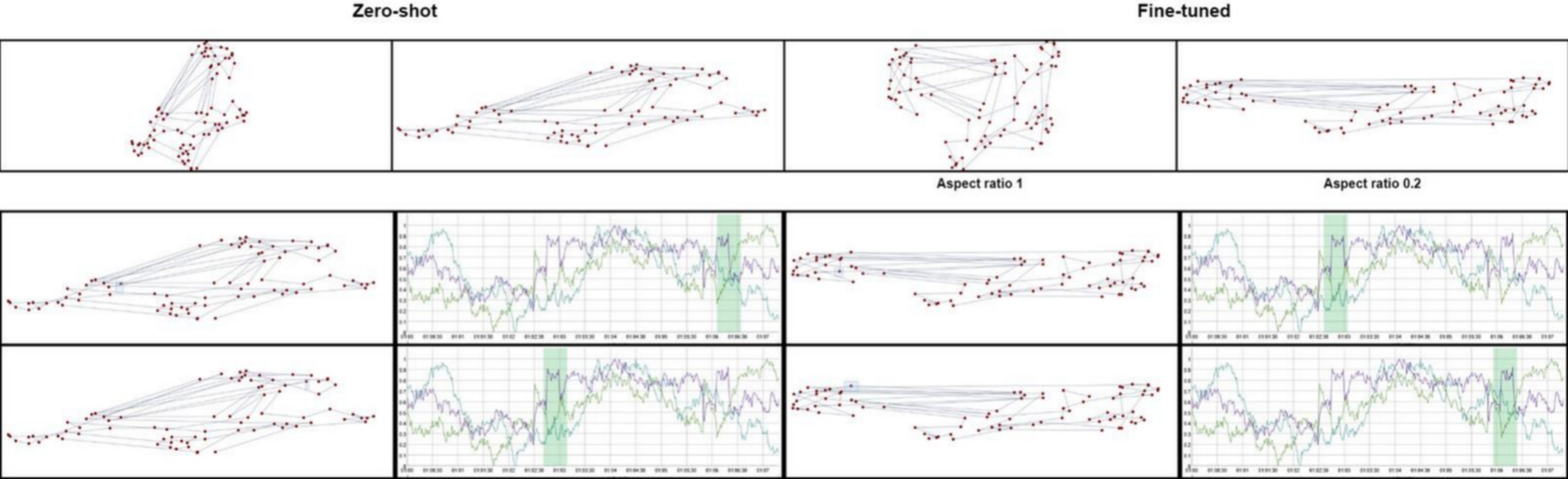}
    \caption{Embeddings projections of MOMENT-base applied to \texttt{M-Toy}.} 
    \label{fig:toy:moment:base}
\end{figure}

\begin{figure}[H]
    \centering
    \includegraphics[width=1\linewidth, pagebox=artbox]{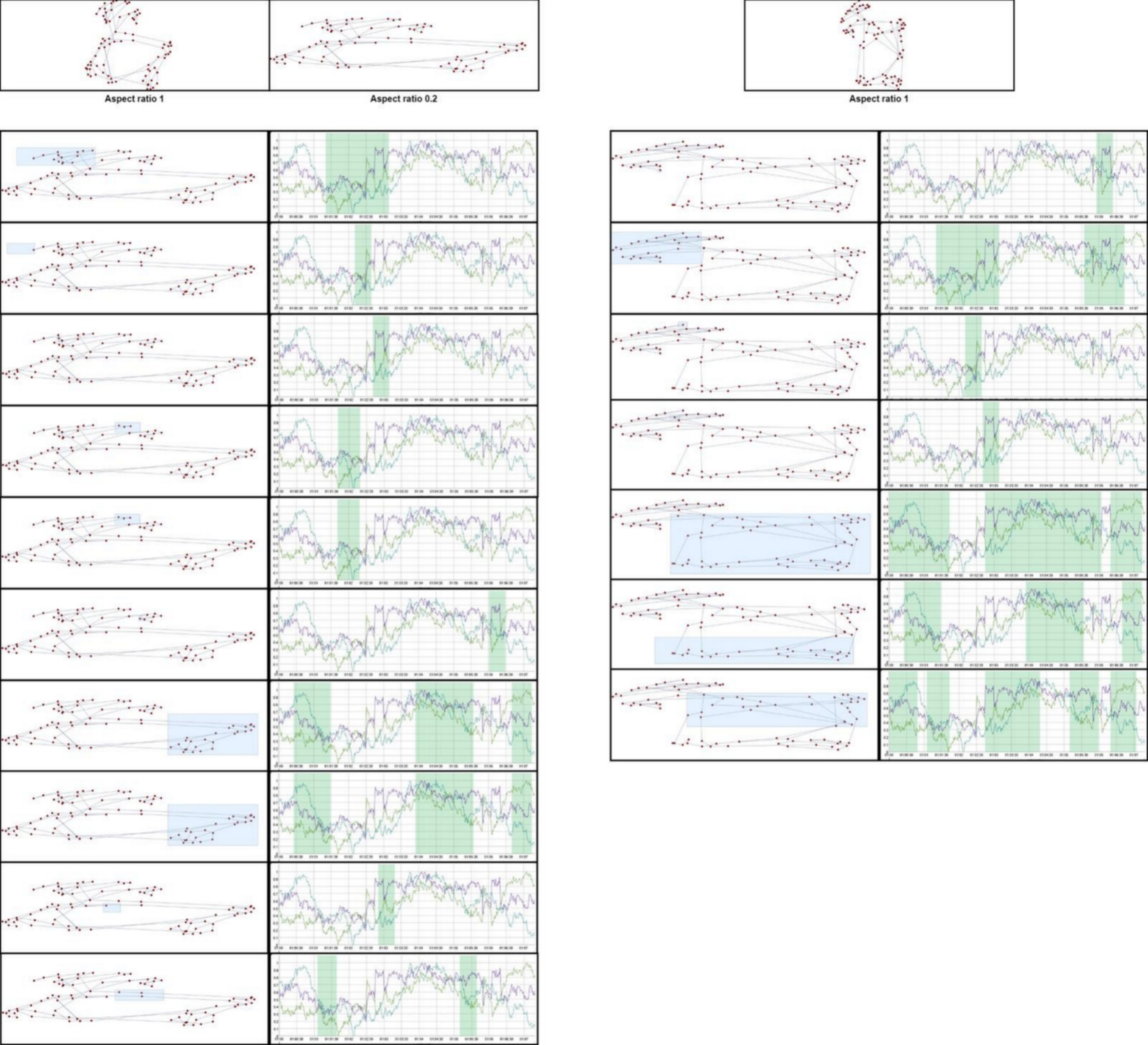}
    \caption{Embeddings projections of MOMENT-large applied to \texttt{M-Toy}.} 
    \label{fig:toy:moment:large}
\end{figure}

Figures~\ref{fig:toy:moment:base} and~\ref{fig:toy:moment:large} show the 
embedding projections for the MOMENT-base and MOMENT-large versions applied to \texttt{M-Toy}.

\FloatBarrier
\subsubsection{\texttt{Kohl's} auxiliar plots}
\begin{figure}[H]
    \centering
    \includegraphics[width=1\linewidth, pagebox=artbox]{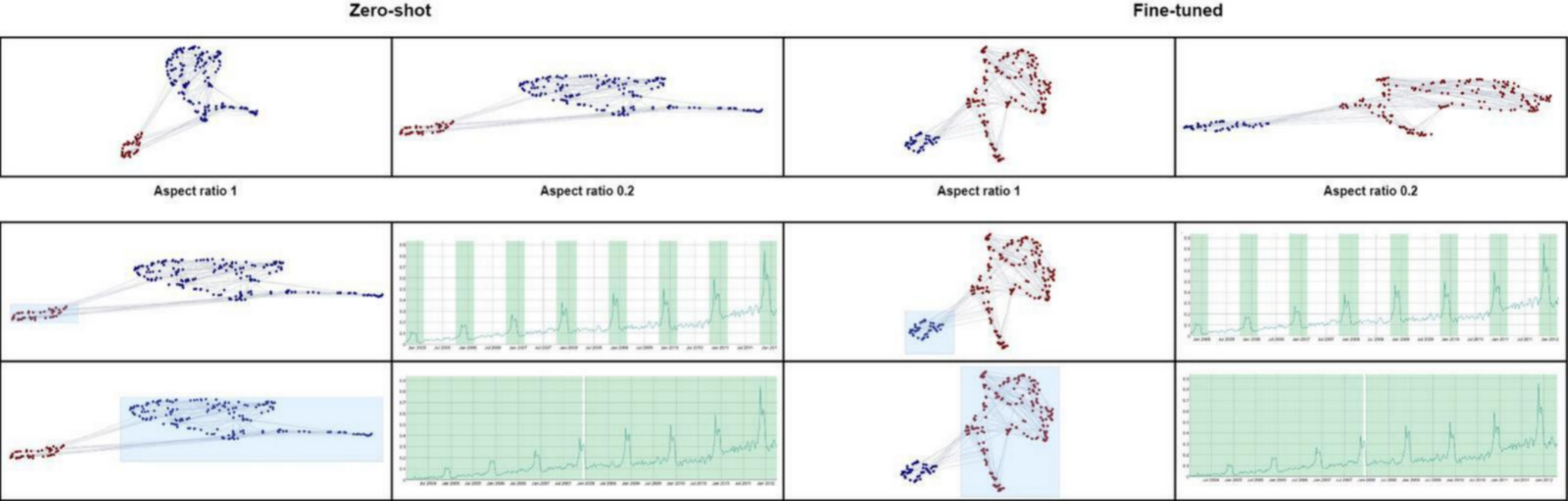}
    \caption{Embeddings projections of MOMENT-base applied to \texttt{Kohl's}.} 
    \label{fig:kohls:moment:base}
\end{figure}

\begin{figure}[H]
    \centering
    \includegraphics[width=1\linewidth, pagebox=artbox]{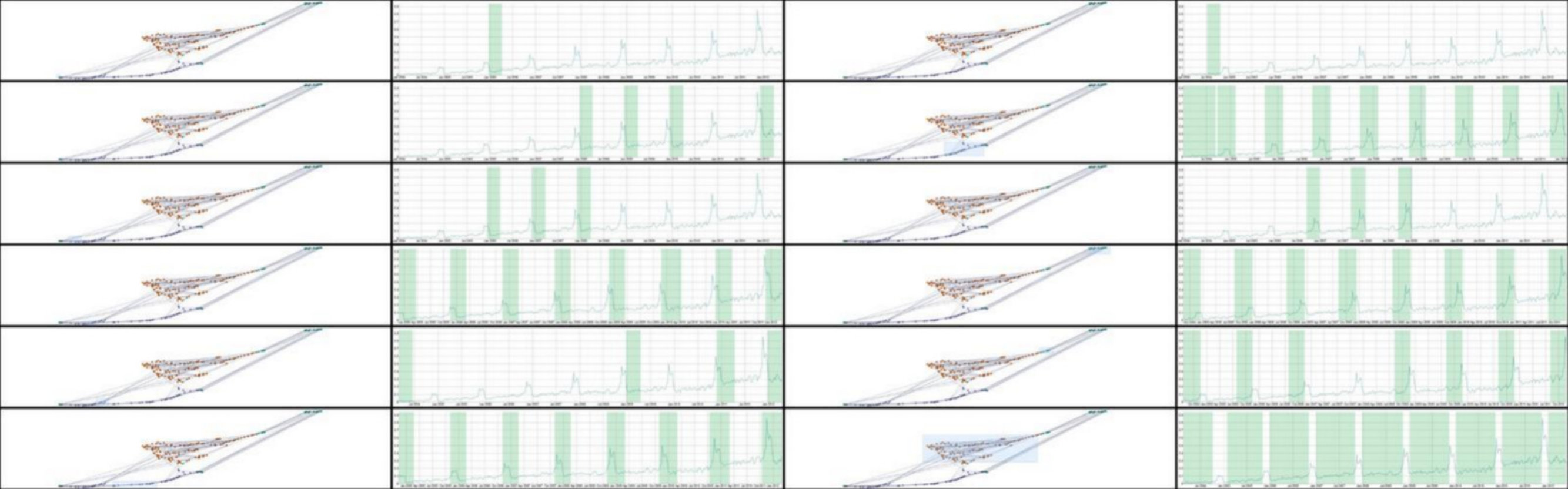}
    \caption{Embeddings projections of MOMENT-large applied to \texttt{Kohl's}.} 
    \label{fig:kohls:moment:large:zero-shot}
\end{figure}

Figures~\ref{fig:kohls:moment:base} and~\ref{fig:kohls:moment:large:zero-shot} show the visual analysis of the zero-shot executions for MOMENT-base and MOMENT-small for the \texttt{Kohl's} dataset. 

\begin{figure}[H]
    \centering
    \includegraphics[width=1\linewidth, pagebox=artbox]{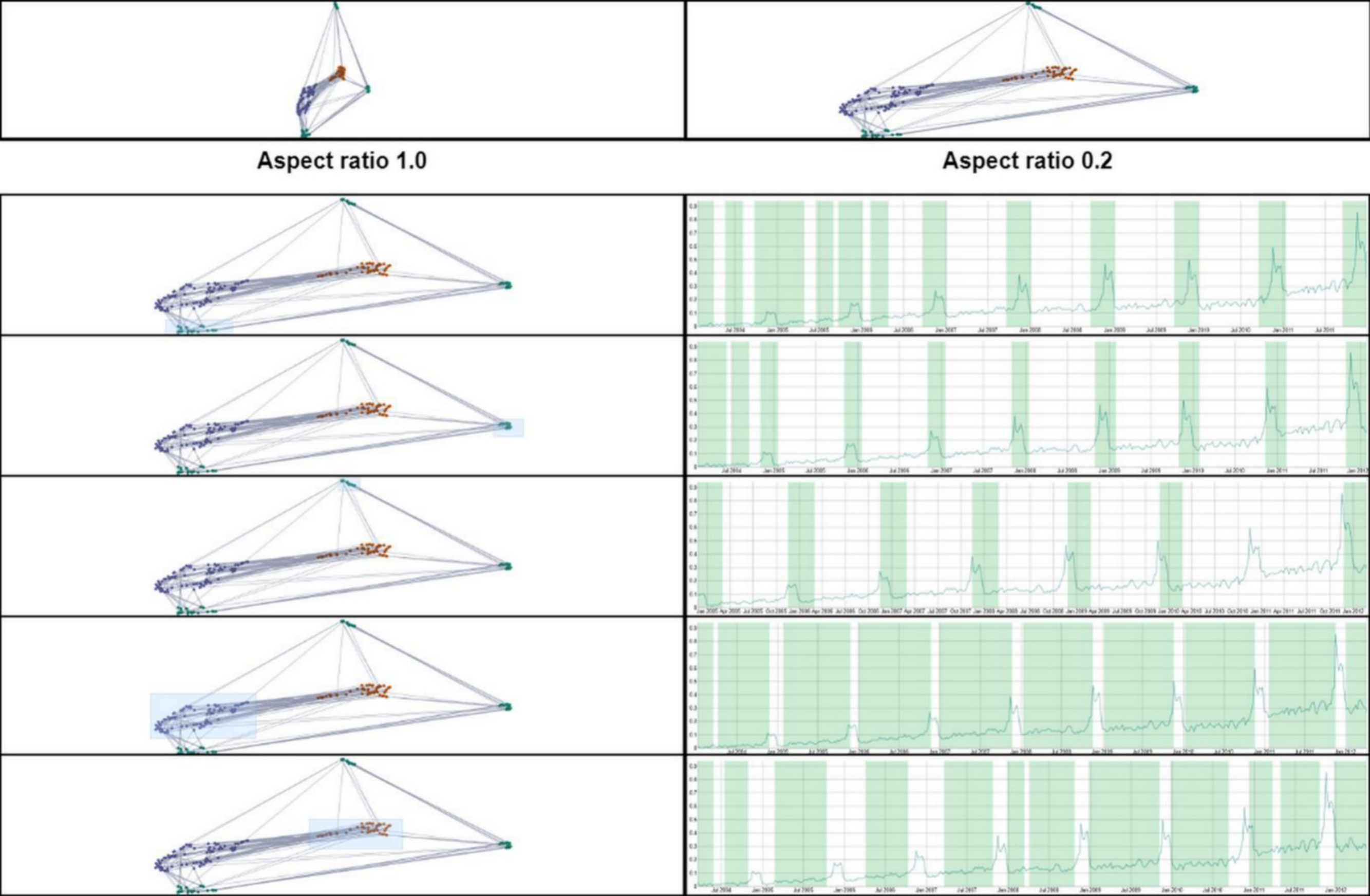}
    \caption{Embeddings projections of MOMENT-large applied to \texttt{Kohl's}.} 
    \label{fig:kohls:moment:large:finetuned}
\end{figure}

Figure~\ref{fig:kohls:moment:large:finetuned} shows the execution of the fine-tuned version of MOMENT large for \texttt{Kohl's}.

\FloatBarrier

\section{Acknowledgment}
This work has been supported by the following projects: H2020 TMA-MSCA-DN TUAI project ``Towards an Understanding of Artificial Intelligence via a transparent, open and explainable perspective'' (HORIZON-MSCA-2023-DN-01-01, Grant agreement nº: 101168344); by Strategic Networking and Development Program funded by the Ministry of Science and ICT through the National Research Foundation of Korea (RS-2023-00267476); by project PCI2022-134990-2 (MARTINI) of the CHISTERA IV Cofund 2021 program; by European Comission under IBERIFIER Plus - Iberian Digital Media Observatory (DIGITAL-2023-DEPLOY- 04-EDMO-HUBS 101158511); by EMIF managed by the Calouste Gulbenkian Foundation, in the project MuseAI; and by Comunidad Autonoma de Madrid, CIRMA-CAM Project (TEC-2024/COM-404).


\bibliographystyle{ijimai}
\bibliography{biblio}
\let\oldincludegraphics\includegraphics
\renewcommand{\includegraphics}[2][]{%
    \oldincludegraphics[#1,trim=100 40 0 0,clip]{#2}%
}

\authorbio{macu_aida}{Inmaculada Santamaria-Valenzuela}{Inmaculada Santamaria-Valenzuela is an Assistant Professor at the School of Computer Systems Engineering, Universidad Politécnica de Madrid (UPM). Her research focuses on deep learning and visual analytics for time series analysis, with particular interest in model interpretability and scalability. She has participated in the development of DeepVATS, a framework that combines deep learning with interactive visualisation tools to explore latent representations. She has presented her work at international conferences and journals and is an active member of the Applied Intelligence \& Data Analysis research group at UPM.}

\renewcommand{\includegraphics}[2][]{%
    \oldincludegraphics[#1,trim=10 10 10 10, clip]{#2}%
}
\vspace{2pt}
\authorbio{victor_google}{Victor Rodriguez-Fernandez}{Dr. Victor Rodriguez-Fernandez is an Associate Professor at the Universidad Politécnica de Madrid (UPM) and a core member of the Applied Intelligence and Data Analysis (AIDA) group. His research focuses on deep learning applications for time series analysis, emphasizing scalability and interpretability. He is the lead author of DeepVATS, a visual analytics framework for exploring latent representations in time series data. Dr. Rodríguez-Fernández has collaborated with institutions like MIT and ESA, contributing to projects on space traffic management, solar storm forecasting, and autonomous spacecraft control using large language models.}

\renewcommand{\includegraphics}[2][]{%
    \oldincludegraphics[#1,trim=40 200 80 70,clip]{#2}%
}
\vspace{2pt}
\authorbio{javi_master}{Javier Huertas-Tato}{Dr. Javier Huertas-Tato is an Assistant Professor at Universidad Politécnica de Madrid (UPM) and a researcher in the Naturla Language Processing and Deep Learning (NLP-DL) group, specializing in machine learning, disinformation detection, and social media analysis. With expertise in deep learning techniques like transformers and adversarial robustness, he has contributed to projects such as FightDIS and IBERIFIER, developing AI-driven solutions for misinformation mitigation. His work spans text and multimodal content verification, cybersecurity, and environmental applications. His latest contributions including representation learning transformer models to detect writing styles in social media18, and innovations in multimodal content verification to combat manipulated audiovisual data.}
\newline
\newline
\renewcommand{\includegraphics}[2][]{%
    \oldincludegraphics[#1,trim=30 30 0 10,clip]{#2}%
}
\vspace{2pt}
\authorbio{jong}{Jhon Hyuk Park}{Dr. James J. (Jong Hyuk) Park is a professor at Seoul National University of Science and Technology (SeoulTech), Korea, with dual Ph.D. degrees from Korea University and Waseda University. He has an extensive academic and research background, having published around 400 papers in international journals and conferences. He actively contributes to the global research community through leadership roles in conferences and editorial positions in various journals. Dr. Park has received multiple awards for his research and leadership and was recognized among the World’s Top 2\% Scientists by Stanford University in 2021. His research focuses on areas such as IoT, Cloud Computing, Blockchain, Quantum Information, Information Security, and the Metaverse.
}
\vspace{2pt}
\renewcommand{\includegraphics}[2][]{%
    \oldincludegraphics[#1,trim=0 0 0 10,clip]{#2}%
}
\authorbio{david}{David Camacho-Fernandez}{David Camacho (Senior Member, IEEE) received the Ph.D. degree (with Hons.) in computer science fromUniversidadCarlos III deMadrid,Getafe, Spain, in 2001. He is currently a Full Professor with Computer Systems Engineering Department, Universidad Politécnica deMadrid (UPM),Madrid, Spain, and the Head of the Applied Intelligence and Data Analysis researchGroup,UPM.He has authored or coauthored more than 300 journals, books, and conference papers. His research interests include machine learning (clustering/deep learning), computational intelligence (evolutionary computation, swarm intelligence), social network analysis, fake news and disinformation analysis. He has participated/led more than 50 research projects (Spanish and European: H2020, DG Justice, ISFP, and Erasmus+), related to the design and application of artificial intelligence methods for data mining and optimization for problems emerging in industrial scenarios, aeronautics, aerospace engineering, cybercrime/cyber intelligence, social networks applications, or video games among others.He was the recipient of the best thesis award in Computer Science for his Ph.D. degree}

\end{document}